%% file: main.tex
\documentclass{article}
\usepackage{adjustbox}
\usepackage{makecell}
\usepackage{booktabs}
\usepackage{tabularx}
\usepackage{array}
\usepackage{booktabs}
\usepackage{makecell}
\usepackage{multirow}
\usepackage{graphicx}
\usepackage{booktabs, makecell, multirow, amsmath, amssymb, hyperref}

\usepackage{hyperref}
\usepackage{url}
\usepackage{booktabs}
\usepackage{amsfonts}
\usepackage{amsmath}
\usepackage{amssymb}
\usepackage{nicefrac}
\usepackage{microtype}
\usepackage{xcolor}
\usepackage{graphicx}

\usepackage{wrapfig}

\usepackage{subcaption}
\usepackage{multirow}
\usepackage{algorithm}
\usepackage{algpseudocode}
\usepackage{makecell}
\usepackage{colortbl}      
\usepackage{listings}      
\usepackage{amsthm}

\newtheorem{corollary}{Corollary}
\newtheorem{proposition}{Proposition}

\definecolor{ourTeal}{HTML}{2A6F7A}
\definecolor{ourTealLight}{HTML}{3F8F9B}
\definecolor{psfRed}{HTML}{B5503C}
\definecolor{pmbPurple}{HTML}{7A5BA4}
\definecolor{mlpinnBlue}{HTML}{5683B5}
\definecolor{neusaOrange}{HTML}{F0A848}
\definecolor{refGold}{HTML}{C9A96E}
\definecolor{stubGray}{HTML}{7A7A7A}

\definecolor{bestC}{HTML}{0058A8}         
\definecolor{secondC}{HTML}{1B7A1B}        
\definecolor{thirdC}{HTML}{A5292A}

\definecolor{codeKeyword}{HTML}{0058A8}
\definecolor{codeComment}{HTML}{6A8759}
\definecolor{codeString}{HTML}{B5503C}
\definecolor{codeBG}{HTML}{F7F7F7}
\lstdefinestyle{ossm}{
  language=Python,
  basicstyle=\ttfamily\footnotesize,
  keywordstyle=\color{codeKeyword}\bfseries,
  commentstyle=\color{codeComment}\itshape,
  stringstyle=\color{codeString},
  backgroundcolor=\color{codeBG},
  frame=single,
  rulecolor=\color{black!30},
  framesep=6pt,
  xleftmargin=8pt,
  xrightmargin=8pt,
  breaklines=true,
  columns=flexible,
  showstringspaces=false,
  morekeywords={self,def,class,return,for,in,None,True,False},
  literate={->}{$\rightarrow$}1 {>=}{$\geq$}1 {<=}{$\leq$}1,
}

\usepackage[numbers,sort&compress]{natbib}
\usepackage[preprint]{neurips_2026}

\usepackage[utf8]{inputenc} % allow utf-8 input
\usepackage[T1]{fontenc}    % use 8-bit T1 fonts
\usepackage{hyperref}       % hyperlinks
\usepackage{url}            % simple URL typesetting
\usepackage{booktabs}       % professional-quality tables
\usepackage{amsfonts}       % blackboard math symbols
\usepackage{nicefrac}       % compact symbols for 1/2, etc.
\usepackage{microtype}      % microtypography
\usepackage{xcolor}         % colors
\usepackage{algorithm}
\usepackage{algpseudocode}

\title{Oscillatory State-Space Models as Inductive Biases for Physics-Informed Neural PDE Solvers}

\author{%
  Abhishek Chandra\thanks{Both authors contributed equally to this work.} \\
  KTH Royal Institute of Technology\\
  Stockholm, Sweden \\
  \And
  Taniya Kapoor\footnotemark[1]\textsuperscript{,}\thanks{Corresponding author: Taniya Kapoor (\texttt{taniya.kapoor@wur.nl}).} \\
  Wageningen University \& Research\\
  Wageningen, The Netherlands \\
  \texttt{taniya.kapoor@wur.nl} \\
}

\begin{document}

\maketitle

\begin{abstract}
Solving time-dependent partial differential equations (PDEs) is an important problem in computational science and engineering. Physics-informed neural networks (PINNs) learn PDE solutions from governing equations. However, accurately capturing temporal evolution remains challenging. Recent sequence-model-based approaches parameterize time evolution using general-purpose sequence models, which capture temporal dependencies but do not explicitly encode the structured dynamics of PDE solutions. In addition, their memory requirements can scale unfavorably with sequence length and resolution, limiting applicability in large-scale or high-dimensional settings. This work introduces a PINN approach that incorporates oscillatory state-space dynamics to represent the modal structure of PDE solutions. The proposed method leverages a linear-oscillator-based temporal evolution, together with a PDE-aware spectral basis in space. This design enables closed-form spatial differentiation and consistent enforcement of boundary conditions. The method is evaluated on forward, inverse, and high-dimensional PDE problems, including cases up to 100 spatial dimensions. The results show improved accuracy and reduced memory usage compared to recent sequence-model-based PINN approaches. Overall, this work highlights the benefits of incorporating structured dynamical priors into the temporal evolution of neural PDE solvers and suggests designing more physics-aligned and computationally efficient PINN architectures.
\end{abstract}

\section{Introduction}
\label{sec:introduction}
Partial differential equations (PDEs) are ubiquitous across science and engineering~\citep{evans2022partial}. Physics-informed neural networks (PINNs)~\citep{raissi2019physics,karniadakis2021physics} are a mesh-free machine learning method that can simulate PDEs via well-posed governing equations and incorporate sparse measurements to solve inverse problems. Although successful~\citep{molina2024modeling, moseley2022physics}, PINNs face challenges for time-dependent PDEs, including high-frequency dynamics, stiffness, long-time integration, high-order operators, and high-dimensional domains~\citep{krishnapriyan2021characterizing,wang2021understanding, wang2022when, kapoor2023physics, rathore2024challenges}. The temporal component poses particular difficulty, motivating works on causal and curriculum training~\citep{wang2024respecting,krishnapriyan2021characterizing}, time-marching and domain-decomposition methods~\citep{wight2021solving,mattey2022novel,penwarden2023unified, meng2020ppinn}.

\begin{figure}[t]
\centering
\includegraphics[width=1.0\textwidth]{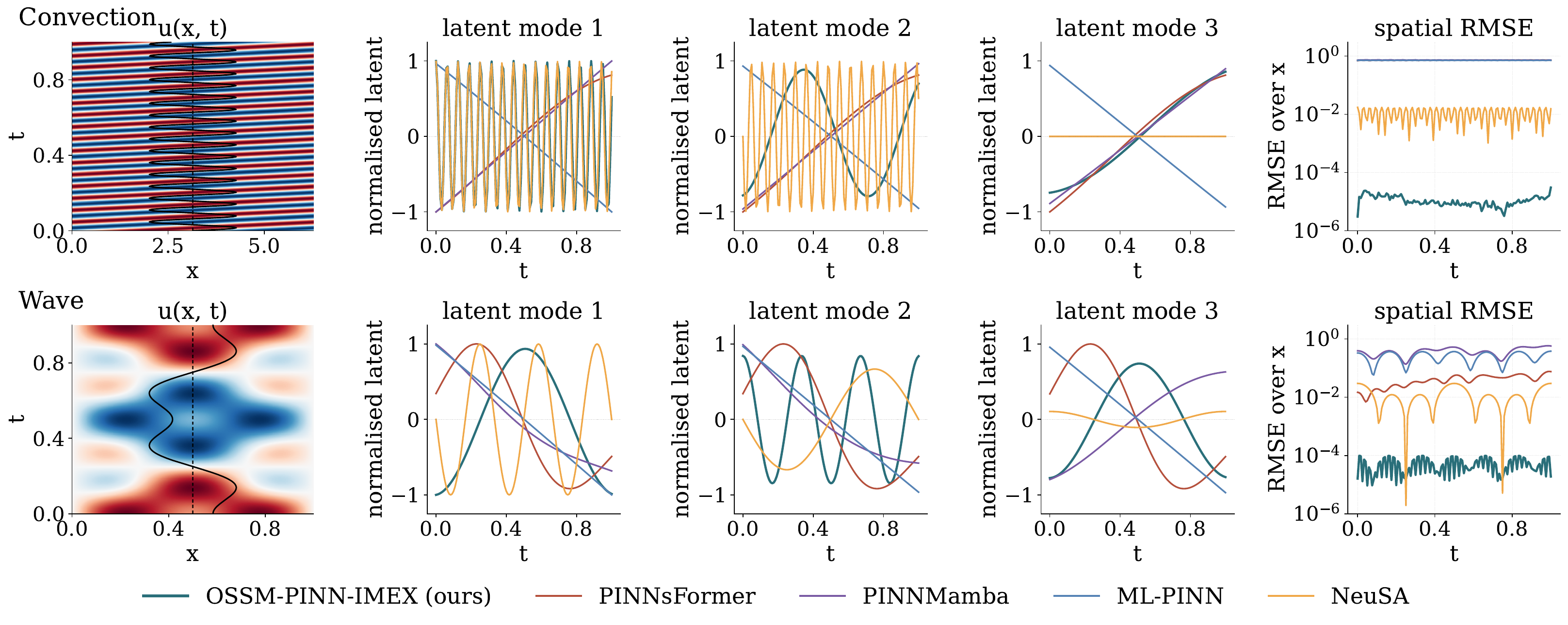}
\caption{Latent dynamics on convection (top) and wave (bottom). Columns: solution heatmap, top-3 latent dimensions per method, predicted-field error. OSSM-PINN produces bounded oscillatory latents, yielding lower error than sequence-model and neuro-spectral baselines.}
\label{fig:latent}
\vspace{-22pt}
\end{figure}

In addition, several architectural advancements that evolve PDE solutions via sequence backbones such as Transformers or modern state-space models~\citep{zhao2024pinnsformer,xu2025pinnmamba,gao2025ml} have been proposed and are hereafter referred to as \emph{sequence-model-based PINNs}. These methods provide better accuracy than coordinate multi-layer perceptron (MLP) PINNs, owing to the expressivity of the underlying sequence backbones. However, they also inherit design choices common to language and generic time-series modeling that do not reflect the modal evolution induced by PDE operators. This can be visualized in Figure~\ref{fig:latent}, top row: on convection, all three sequence-model baselines (PINNsFormer~\citep{zhao2024pinnsformer}, PINNMamba~\citep{xu2025pinnmamba}, ML-PINN~\citep{gao2025ml}) collapse their latent state into smooth, unbounded drifts rather than the oscillations the PDE inherits from its dispersion relation, leading to higher error. A second issue is computational: the pseudo-sequence strategy scales unfavorably with spatial resolution, exhausting single-GPU memory on coupled 2D problems and becoming infeasible in high-dimensional, non-rectangular, and inverse problems with real-world data settings.

A complementary direction, neuro-spectral architectures such as NeuSA~\citep{bizzi2025neuro}, projects the spatial dependence onto a fixed basis and integrates the resulting coefficient dynamics through an ordinary differential equation (ODE)-type rollout. Embedding modal structure into the architecture produces visibly improved latent dynamics; Figure~\ref{fig:latent}, bottom row, shows that on the wave equation, NeuSA's latents oscillate in a way the sequence-model baselines do not. Two challenges nevertheless remain: the time evolution is not learned from the residual, and the rollout is a sequential ODE solve that cannot be parallelized. Extended related work is discussed in SM\,\S\ref{app:extended_related_work}.

These observations lead to an open question: \emph{what inductive bias should the temporal evolution of a neural PDE solver encode?} The answer to this question should satisfy several constraints. The temporal evolution should be \emph{learnable from the residual}, so the same architecture transfers across problem classes. It should produce \emph{modal trajectories by construction}, so that natural frequencies, phase shifts, and damping patterns are available to the optimizer as the default trajectory shape. It should be \emph{parallelizable in time}, avoiding the $\mathcal{O}(N^2)$ cost of attention and the $\mathcal{O}(N)$ cost of sequential rollouts. It should be \emph{compact}, so that capacity is spent on representing the PDE rather than on a generic sequence prior; and it should \emph{decouple from the spatial representation}, so that the basis can be chosen to provide exact spatial derivatives by definition and, if possible, hard boundary-condition enforcement.

These requirements are realized jointly in the architecture proposed in this paper, \emph{oscillatory state-space model physics-informed neural networks} (OSSM-PINNs). Temporal evolution is carried out by a linear oscillatory state-space (LinOSS) block~\citep{rusch2025oscillatory}, and the spatial representation is provided by a fixed analytical basis chosen per problem family. The two halves are coupled so that the architecture learns time-evolving modal coefficients rather than an unconstrained space-time map $\mathrm{(x,t)}\mapsto \mathrm{u(x,t)}$. The method is evaluated across forward, inverse, high-dimensional, non-rectangular, large-domain, and problem-adapted-basis settings and is found to extend the reach of neural PDE solvers into regimes where sequence-model and neuro-spectral baselines face challenges.

The contributions of this work are as follows. (1) This paper identifies the temporal component of a neural PDE solver as the principal locus at which physics-aligned inductive bias should contribute, and shows that an oscillatory state-space recurrence whose spectrum is constrained to undamped oscillation is a structurally appropriate prior for this role. (2) OSSM-PINN is introduced, a PINN architecture in which an oscillatory state-space cell is coupled to a fixed spatial basis through a modewise decoder, with a single backbone applied unchanged across every benchmark. (3) Empirical results across forward, inverse, high-dimensional, non-rectangular, and large-domain benchmarks indicate that the proposed architecture addresses, under one training recipe, a set of PINN failure modes, with PDE-specific characteristic frequencies recovered from the residual. (4) A cell-swap analysis, in which only the temporal cell is replaced while the encoder, decoder, basis, optimizer, and parameter budget are held fixed, indicates that the observed performance is attributable to the oscillatory eigenstructure of the temporal cell rather than to capacity, basis, or decoder.

\section{Method}
\label{sec:method}

\begin{wrapfigure}{r}{0.45\textwidth}
  \vspace{-20pt}
  \centering
  \includegraphics[width=0.45\textwidth]{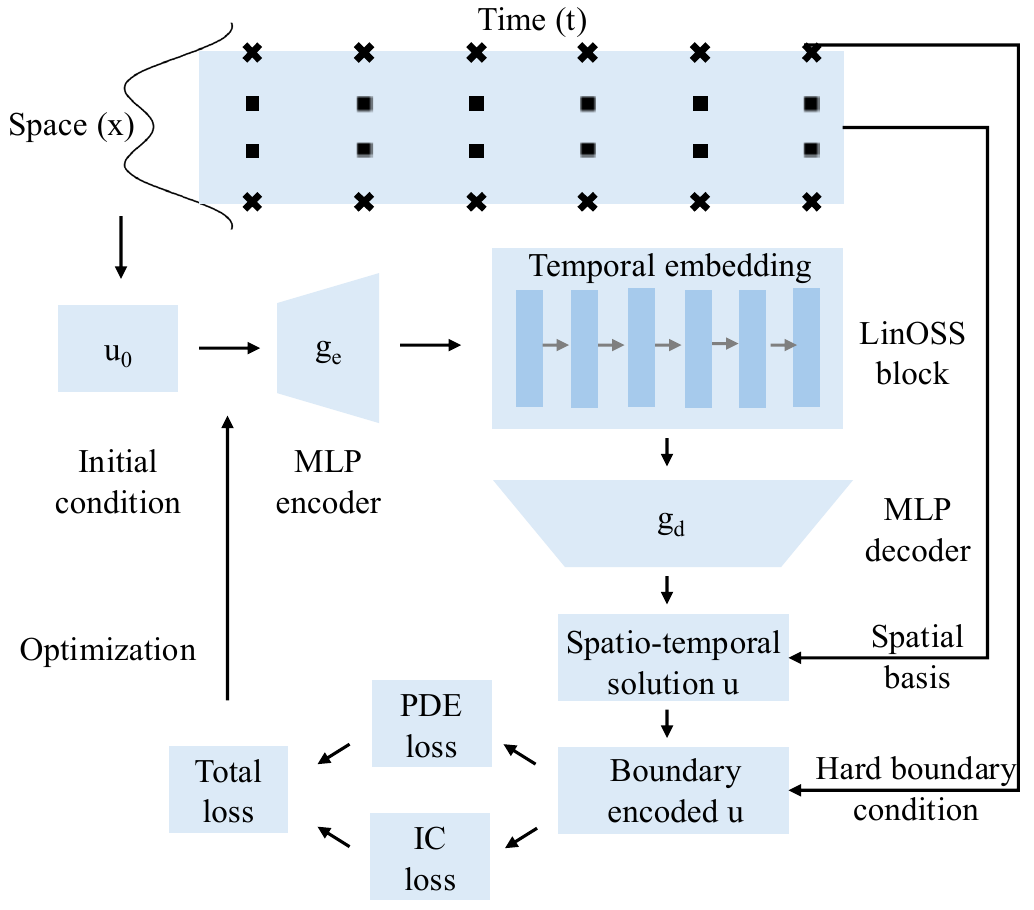}
  \vspace{-8pt}
  \caption{OSSM-PINN architecture: The initial condition is encoded into an oscillatory
  LinOSS latent state, whose temporal rollout produces modal
  coefficients through an MLP decoder. Coefficients are
  combined with spatial basis and boundary factor
  to form the spatio-temporal solution, which is trained through
  physics-informed loss.}
  \label{fig:my_figure}
  \vspace{-50pt}
\end{wrapfigure}
This section introduces OSSM-PINNs for solving time-dependent PDEs. Let $\Omega \subset \mathbb{R}^d$ be a spatial domain and $[0, T]$ the time interval. The abstract PDE could be represented as $\mathcal{F}(\mathrm{u}, \mathrm{u}_\mathrm{t}, \mathrm{u}_{\mathrm{tt}}, \nabla \mathrm{u}, \Delta \mathrm{u}, \ldots, \mathrm{x}, \mathrm{t}) = 0$ on $\Omega \times [0,T]$, where $\mathrm{u}$ is the unknown, with initial condition $\mathrm{u}(\mathrm{x},0) = \mathrm{u}_0(\mathrm{x})$, boundary condition $\mathcal{B}[\mathrm{u}](\mathrm{x},\mathrm{t}) = 0$ on $\partial\Omega \times [0,T]$. Standard PINNs learn a coordinate map $(\mathrm{x},\mathrm{t}) \mapsto \mathrm{u}(\mathrm{x},\mathrm{t})$ through a multi-layer perceptron evaluated at collocation points in the computational domain. OSSM-PINN takes a different approach by decoupling the space-time treatment, as described in detail in the following subsections. The architecture of OSSM-PINN is shown in Figure~\ref{fig:my_figure}. The pseudo code of the method is presented in SM\,\S\ref{app:pseudocode}.

\subsection{Modal Factorization}
\label{subsec:modal_factorization}
OSSM-PINN approximates the PDE solution as
\begin{equation}
\widehat{\mathrm{u}}(\mathrm{x},\mathrm{t})
=
D(\mathrm{x})
\sum_{k=1}^{K}
c_k(h(\mathrm{t}))\,\phi_k(\mathrm{x})
+
g(\mathrm{x}),
\label{eq:ossm_ansatz}
\end{equation}
where $\{\phi_k\}_{k=1}^{K}$ are spatial basis functions, $h(\mathrm{t}) \in \mathbb{R}^H$ is a temporal latent state, $c_k(\cdot)$ are scalar coefficients generated by a shared modewise decoder $g_d : \mathbb{R}^H \to \mathbb{R}^K$ (extended to $\mathbb{R}^{K \times q}$ for vector-valued PDEs of output dimension $q$), $D(\mathrm{x})$ and $g(\mathrm{x})$ are used for boundary-condition handling. The neural network thus does not learn an unconstrained space-time field but instead learns a finite set of time-evolving modal coefficients. As the basis is analytical, spatial derivatives of any order can be computed in closed form, and boundary-factor terms could be leveraged through Leibniz rule (\S\ref{subsec:basis_and_bc}), avoiding autograd evaluations through a spatial network for high-order operators.

\subsection{Oscillatory Temporal Evolution}
\label{subsec:oscillatory_evolution}
This subsection specifies the temporal evolution leading to the
modal coefficients in~\eqref{eq:ossm_ansatz}. The latent trajectory
is initialized from the initial condition, evolved through forced
linear oscillators, and discretized into a uniform time grid via a parallel scan. The initial latent state is generated by an encoder
$g_e: \mathbb{R}^{N_{\mathrm{IC}} q} \to \mathbb{R}^{2H}$, a
MLP that maps a fixed sample
$U_0 = [\mathrm{u}_0(\mathrm{x}_1^{\mathrm{IC}}), \ldots,
\mathrm{u}_0(\mathrm{x}_{N_{\mathrm{IC}}}^{\mathrm{IC}})]$ of the
initial condition (of dimension $N_{\mathrm{IC}} q$) to the
$2H$-dimensional initial LinOSS state $(y_0, z_0) = g_e(U_0)$. The latent state is then propagated by the state-space LinOSS cell, defined for hidden states $y(\mathrm{t}),
z(\mathrm{t}) \in \mathbb{R}^H$ by
\begin{equation}
\dot{y} = z, \qquad
\dot{z} = -A\, y + B\, s(\mathrm{t}),
\label{eq:linoss_continuous}
\end{equation}
where $A \in \mathbb{R}^{H \times H}$ is a nonnegative diagonal
matrix parameterized as $A = \mathrm{ReLU}(\widehat{A})$ with
$\widehat{A} \in \mathbb{R}^{H \times H}$ a freely learnable
diagonal matrix, $B \in
\mathbb{R}^H$ is a learnable input projection, and $s(\mathrm{t}) =
\mathrm{t}/T$ is a normalized time forcing. The temporal feature
passed to the modewise decoder is $h(\mathrm{t}) = y(\mathrm{t}) \in
\mathbb{R}^H$. With $A_i \geq 0$, each hidden coordinate is a forced
harmonic oscillator with angular frequency $\omega_i = \sqrt{A_i}$,
and the continuous-time eigenvalues lie on the imaginary axis at
$\pm i \sqrt{A_i}$.

Two stable discretizations are used~\citep{rusch2025oscillatory}.
The implicit-explicit (IMEX) scheme
\begin{equation}
z_n = z_{n-1} + \Delta \mathrm{t}\,(-A y_{n-1} + B s_n),
\qquad y_n = y_{n-1} + \Delta \mathrm{t}\, z_n,
\label{eq:imex}
\end{equation}
is symplectic for the autonomous part and has a discrete spectrum on
the unit circle, suiting conservative dynamics. The implicit (IM)
scheme
\begin{equation}
S = (I + \Delta \mathrm{t}^{2} A)^{-1},\quad
z_n = S(z_{n-1} - \Delta \mathrm{t}\, A y_{n-1}
+ \Delta \mathrm{t}\, B s_n),\quad
y_n = y_{n-1} + \Delta \mathrm{t}\, z_n,
\label{eq:im}
\end{equation}
introduces controlled dissipation with a discrete spectrum strictly
inside the unit disk for $A_i > 0$, suiting dissipative dynamics.
The two models are denoted OSSM-PINN-IMEX and OSSM-PINN-IM.

Both discretizations admit the linear-recurrence form
$\xi_{n+1} = M \xi_n + F_n$ with
$\xi_n = [y_n;\, z_n] \in \mathbb{R}^{2H}$, where $M$ depends on the
chosen scheme and $F_n$ carries the discrete forcing. This recurrence
is associative under $(M_2, F_2) \circ (M_1, F_1) =
(M_2 M_1, M_2 F_1 + F_2)$, so the full $N_t$-step trajectory is
computed by a parallel scan~\citep{blelloch1990prefix} in $\mathcal{O}(\log N_t)$ depth on
parallel hardware, in contrast to the $\mathcal{O}(N_t)$ depth of
sequential integrators. The output of the rollout is the trajectory
$\{h(\mathrm{t}_n)\}_{n=0}^{N_t}$, which is then mapped through the
modewise decoder $g_d : \mathbb{R}^H \to \mathbb{R}^{K}$ to generate,
at each time step, the $K$ modal coefficients
$\{c_k(h(\mathrm{t}_n))\}_{k=1}^{K}$ that drive the spatial expansion
in~\eqref{eq:ossm_ansatz}, where $K$ denotes the number of spatial
basis functions.

\subsection{Spatial Basis and Boundary-Condition Enforcement}
\label{subsec:basis_and_bc}

The spatial basis is chosen per PDE domain and
boundary condition. Periodic problems on $[a,b]$ use $K$ Fourier
features $\phi_k(\mathrm{x}) \in \{1, \sin(2\pi k(\mathrm{x}-a)/
(b-a)), \cos(2\pi k(\mathrm{x}-a)/(b-a))\}$ for $k = 1, \ldots, K$;
homogeneous Dirichlet problems on intervals use the sine modes
$\phi_k(\mathrm{x}) = \sin(k\pi(\mathrm{x}-a)/(b-a))$. Multi-
dimensional rectangular domains use the tensor product
$\Phi_{\mathbf{k}}(\mathrm{x}) = \prod_{j=1}^{d}
\phi^{(j)}_{k_j}(\mathrm{x}_j)$ over multi-indices
$\mathbf{k} = (k_1, \ldots, k_d)$ with $k_j = 1, \ldots, K_j$,
giving $K = \prod_{j=1}^{d} K_j$ effective basis functions.
%Bound-state quantum problems use Hermite or P\"oschl--Teller eigenfunctions. 
When the basis itself does not enforce the boundary condition, a
smooth boundary factor $D(\mathrm{x})$ is applied to the raw modal
sum. For homogeneous Dirichlet conditions on
$\Omega = \prod_{j=1}^{d}[a_j, b_j]$
\begin{equation}
D(\mathrm{x}) = \prod_{j=1}^{d}
\frac{(\mathrm{x}_j - a_j)(b_j - \mathrm{x}_j)}
{((b_j - a_j)/2)^{2}}
\label{eq:dirichlet_factor}
\end{equation}
vanishes on $\partial\Omega$, yielding
$\widehat{\mathrm{u}} = D\, \widehat{\mathrm{u}}_{\mathrm{raw}} + g$
with the raw modal sum
$\widehat{\mathrm{u}}_{\mathrm{raw}}(\mathrm{x}, \mathrm{t}) =
\sum_{k=1}^{K} c_k(h(\mathrm{t}))\, \phi_k(\mathrm{x})$. Spatial
derivatives of the wrapped prediction follow by the Leibniz rule,
$\partial_\mathrm{x}^m \widehat{\mathrm{u}} = \sum_{\ell=0}^{m}
\binom{m}{\ell} \partial_\mathrm{x}^\ell D \cdot
\partial_\mathrm{x}^{m-\ell} \widehat{\mathrm{u}}_{\mathrm{raw}}
+ \partial_\mathrm{x}^m g$. Basis values
$\{\phi_k(\mathrm{x}_i)\}$ and their derivatives, together with
the derivatives of $D$, are precomputed at the fixed collocation
points $\{\mathrm{x}_i\}_{i=1}^{N_x}$.

\subsection{Training Objective and Algorithm}
\label{subsec:objective_and_training}

\IfFileExists{F3_algorithm.tex}{\input{F3_algorithm.tex}}{}

The temporal evolution of \S\ref{subsec:oscillatory_evolution} and
the spatial basis of \S\ref{subsec:basis_and_bc} are combined to
form the prediction $\widehat{\mathrm{u}}(\mathrm{x},\mathrm{t})$
in~\eqref{eq:ossm_ansatz}, which is trained by minimizing a
physics-informed loss. Let $\mathcal{X} =
\{\mathrm{x}_i\}_{i=1}^{N_x}$ be the spatial collocation points and let
$\{\mathrm{t}_n\}_{n=0}^{N_t}$ be the uniform time grid produced by
the LinOSS rollout. Temporal derivatives required by $\mathcal{F}$
are computed from the predicted sequence
$\{\widehat{\mathrm{u}}(\mathrm{x}_i, \mathrm{t}_n)\}$ via $p$th-order
finite differences ($p \in \{4, 6\}$, sixth-order by default) with
one-sided stencils near the time boundaries. 
The training objective is
\begin{equation}
\mathcal{L} = \frac{\lambda_{\mathrm{PDE}}}{N_x(N_t+1)}
\sum_{i,n} \|\mathcal{F}(\widehat{\mathrm{u}};\mathrm{x}_i,
\mathrm{t}_n)\|_2^2
+ \frac{\lambda_{\mathrm{IC}}}{N_x}
\sum_i \|\widehat{\mathrm{u}}(\mathrm{x}_i, 0)
- \mathrm{u}_0(\mathrm{x}_i)\|_2^2,
\label{eq:total_loss}
\end{equation}
where $\mathcal{F}(\widehat{\mathrm{u}}; \mathrm{x}_i,
\mathrm{t}_n)$ denotes the PDE differential operator evaluated on
the prediction at the grid point $(\mathrm{x}_i, \mathrm{t}_n)$. The hyperparameters $\lambda_{\mathrm{PDE}}$ and $\lambda_{\mathrm{IC}}$ denoted the loss weights.
Optimization uses an Adam warm-up phase followed by L-BFGS. 
The full training procedure, including the modification used for inverse problems, is summarized in Algorithm~\ref{alg:ossm-pinn}.

\subsection{Approximation Guarantees}
\label{subsec:consistency}

The modal factorization in~\eqref{eq:ossm_ansatz}, together with the
discretization choices in \S\ref{subsec:oscillatory_evolution} and
\S\ref{subsec:objective_and_training}, admits a residual-loss bound
that vanishes as the basis size, the LinOSS hidden width, and the
inverse time step grow. The result is a composition of three
factors: spectral basis truncation, the universal
approximation property of the LinOSS
cell~\citep{rusch2025oscillatory}, and the
$p$th-order accuracy of the finite-difference temporal derivative.

\begin{proposition}[Residual consistency]
\label{prop:ossm_consistency}
Let $\Omega \subset \mathbb{R}^d$ be compact, let
$\mathrm{u}^\star : \Omega \times [0,T] \to \mathbb{R}^q$ be a
solution of the PDE with $\mathcal{F}$ locally Lipschitz
in its arguments on a bounded neighborhood of $\mathrm{u}^\star$,
and let $\mathrm{u}^\star$ admit the modal representation
$\mathrm{u}^\star = D(\mathrm{x}) \sum_{k=1}^{\infty} a_k(\mathrm{t})
\phi_k(\mathrm{x}) + g(\mathrm{x})$ with $D, g, \{\phi_k\}$
sufficiently smooth and the basis compatible with the prescribed
boundary condition. Let $m_x, m_t$ be the highest spatial and
temporal derivative orders in $\mathcal{F}$, and denote by
$\varepsilon_{\mathrm{basis}}(K)$ and
$\varepsilon_{\mathrm{coeff}}(K, H, \theta)$ the
$C^{m_x, m_t}(\Omega \times [0,T])$-norm errors of, respectively,
the order-$K$ modal truncation of $\mathrm{u}^\star$ and the
OSSM-PINN approximation $\widehat{\mathrm{u}}_{K, H, \theta}$ at
fixed truncation. Assume $\varepsilon_{\mathrm{basis}}(K) \to 0$ as
$K \to \infty$, and that for each $K$ there exists a parameterization
$\theta$ such that $\varepsilon_{\mathrm{coeff}}(K, H, \theta) \to 0$
as the hidden width $H$ and decoder capacity grow. Then
$\widehat{\mathrm{u}}_{K, H, \theta} \to \mathrm{u}^\star$ in the
$C^{m_x, m_t}$ norm, and the continuous residual converges to zero in
$L^2$. If temporal derivatives are evaluated by a consistent
$p$th-order finite-difference scheme on a uniform grid with step
$\Delta \mathrm{t}$, the discrete PDE residual term
in~\eqref{eq:total_loss} satisfies
\begin{equation}
\frac{1}{N_x(N_t+1)} \sum_{i,n}
\| \mathcal{F}(\widehat{\mathrm{u}}_{K,H,\theta};
\mathrm{x}_i, \mathrm{t}_n) \|_2^2
\leq C \left(
\varepsilon_{\mathrm{basis}}(K)^{2}
+ \varepsilon_{\mathrm{coeff}}(K, H, \theta)^{2}
+ \Delta \mathrm{t}^{\,2p}
\right),
\label{eq:ossm_residual_consistency_bound}
\end{equation}
for $C > 0$ depending on the Lipschitz constant of $\mathcal{F}$ and
on derivative bounds for $\widehat{\mathrm{u}}_{K,H,\theta}$.
\end{proposition}

\begin{corollary}[Exactness for finite-mode dynamics]
\label{cor:finite_oscillatory_exactness}
If $\mathrm{u}^\star$ has exactly $K$ active modes, $a(\mathrm{t})$
is realized exactly by the OSSM-PINN cell and decoder, and the
boundary condition is enforced exactly by $D$, $g$, and the basis,
then $\varepsilon_{\mathrm{basis}} = \varepsilon_{\mathrm{coeff}} =
0$ and \eqref{eq:ossm_residual_consistency_bound} reduces to
$\mathcal{O}(\Delta \mathrm{t}^{\,2p})$.
\end{corollary}

\section{Experiments}
\label{sec:numerical-experiments}

\textbf{Goal:} The empirical evaluations aim to demonstrate four key advantages
of OSSM-PINN. First, OSSM-PINN accurately solves time-dependent
PDEs with stiff, oscillatory, high-order, nonlinear, and high-dimensional
dynamics, outperforming existing sequence-model-based and neuro-spectral PINN
approaches. Second, the performance gains arise from the
proposed oscillator-spectral inductive bias rather than increased model
capacity. Third, the same architecture applies to forward problems, inverse problems, problem-adapted bases,
non-rectangular domains, and large spatial domains. Fourth, the
modal factorization leads to compact memory usage and fast inference relative
to sequence-model-based baselines.

\textbf{Experiment setup:} OSSM-PINN is evaluated on twelve time-dependent PDE benchmarks spanning
forward, inverse, non-rectangular geometry, high-dimensional, and problem-adapted-basis
settings. The suite includes transport, reaction, wave, beam, Taylor--Green
vortex, triangular-domain heat, Schr\"odinger in 5 and 100 spatial domains, KdV inverse
recovery, and sea surface temperature (SST) advection--diffusion~\citep{huang2021oisst}. Full equations and benchmark details
are given in SM\,\S\ref{app:forward_problems} and
Table~\ref{tab:experiment-suite}, with overview visualizations in
Figures~\ref{fig:F8a-forward} and~\ref{fig:F8b-specialised}. The baselines are PINNsFormer~\citep{zhao2024pinnsformer},
PINNMamba~\citep{xu2025pinnmamba}, ML-PINN~\citep{gao2025ml}, and
NeuSA~\citep{bizzi2025neuro} under the same 24~GiB single-GPU compute
envelope. Methods exceeding memory are marked \textsc{OOM}; \textsc{ST} and
\textsc{IG} denote NeuSA stiff-integrator failure and geometry infeasibility.
Both OSSM-PINN-IM and OSSM-PINN-IMEX are reported, using the same backbone across all benchmarks and changing only the PDE residual, spatial basis, number of modes, collocation budget, and temporal rollout length. Hyperparameters for OSSM-PINN and
all four baselines are collected in SM\,\S\ref{app:hyperparameters} Tables~\ref{tab:s4-ossm-recipe}--\ref{tab:s8-neusa-recipe}.

\begin{figure}[t]
  \centering
  \includegraphics[width=0.9\textwidth]{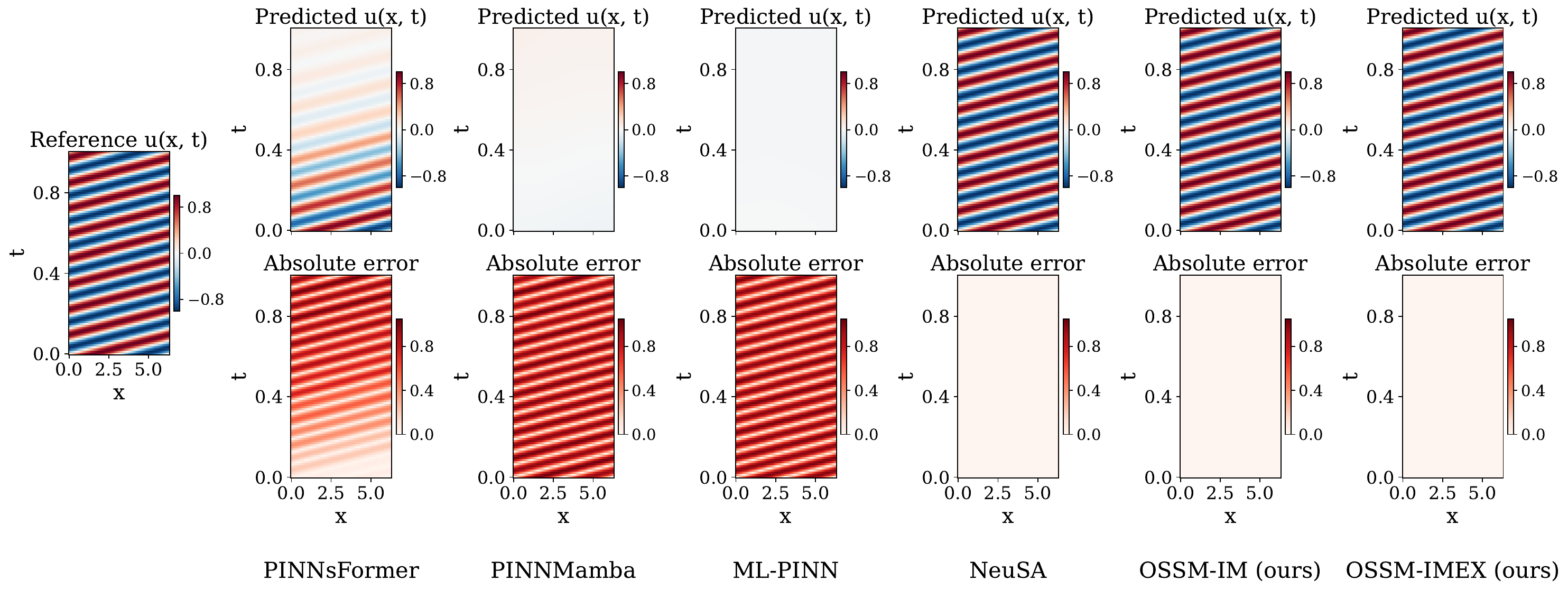}
  \caption{Convection ($\beta=50$): predicted $u(x,t)$ fields and absolute errors for each method.}
  \label{fig:fwd-convection}
\end{figure}

\begin{figure}[t]
  \centering
  \includegraphics[width=0.9\textwidth]{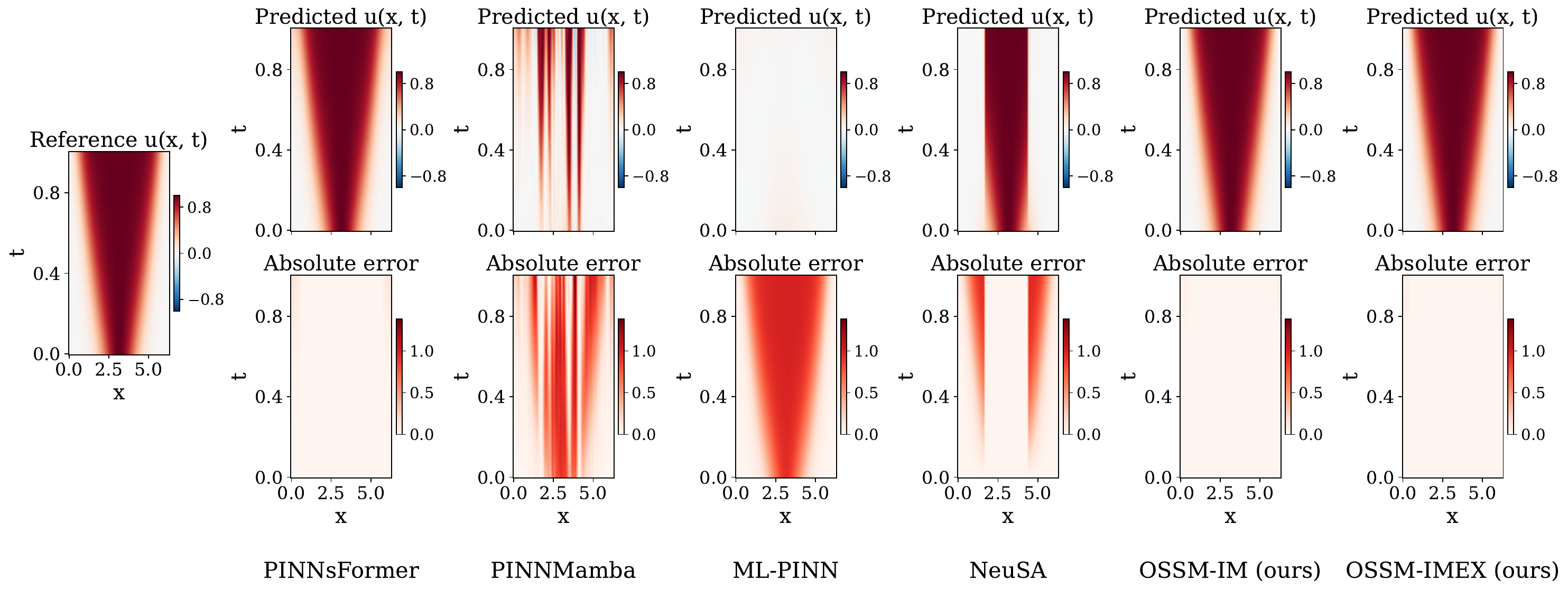}
  \caption{Reaction: predicted $u(x,t)$ fields and absolute errors for each method.}
  \label{fig:fwd-reaction}
\end{figure}

\textbf{Evaluation:} Approximately comparable parameter counts are maintained across all models to highlight advantages from the oscillatory-spectral inductive bias rather than overparameterization. All models are trained using Adam warm-up followed by L-BFGS with strong-Wolfe line search.
Evaluation metrics are relative mean absolute error (rMAE), relative root mean squared error (rRMSE), and maximum error defined in SM\,\S\ref{app:implementation}. Unless stated otherwise, results are averaged over three random seeds $s\in\{0,1,2\}$; seed-to-seed variability is shown in Figure~\ref{fig:seed-stability}.
\begin{table}[t]
  \caption{Performance on twelve PDE benchmarks. Best in \textcolor{bestC}{\textbf{blue}}, second in \textcolor{secondC}{green}, third in \textcolor{thirdC}{red}. Italic factors: improvement over best baseline.}
  \label{tab:headline}
  \centering
  \footnotesize
  \setlength{\tabcolsep}{2pt}
  \renewcommand{\arraystretch}{1.0}
  \resizebox{\textwidth}{!}{%
\begin{tabular}{c l c c c c c c}
\toprule
& PDE & PINNsFormer & PINNMamba & ML-PINN & NeuSA & \makecell{OSSM-PINN-IM\\\scriptsize\emph{(ours)}} & \makecell{OSSM-PINN-IMEX\\\scriptsize\emph{(ours)}} \\
\midrule
\multirow{6}{*}{\rotatebox[origin=c]{90}{Forward}} & Convection ($\beta{=}50$) & 6.7e-1 & 1.0e+0 & 1.0e+0 & \textcolor{thirdC}{3.2e-3} & \textcolor{secondC}{3.3e-5}\,{\scriptsize\itshape ($\times$99)} & \textbf{\textcolor{bestC}{3.0e-5}}\,{\scriptsize\itshape ($\times$107)} \\
\cmidrule(lr){2-8}
 & Convection ($\beta{=}100$) & 1.0e+0 & 1.0e+0 & 1.0e+0 & \textcolor{thirdC}{3.2e-3} & \textcolor{secondC}{5.6e-4}\,{\scriptsize\itshape ($\times$6)} & \textbf{\textcolor{bestC}{1.5e-5}}\,{\scriptsize\itshape ($\times$219)} \\
\cmidrule(lr){2-8}
 & Reaction & \textcolor{thirdC}{1.1e-2} & 7.0e-1 & 9.8e-1 & 2.9e-1 & \textbf{\textcolor{bestC}{2.9e-3}}\,{\scriptsize\itshape ($\times$4)} & \textcolor{secondC}{3.1e-3}\,{\scriptsize\itshape ($\times$4)} \\
\cmidrule(lr){2-8}
 & Wave & 7.6e-2 & 7.3e-1 & 4.9e-1 & \textcolor{thirdC}{5.5e-3} & \textcolor{secondC}{4.3e-5}\,{\scriptsize\itshape ($\times$128)} & \textbf{\textcolor{bestC}{4.2e-5}}\,{\scriptsize\itshape ($\times$132)} \\
\cmidrule(lr){2-8}
 & Euler-Bernoulli (classical) & \textcolor{stubGray}{\scriptsize\textsc{OOM}} & \textcolor{stubGray}{\scriptsize\textsc{OOM}} & \textcolor{thirdC}{3.3e-1} & 3.3e+0 & \textbf{\textcolor{bestC}{6.6e-5}}\,{\scriptsize\itshape ($\times$5.0k)} & \textcolor{secondC}{6.7e-5}\,{\scriptsize\itshape ($\times$4.9k)} \\
\cmidrule(lr){2-8}
 & Euler-Bernoulli (extended) & \textcolor{stubGray}{\scriptsize\textsc{OOM}} & \textcolor{stubGray}{\scriptsize\textsc{OOM}} & \textcolor{thirdC}{1.1e+0} & 5.8e+0 & \textbf{\textcolor{bestC}{1.1e-5}}\,{\scriptsize\itshape ($\times$98.4k)} & \textcolor{secondC}{1.5e-5}\,{\scriptsize\itshape ($\times$71.1k)} \\
\midrule
\midrule
\multirow{4}{*}{\rotatebox[origin=c]{90}{High-dimension}} & Taylor-Green 2D & \textcolor{stubGray}{\scriptsize\textsc{OOM}} & \textcolor{stubGray}{\scriptsize\textsc{OOM}} & \textcolor{stubGray}{\scriptsize\textsc{OOM}} & \textcolor{stubGray}{\scriptsize\textsc{OOM}} & \textbf{\textcolor{bestC}{2.6e-3}} & \textcolor{secondC}{3.9e-3} \\
\cmidrule(lr){2-8}
 & Heat (triangular domain) & \textcolor{stubGray}{\scriptsize\textsc{OOM}} & \textcolor{stubGray}{\scriptsize\textsc{OOM}} & \textcolor{stubGray}{\scriptsize\textsc{OOM}} & \textcolor{stubGray}{\scriptsize\textsc{IG}} & \textbf{\textcolor{bestC}{7.7e-3}} & \textcolor{secondC}{7.7e-3} \\
\cmidrule(lr){2-8}
 & Schr\"odinger (5D) & \textcolor{stubGray}{\scriptsize\textsc{OOM}} & \textcolor{stubGray}{\scriptsize\textsc{OOM}} & \textcolor{stubGray}{\scriptsize\textsc{OOM}} & \textcolor{stubGray}{\scriptsize\textsc{OOM}} & \textbf{\textcolor{bestC}{1.1e-3}} & \textcolor{secondC}{1.1e-3} \\
\cmidrule(lr){2-8}
 & Schr\"odinger (100D) & \textcolor{stubGray}{\scriptsize\textsc{OOM}} & \textcolor{stubGray}{\scriptsize\textsc{OOM}} & \textcolor{stubGray}{\scriptsize\textsc{OOM}} & \textcolor{stubGray}{\scriptsize\textsc{OOM}} & \textcolor{secondC}{9.3e-4} & \textbf{\textcolor{bestC}{8.7e-4}} \\
\midrule
\midrule
\multirow{2}{*}{\rotatebox[origin=c]{90}{Inverse}} & KdV & \textcolor{stubGray}{\scriptsize\textsc{OOM}} & \textcolor{stubGray}{\scriptsize\textsc{OOM}} & \textcolor{stubGray}{\scriptsize\textsc{OOM}} & \textcolor{stubGray}{\scriptsize\textsc{ST}} & \textcolor{secondC}{6.4e-3} & \textbf{\textcolor{bestC}{6.1e-3}} \\
\cmidrule(lr){2-8}
 & SST 2D adv-diff & \textcolor{stubGray}{\scriptsize\textsc{OOM}} & \textcolor{stubGray}{\scriptsize\textsc{OOM}} & \textcolor{stubGray}{\scriptsize\textsc{OOM}} & \textcolor{stubGray}{\scriptsize\textsc{OOM}} & \textcolor{secondC}{8.1e-2} & \textbf{\textcolor{bestC}{8.0e-2}} \\
\bottomrule
\end{tabular}}
  \\[3pt]
  {\scriptsize\textcolor{stubGray}{OOM: out-of-memory; IG: infeasible geometry; ST: stiff PDE.}}
\end{table}

\subsection{Forward Time-Dependent PDEs}
\label{sec:exp:forward}

The first experiment evaluates whether the proposed modal architecture
generalizes across diverse time-dependent PDE regimes (Exp.~1 in
SM\,\S\ref{app:experiment_overview} Table~\ref{tab:experiment-suite}), including transport, reaction, hyperbolic,
and high-order dynamics. Table~\ref{tab:headline} summarizes the
results, with detailed rMAE, rRMSE, and maximum-error in SM\,\S\ref{app:forward_problems}
Table~\ref{tab:s1-full-results}. OSSM-PINN is consistently the most accurate
method across the forward benchmarks, with the largest gains on
high-frequency convection, wave propagation, and Euler--Bernoulli beam
problems. These results indicate that the oscillator--spectral factorization is
effective across both first- and second-order-in-time PDEs, as well as
high-order spatial operators. Figures~\ref{fig:fwd-convection} and \ref{fig:fwd-reaction}, together with SM\,\S\ref{app:forward_problems} Figures~\ref{fig:fwd-wave}, \ref{fig:fwd-eb-extended}, and
\ref{fig:sup-convection-beta100}--\ref{fig:sup-eb}, show that OSSM-PINN closely
matches the reference fields while avoiding common baseline failure modes such
as near-zero collapse and spurious oscillations. A hyperparameter sensitivity analysis in Table~\ref{tab:sensitivity} confirms that the proposed model is robust to variations in mode count and rollout depth.

\begin{figure}[t]
  \vspace{-8pt}
  \centering
  \begin{minipage}[t]{0.48\textwidth}
    \centering
    \IfFileExists{figs/figs5.pdf}{%
      \includegraphics[width=\textwidth]{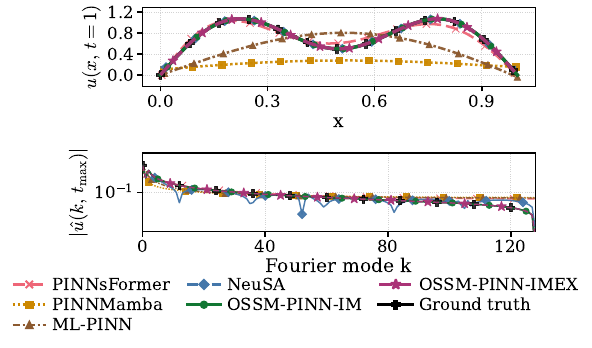}}{}
    \vspace{-8pt}
    \caption{Frequency-domain comparison on the wave equation: predictions and Fourier spectra.}
    \label{fig:sup-freq-wave}
  \end{minipage}\hfill
  \begin{minipage}[t]{0.48\textwidth}
    \centering
    \IfFileExists{figs/figs6.pdf}{%
      \includegraphics[width=\textwidth]{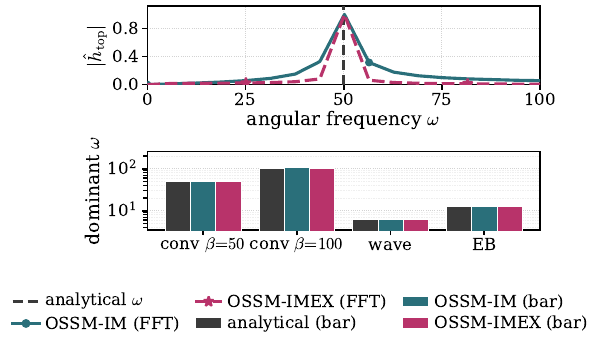}}{}
    \vspace{-8pt}
    \caption{Learned dispersion: dominant latent frequencies match analytical PDE dispersion.}
    \label{fig:sup-dispersion}
  \end{minipage}
  \vspace{-20pt}
\end{figure}

\textbf{Spectral and latent structure:}
Figures~\ref{fig:sup-freq-wave} and \ref{fig:sup-dispersion} probe the
mechanism behind these gains. In the wave benchmark, OSSM-PINN preserves the
reference Fourier spectrum while baselines either introduce spurious
high-frequency content or do not match the ground truth. In latent space, the dominant LinOSS frequencies align
with analytical PDE dispersion across convection, wave, and Euler--Bernoulli
benchmarks. These diagnostics confirm that OSSM-PINN improves accuracy by learning physically grounded modal time scales rather than acting as a generic sequence model.
Additional spectral and latent results are in SM\,\S\ref{app:spectral_latent} Figure~\ref{fig:sup-latents-all} and
Figures~\ref{fig:sup-freq-conv50}, \ref{fig:sup-freq-conv100}, \ref{fig:sup-freq-reaction}, \ref{fig:sup-freq-eb}, and~\ref{fig:sup-freq-eb-ext}.
% ----------------------------------------------------------------------
\subsection{Ablations: Which Components Matter?}
\label{sec:exp:ablations-main}

\begin{wraptable}{r}{0.5\textwidth}
  \vspace{-12pt}
  \caption{Architecture ablations (rMAE, IMEX). ``Default'' is the full OSSM-PINN; each row removes or replaces one component.}
  \label{tab:ablations}
  \centering
  \footnotesize
  \setlength{\tabcolsep}{3pt}
  \begin{tabular}{l c c c}
  \toprule
  Variant & Conv.\ $\beta{=}50$ & Euler-Bern. & Wave \\
  \midrule
  Default & 3.04e-5 & 6.72e-5 & 4.15e-5 \\
  $-$ IC encoder & 5.37e-5 & 3.84e-2 & 7.30e-2 \\
  $-$ MLP decoder & 1.27e-4 & 4.34e-5 & 6.79e-5 \\
  $-$ Fourier basis & 7.10e-1 & 9.95e-3 & 1.00e+0 \\
  $-$ 6th-order FD & 2.75e-2 & 1.24e-4 & 3.99e-4 \\
  $-$ FP64 & 7.97e-4 & 2.38e-3 & 8.30e-3 \\
  $-$ LinOSS & 1.00e+0 & 1.00e+0 & 1.00e+0 \\
  \bottomrule
  \end{tabular}
  \vspace{-10pt}
\end{wraptable}

This section analyzes the contribution of each structural component (Exp.~2 in
SM\,\S\ref{app:experiment_overview} Table~\ref{tab:experiment-suite}). Each ablation removes or alters one
component at a time and is evaluated on three one-dimensional benchmarks:
convection ($\beta=50$), wave, and Euler--Bernoulli beam.

\begin{wraptable}{r}{0.5\textwidth}
  \vspace{-14pt}
  \caption{Plug-in extensions (rMAE, IMEX). Each row adds one enhancement on top of the default OSSM-PINN backbone.}
  \label{tab:additions}
  \centering
  \footnotesize
  \setlength{\tabcolsep}{3pt}
  \begin{tabular}{l c c c}
  \toprule
  Variant & Conv.\ $\beta{=}50$ & E-Bern. & Wave \\
  \midrule
  Default & 3.04e-5 & 6.72e-5 & 4.15e-5 \\
  $+$ gPINN & 2.74e-5 & 5.28e-5 & 6.75e-5 \\
  $+$ RWF & 3.00e-5 & 2.73e-5 & 1.60e-5 \\
  $+$ post-L-BFGS & 3.04e-5 & 4.74e-5 & 3.75e-5 \\
  $+$ time FD-6 & 3.04e-5 & 7.83e-5 & 2.91e-5 \\
  \bottomrule
  \end{tabular}
  \vspace{-14pt}
\end{wraptable}

The ablation results are presented in Table~\ref{tab:ablations} for the IMEX
discretization; the full table for both IM and IMEX variants is
SM\,\S\ref{app:ablations} Table~\ref{tab:ablations-full}. The results indicate
that the performance gains are not attributable to model size alone. Removing
the LinOSS rollout ($-$LinOSS) causes all three benchmarks to collapse to
rMAE $\approx 1$, since the latent state no longer evolves in time. Replacing
the spectral basis with a learned MLP basis ($-$Fourier basis) degrades
accuracy by four to five orders of magnitude on convection and wave,
demonstrating the value of fixed spatial modes and analytic derivatives.
Reducing the temporal finite-difference order from sixth to second
($-$6th-order FD) increases convection rMAE by nearly three orders of
magnitude, while switching to single precision ($-$FP64) uniformly degrades
all benchmarks by one to two orders of magnitude. These results confirm that
OSSM-PINN's accuracy depends on the joint contribution of the oscillatory
rollout, the spectral basis, high-order temporal derivatives, and
double-precision arithmetic. A cell-swap analysis (SM\,\S\ref{app:ablations} Tables~\ref{tab:ssm-ablation-conv} and~\ref{tab:ssm-ablation-reac}) further isolates the oscillatory eigenstructure as the source of gains compared to state-of-the-art non-oscillatory state-space models, and a method-attribute comparison is provided in SM\,\S\ref{app:experiment_overview} Table~\ref{tab:T1-attributes}.

Furthermore, four plug-in enhancements are considered in Table~\ref{tab:additions} (Exp.~3):
gradient-enhanced residuals (gPINN)~\citep{yu2022gradient}, random weight factorization (RWF)~\citep{wang2022random},
post-training L-BFGS refinement~\citep{kiyani2025which}, and sixth-order temporal finite differences.
All four integrate without modification to the oscillator rollout, spectral
basis, or decoder; the full IM/IMEX results are in SM\,\S\ref{app:ablations} Table~\ref{tab:additions-full}. Among these, RWF provides the most consistent improvement,
reducing rMAE by up to $2\times$ on the Euler--Bernoulli benchmark.

% ----------------------------------------------------------------------
\subsection{Inverse Problems}
\label{sec:exp:inverse-main}
\begin{wrapfigure}{r}{0.28\textwidth}
  \vspace{-14pt}
  \centering
  \includegraphics[width=0.27\textwidth]{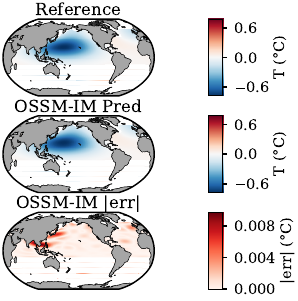}
  \vspace{-8pt}
  \caption{SST inverse at $t\!=\!23$\,mo: reference (top), OSSM-IM prediction (mid), absolute error (bot). Full results in Figure~\ref{fig:sup-inv-sst}.}
  \label{fig:sst-preview}
  \vspace{-15pt}
\end{wrapfigure}
This section evaluates whether the same architecture can learn unknown PDE coefficients from sparse observations without architectural changes (Exp.~4 in
SM\,\S\ref{app:experiment_overview} Table~\ref{tab:experiment-suite}). Unknown coefficients are trainable scalars optimized jointly with the network weights.
Two inverse problems are considered: KdV $u_t+\lambda_1 u u_x+\lambda_2 u_{xxx}=0$ ($\lambda_1=1$, $\lambda_2=0.0025$) and a 2D sea-surface-temperature advection--diffusion problem $T_t+c_xT_x+c_yT_y=\kappa(T_{xx}+T_{yy})$ ($c_x=0.5$, $c_y=0.05$, $\kappa=0.01$).

The learned coefficients are close to the ground truth in both inverse
problems: on KdV, OSSM-PINN-IM estimates $\lambda_1$ and $\lambda_2$ with
relative errors of $0.07\%$ and $0.06\%$, respectively, while all four
baselines fail on both benchmarks. Detailed quantitative results are reported
in SM\,\S\ref{app:inverse_problems} Table~\ref{tab:s3-inverse-recovery}.
A representative snapshot at $t=23$\,mo is shown in Figure~\ref{fig:sst-preview}; full predicted fields across four monthly snapshots are provided in SM\,\S\ref{app:inverse_problems} Figures~\ref{fig:sup-inv-kdv} and~\ref{fig:sup-inv-sst}, while
Figures~\ref{fig:sup-inv-kdv-evolution} and~\ref{fig:sup-inv-sst-evolution}
show the coefficient convergence trajectories.
The frequency-domain analysis in Figure~\ref{fig:sup-freq-kdv} further confirms that the inferred KdV forward solution matches the reference spectrum down to the FFT noise floor, demonstrating that the oscillator-spectral factorization is effective for inverse parameter estimation.

% ----------------------------------------------------------------------
\subsection{Problem-Adapted Spectral Bases}
\label{sec:exp:basis-main}

\begin{wraptable}{r}{0.55\textwidth}
  \vspace{-12pt}
  \caption{Spatial-basis ablation on quantum benchmarks (rMAE); bracketed factors give the win over NeuSA.}
  \label{tab:basis}
  \centering
  \footnotesize
  \setlength{\tabcolsep}{4pt}
  \begin{tabular}{l l l c c}
  \toprule
  & Method & Basis & IM & IMEX \\
  \midrule
  \multirow{3}{*}{\rotatebox[origin=c]{90}{QHO}\hspace{-2pt}}
   & NeuSA & Fourier & \multicolumn{2}{c}{2.67e-2} \\
   \cmidrule(l){2-5}
   & \multirow{2}{*}{OSSM-PINN}
       & Fourier  & \textcolor{thirdC}{9.18e-3}\,{\scriptsize\itshape ($\times$3)} & 9.51e-3\,{\scriptsize\itshape ($\times$3)} \\
   & & Hermite  & \textbf{\textcolor{bestC}{1.47e-4}}\,{\scriptsize\itshape ($\times$181)} & \textcolor{secondC}{1.87e-4}\,{\scriptsize\itshape ($\times$143)} \\
  \midrule
  \multirow{3}{*}{\rotatebox[origin=c]{90}{PT}\hspace{-2pt}}
   & NeuSA & Fourier & \multicolumn{2}{c}{\textcolor{thirdC}{1.77e-2}} \\
   \cmidrule(l){2-5}
   & \multirow{2}{*}{OSSM-PINN}
       & Fourier  & 1.92e-1 & 6.63e-2 \\
   & & PT  & \textcolor{secondC}{2.19e-3}\,{\scriptsize\itshape ($\times$8)} & \textbf{\textcolor{bestC}{1.68e-3}}\,{\scriptsize\itshape ($\times$11)} \\
  \bottomrule
  \end{tabular}
  \vspace{-10pt}
\end{wraptable}
A key advantage of OSSM-PINN is that the spatial basis is modular: it can match the eigenstructure of the target PDE while leaving the temporal rollout, decoder, and training unchanged (Exp.~5 in Table~\ref{tab:experiment-suite}). We replace the default Fourier basis with Hermite eigenfunctions for the quantum harmonic oscillator and P\"oschl--Teller eigenfunctions for the Schr\"odinger equation with a $\mathrm{sech}^2$ potential. Basis functions are illustrated in Figure~\ref{fig:F7-bases}, quantitative results in Table~\ref{tab:basis}, and prediction fields in Figures~\ref{fig:basis-qho} and~\ref{fig:basis-pt}; full values are in SM\,\S\ref{app:adapted_basis} Table~\ref{tab:basis-full}.
Using the Hermite basis reduces QHO rMAE from $9.5\times10^{-3}$ to $1.9\times10^{-4}$ ($50\times$), and the PT eigenbasis reduces rMAE from $6.6\times10^{-2}$ to $1.7\times10^{-3}$ ($40\times$), both at zero architectural cost. These results confirm the modal interpretation: when the spatial basis aligns with the PDE operator, OSSM-PINN only needs to learn the time evolution of modal coefficients.

% ----------------------------------------------------------------------
\subsection{Geometry, Large Domains, High Dimensions, and Efficiency}
\label{sec:exp:geometry-cost-main}

This section tests whether the modal factorization generalizes beyond standard 1D periodic settings to diverse geometric and dimensional regimes.

\textbf{Non-rectangular and large spatial domains:} On a triangular heat equation, OSSM-PINN handles the non-rectangular geometry without mesh generation: the spatial basis is constructed on the triangle directly and the LinOSS rollout proceeds identically. On the extended Euler--Bernoulli beam ($[0,8\pi]$), the same architecture remains accurate on a domain $8\times$ larger than the standard benchmark. Errors are in Table~\ref{tab:headline}; prediction fields in SM\,\S\ref{app:highdim_geometry} Figures~\ref{fig:sup-triangle} and~\ref{fig:fwd-eb-extended}.

\textbf{Coupled multi-component PDEs:} The Taylor--Green vortex evaluates coupled 2D flow dynamics (velocity and pressure). OSSM-PINN accurately reconstructs all three field components (SM\,\S\ref{app:highdim_geometry} Figure~\ref{fig:sup-tgv}), while all sequence-model baselines exceed the 24\,GiB memory budget since their collocation cost scales quadratically in two dimensions.

\begin{figure}[t]
  \vspace{-14pt}
  \centering
  \begin{minipage}[t]{0.48\textwidth}
    \centering
    \includegraphics[width=\textwidth]{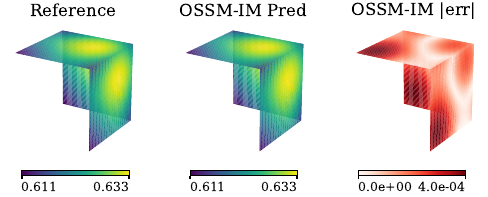}
    \vspace{-10pt}
    \caption{Schr\"odinger 100D: $|\psi(x_1,\ldots,x_{100},t)|^2$ at $t\!=\!\pi/2$ showing a slice in $(x_1,x_2,x_3)$. Reference (left), OSSM-IM prediction (center), absolute error (right).}
    \label{fig:schrodinger-preview}
  \end{minipage}\hfill
  \begin{minipage}[t]{0.48\textwidth}
    \centering
    \includegraphics[width=\textwidth]{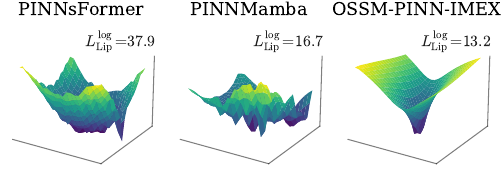}
    \vspace{-10pt}
    \caption{Loss landscape on convection ($\beta\!=\!100$) along top-2 Hessian eigenvectors: PINNsFormer, PINNMamba, OSSM-PINN-IMEX. Lower $L_{\mathrm{Lip}}^{\log}$ indicates a smoother basin.}
    \label{fig:landscape-preview}
  \end{minipage}
  \vspace{-18pt}
\end{figure}

\textbf{High-dimensional PDEs:} The 5D and 100D Schr\"odinger benchmarks test scalability with spatial dimension. Figure~\ref{fig:schrodinger-preview} shows the 100D slice: the prediction matches the reference with errors below $5\times10^{-4}$. Full results are in SM\,\S\ref{app:highdim_geometry} Figures~\ref{fig:sup-schrodinger-5d} and~\ref{fig:sup-schrodinger-100d}. Because the basis is tensorized and the rollout is dimension-agnostic, the architecture applies without modification.

\textbf{Optimization landscape and computational cost:} Figure~\ref{fig:landscape-preview} contrasts loss landscapes on high-frequency convection: OSSM-PINN-IMEX ($L_{\mathrm{Lip}}^{\log}\!=\!13.2$) yields the smoothest basin, versus PINNMamba ($16.7$) and PINNsFormer ($37.9$) which exhibits multiple sharp local minima.
The full comparison is in SM\,\S\ref{app:spectral_latent} Figure~\ref{fig:sup-landscape-conv100}; training loss histories in SM\,\S\ref{app:training_dynamics} Figures~\ref{fig:sup-loss-curves} and~\ref{fig:sup-loss-curves-im}; hardware and cost details in SM\,\S\ref{app:cost} Table~\ref{tab:s2-cost}.

\section{Conclusion, limitations and future work}
\textbf{Conclusion: }We introduced OSSM-PINNs, a physics-informed neural PDE solver that factorizes
time-dependent solutions into spatial modes whose coefficients evolve
through a learnable oscillatory state-space rollout. This design answers the
motivating question of what temporal inductive bias a neural PDE solver should
encode: the temporal model should be learnable from the residual, be oscillatory and modal by
construction, and should be compact, parallelizable, and decoupled from the spatial
representation. 

Across twelve PDE benchmarks spanning forward, inverse, high-dimensional,
non-rectangular, and problem-adapted-basis settings, OSSM-PINNs achieve lower
error and improved memory efficiency relative to sequence-model-based and
neuro-spectral baselines. On representative forward problems, the method reduces
rMAE by up to $219\times$ on high-frequency convection, $132\times$ on wave
propagation, and $9.8\times 10^{4}\times$ on the extended Euler--Bernoulli
benchmark compared with the state-of-the-art baselines. In inverse KdV recovery,
OSSM-PINN estimates the two unknown coefficients with relative errors below
$0.1\%$, while in problem-adapted quantum benchmarks, replacing a generic
Fourier basis with operator-aligned eigenbases improves rMAE by roughly
$40$--$50\times$ without changing the temporal architecture.

Spectral and latent analyses show that these gains are physically
grounded: the learned dominant latent frequencies align
with analytical PDE dispersion, while the predicted fields avoid the collapse
modes and spurious high-frequency spectra observed in several baselines.
Ablations further confirm that the improvements are not due to parameter count
alone, but depend jointly on the oscillatory rollout, analytical spatial basis,
high-order temporal derivatives, and numerical precision. The results suggest
that neural PDE solvers should not treat time as a generic sequence. Encoding oscillatory modal structure provides an
effective and efficient inductive bias for physics-informed learning.

\textbf{Limitations and future work: }OSSM-PINNs assume that the target solution is well represented by a finite set of spatial modes with structured temporal evolution. This assumption may be less effective for shocks, moving discontinuities, intermittent localized structures, or strongly non-modal dynamics, where fixed global bases may require many modes. Future work should extend the framework with adaptive, localized, multi-resolution, or domain-decomposed bases.

The current implementation uses fixed collocation points, uniform temporal grids, and manually selected problem-aware bases. While these choices enable cached analytical derivatives, high-order finite differences, and efficient parallel rollout, they may be inefficient for multiscale or highly localized phenomena. Future directions include adaptive residual sampling, adaptive temporal refinement, causal curricula, automated basis selection, and broader basis libraries.

\newpage
\bibliographystyle{unsrtnat}
\bibliography{ref_main}

% \newpage
% \input{checklist.tex}

\newpage
\appendix

\section*{Supplementary Material}

\vspace{6pt}
\noindent\textbf{Contents}
\vspace{4pt}

\noindent\begin{minipage}{\textwidth}
\renewcommand{\arraystretch}{1.15}
\normalsize
\noindent\begin{tabularx}{\textwidth}{@{}r@{\;\;}X@{\quad}r@{}}
\S\ref{app:extended_related_work} & Extended Related Work & p.\,\pageref{app:extended_related_work} \\
\S\ref{app:pseudocode} & OSSM-PINN Pseudocode & p.\,\pageref{app:pseudocode} \\
\S\ref{app:experiment_overview} & Experiment Overview and Benchmark Suite & p.\,\pageref{app:experiment_overview} \\
\S\ref{app:implementation} & Implementation Details & p.\,\pageref{app:implementation} \\
\S\ref{app:forward_problems} & Forward-Problem Definitions and Results & p.\,\pageref{app:forward_problems} \\
\S\ref{app:spectral_latent} & Spectral and Latent Diagnostics & p.\,\pageref{app:spectral_latent} \\
\S\ref{app:ablations} & Ablation Studies & p.\,\pageref{app:ablations} \\
\S\ref{app:inverse_problems} & Inverse Problems & p.\,\pageref{app:inverse_problems} \\
\S\ref{app:adapted_basis} & Problem-Adapted Spectral Bases & p.\,\pageref{app:adapted_basis} \\
\S\ref{app:highdim_geometry} & High-Dim., Geometry, and Large-Domain Experiments & p.\,\pageref{app:highdim_geometry} \\
\S\ref{app:hyperparameters} & Hyperparameters and Training Recipes & p.\,\pageref{app:hyperparameters} \\
\S\ref{app:cost} & Computational Cost & p.\,\pageref{app:cost} \\
\S\ref{app:training_dynamics} & Training Dynamics and Seed Stability & p.\,\pageref{app:training_dynamics} \\
\end{tabularx}
\end{minipage}
\vspace{6pt}

% ======================================================================
\section{Extended Related Work}
\label{app:extended_related_work}
% ======================================================================

This section expands the related-work discussion in Section~\ref{sec:introduction}.

\subsection{Physics-Informed Neural Networks}
\label{app:rw:pinn_basics}

The idea of training neural networks to satisfy differential equations was introduced by \citet{lagaris1998artificial}, who constructed trial solutions that exactly satisfy boundary conditions and minimized the PDE residual. \citet{berg2018unified} extended this to complex geometries using distance-function masks. The modern PINN formulation~\citep{raissi2019physics, karniadakis2021physics} combined automatic differentiation with deep MLPs and demonstrated forward and inverse solving on a range of canonical problems. DeepXDE~\citep{lu2021deepxde} made the approach widely accessible through software implementation.

PINNs face failure modes that motivate architectural research. \citet{krishnapriyan2021characterizing} showed that vanilla PINNs fail on stiff transport even when the underlying solution is smooth. \citet{wang2021understanding} traced failures to imbalanced gradient flow between loss terms. \citet{wang2022when} provided a neural tangent kernel analysis revealing spectral bias toward low frequencies, and \citet{rathore2024challenges} showed that PINN loss surfaces contain narrow valleys with ill-conditioned curvature. These analyses motivate two complementary directions: improving the training procedure (loss weighting, optimizers, sampling) and improving the architecture.

\subsection{Training Improvements: Weighting, Optimization, and Sampling}
\label{app:rw:loss_opt}

Adaptive loss weighting~\citep{wang2021understanding} adjusts the balance between PDE, IC, and BC loss terms during training based on gradient statistics. Gradient-enhanced PINNs (gPINN)~\citep{yu2022gradient} augment the residual loss with derivative-matching terms, penalizing not only the PDE residual but also its spatial gradients. \citet{kiyani2025which} performed a systematic optimizer comparison and found that a second-order phase (L-BFGS) is often decisive for reaching high accuracy, confirming the Adam-then-L-BFGS recipe used throughout this work. Residual-driven collocation~\citep{wu2023comprehensive} places more training points where the residual is largest, improving efficiency on stiff problems. R3 sampling~\citep{daw2023mitigating} and failure-informed strategies~\citep{gao2023failure} extend this idea.

\subsection{Temporal Treatment: Causality and Time-Marching}
\label{app:rw:temporal}
Causal training~\citep{wang2024respecting} reweights residual contributions so that earlier time steps are satisfied before later ones, preventing the optimizer from fitting late-time artifacts at the expense of early-time accuracy. \citet{wight2021solving} introduced adaptive time-stepping for Allen--Cahn and Cahn--Hilliard equations. \citet{mattey2022novel} proposed sequential segment-by-segment methods that solve the PDE on one time window before advancing to the next. \citet{penwarden2023unified} unified these causal sweeping strategies into a common framework. \citet{meng2020ppinn} introduced parareal PINN (PPINN), applying the parareal algorithm to decompose time into subintervals solved in parallel. \citet{du2021evolutional} treated temporal evolution as a parameterized dynamical system.

\subsection{Sequence-Model-Based PINNs}
\label{app:rw:sequence_pinn}
PINNsFormer~\citep{zhao2024pinnsformer} reformulates the PINN input as a pseudo-sequence of time steps processed by a transformer with a wavelet activation function. PINNMamba~\citep{xu2025pinnmamba} adapts the same pseudo-sequence strategy to the Mamba architecture~\citep{gu2024mamba}, replacing attention with selective state-space scanning. ML-PINN~\citep{gao2025ml} combines Mamba with bidirectional LSTM to capture both forward and backward temporal dependencies.

These methods improve on coordinate MLPs for time-dependent PDEs, but they import a generic sequence inductive bias that was designed for language or audio, not for PDE solutions. Their latent representations drift smoothly rather than oscillate at PDE-characteristic frequencies (Figure~\ref{fig:latent}). The pseudo-sequence formulation also couples temporal and spatial resolution: each time step processes the full spatial collocation grid, causing memory to scale as $\mathcal{O}(N_t \times N_x)$. This leads to out-of-memory failures on 2D and high-dimensional problems (Table~\ref{tab:headline}). OSSM-PINN decouples temporal from spatial resolution, the LinOSS rollout operates on a compact $H$-dimensional state independent of $N_x$, and its oscillatory eigenstructure produces latent trajectories that naturally capture PDE characteristic frequencies.

\subsection{State-Space Models for Sequence Modeling}
\label{app:rw:ssm}

\textbf{Relevance:} OSSM-PINN's temporal cell is a state-space model. This subsection explains the lineage and the specific variant (LinOSS) that makes the architecture suitable for PDE solving.

S4~\citep{gu2022efficiently} demonstrated that structured state-space layers with HiPPO initialization can model long-range dependencies in sequences. S5~\citep{smith2023simplified} simplified to a diagonal state-space formulation. LRU~\citep{orvieto2023resurrecting} showed that carefully initialized linear recurrences without structured parameterization can match SSM performance. Mamba~\citep{gu2024mamba} introduced selective (input-dependent) dynamics, achieving strong results on language modeling.

A critical design choice in SSMs is where the eigenvalues of the state transition matrix lie. Most variants place eigenvalues inside the unit disk, producing exponentially decaying dynamics suited to language and audio where information is consumed and forgotten. This conflicts with conservative PDE dynamics where energy is preserved and solutions oscillate indefinitely. LinOSS~\citep{rusch2025oscillatory} constrains eigenvalues to the imaginary axis, yielding forced harmonic oscillators that neither grow nor decay. The IMEX discretization further ensures $|$eigenvalues$| = 1$ exactly, providing a symplectic time stepper. The cell-swap ablation (Tables~\ref{tab:ssm-ablation-conv} and~\ref{tab:ssm-ablation-reac}) isolates this spectral constraint as the source of OSSM-PINN's gains over alternative SSM cells (S5, LRU, Mamba) used in the same backbone.

\subsection{Operator Learning and Domain Decomposition}
\label{app:rw:operator}
A similar line of research is neural operators. While physics-informed neural network based methods focus on single instance PDE solving without any data, neural operators focus on learning the governing operator with data, mostly. Hence, these domains are not directly comparable. Fourier neural operators (FNO)~\citep{li2021fourier}, DeepONet~\citep{lu2021learning}, and graph-based variants~\citep{li2020neural} learn solution operators from supervised data corpora, mapping initial/boundary conditions to solution fields across a family of PDEs. Physics-informed DeepONet~\citep{wang2021learning} and spectral neural operators~\citep{fanaskov2023spectral} combine operator learning with residual supervision. These methods require training data (typically from a numerical solver), whereas OSSM-PINN trains from the PDE residual alone.

Domain decomposition methods for PINNs---Conservative PINN~\citep{jagtap2020conservative} and XPINN~\citep{jagtap2021extended}---partition the spatial domain into subdomains with interface continuity constraints. Fourier-feature networks~\citep{wang2021eigenvector} mitigate spectral bias through random feature embeddings.

\newpage
\section{OSSM-PINN Pseudocode}
\label{app:pseudocode}

\begin{lstlisting}[style=ossm]
class OSSMCell:  # Linear Oscillatory State-Space cell
    def __init__(self, H, dt):
        self.A = Parameter(uniform(0, A_max))   # learnable omega^2
        self.B = Linear(1, H)                   # input projection
        self.dt = dt
    def propagator(self):
        A, dt = relu(self.A), self.dt           # A >= 0
        M = [[1 - dt**2*A, dt], [-dt*A, 1]]    # IMEX (symplectic)
        return M, dt**2, dt

class OSSMPINN:
    def __init__(self, K, H, N_t, dt, basis, bc):
        self.encoder = MLP(n_ic -> H)           # IC encoder
        self.cell = OSSMCell(H, dt)             # oscillatory cell
        self.decoder = MLP(H -> K * n_out)      # mode-coefficient decoder
        self.basis = basis                      # Fourier / Hermite / PT
        self.bc = bc                            # hard-BC factor B(x)
    def forward(self, u0, x):
        y0, z0 = self.encoder(u0)               # encode IC
        M, ay, az = self.cell.propagator()
        Bu = self.cell.B(linspace(0, 1, N_t))   # forcing
        F = stack([ay*Bu, az*Bu], dim=-1)
        h = parallel_scan(M, F, [y0, z0])[:,:,0]  # rollout
        c = self.decoder(h).reshape(N_t+1, K, n_out)
        phi = self.basis(x)                     # (n_x, K)
        return self.bc(x) * einsum('tkq,xk->txq', c, phi)

def parallel_scan(M, F, x0):  # O(log N) associative scan
    cum_M, cum_F = M.expand(T, ...), F
    for stride in [1, 2, 4, ...]:
        cum_M[stride:] = cum_M[stride:] @ cum_M[:-stride]
        cum_F[stride:] = cum_M[stride:] @ cum_F[:-stride] + cum_F[stride:]
    return cat([x0.unsqueeze(0), cum_M @ x0 + cum_F])

def compute_loss(model, pde, x, u0, dt):
    u = model.forward(u0, x)
    # Spatial derivs: closed-form from basis (no autodiff)
    phi, dphi, d2phi = model.basis.eval_with_derivs(x)
    c = model.decoder(...)
    u_x = model.bc(x) * einsum('tkq,xk->txq', c, dphi) + ...
    u_xx = ...                                  # Leibniz rule
    # Temporal derivs: 6th-order finite differences
    u_t = fd_coeffs(order=6) @ u / dt
    # PDE residual (problem-specific, architecture-agnostic)
    res = pde.residual(u, u_t, u_x, u_xx, ...)
    return w_pde*mean(res**2) + w_ic*mean((u[0]-u0)**2)

def train(model, pde, x, u0):  # Adam warm-up + L-BFGS
    for step in range(n_adam):
        compute_loss(...).backward(); Adam.step()
    for step in range(n_lbfgs):
        LBFGS.step(lambda: compute_loss(...))
\end{lstlisting}

\section{Experiment Overview and Benchmark Suite}
\label{app:experiment_overview}

Figures~\ref{fig:F8a-forward} and~\ref{fig:F8b-specialised} provide annotated visual overviews of all forward and inverse benchmarks. Table~\ref{tab:experiment-suite} summarizes the full experimental suite, listing each experiment, the problems it contains, and its purpose. Table~\ref{tab:T1-attributes} compares method-level attributes across the PINN-family architectures evaluated in this work.

\IfFileExists{figs/F8a_forward_showcase.pdf}{%
\begin{figure}[ht]
  \centering
  \includegraphics[width=\textwidth]{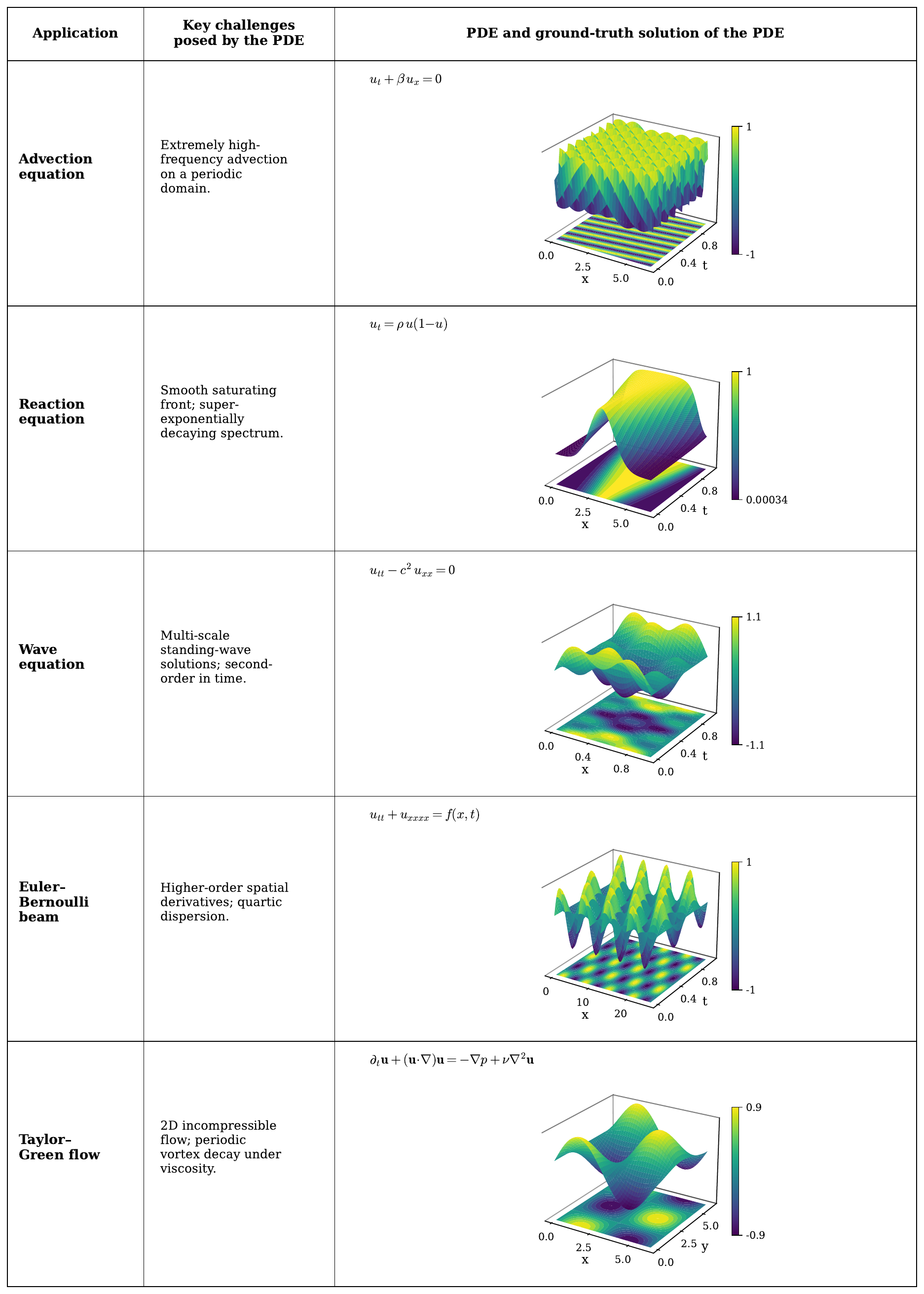}
  \caption{Forward-problem benchmark overview: key challenge and ground-truth solution for each PDE.}
  \label{fig:F8a-forward}
\end{figure}}{}

\IfFileExists{figs/F8b_specialised_showcase.pdf}{%
\begin{figure}[ht]
  \centering
  \includegraphics[width=\textwidth]{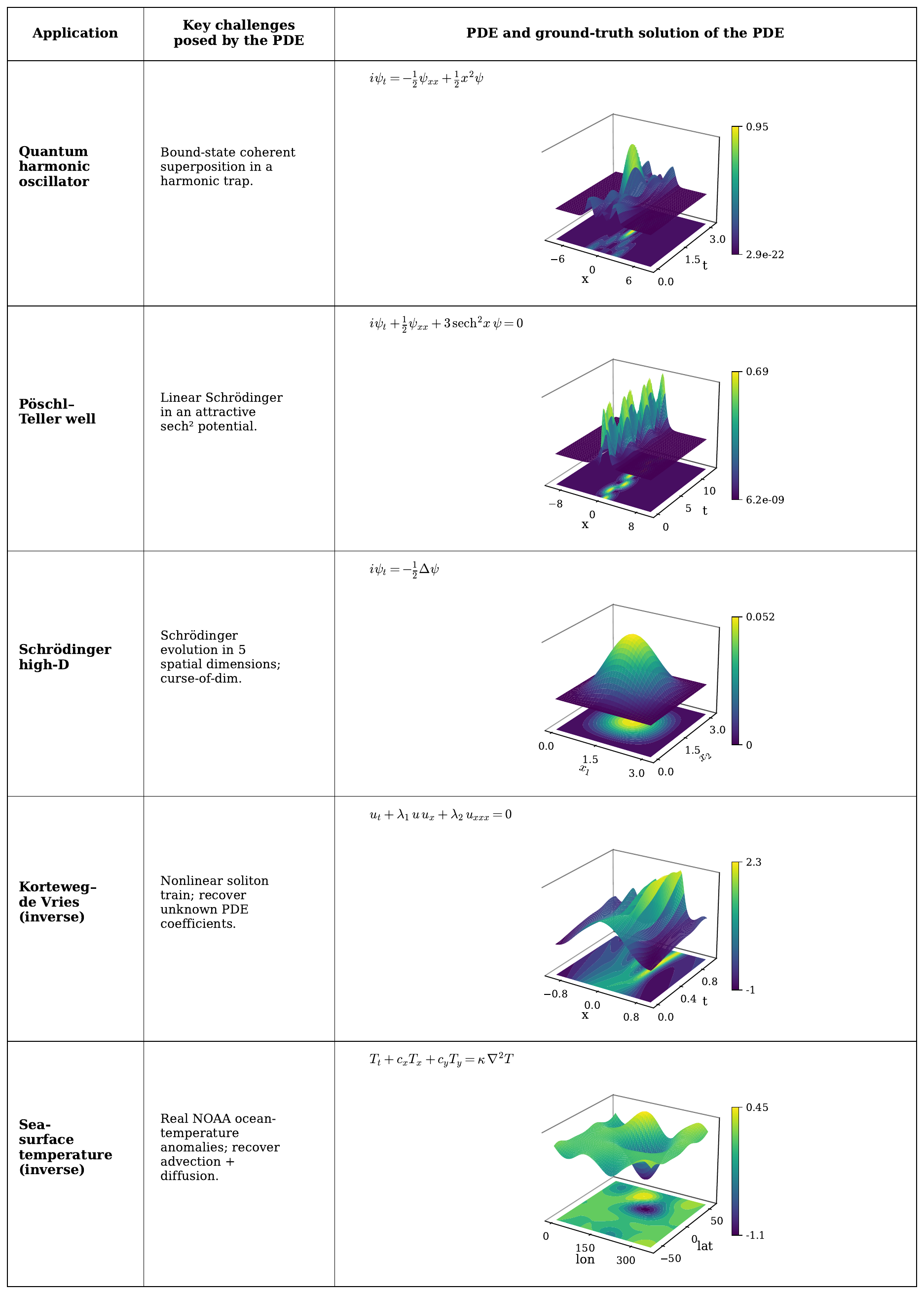}
  \caption{Benchmark overview (inverse, geometry, high-dimensional, problem-adapted-basis).}
  \label{fig:F8b-specialised}
\end{figure}}{}

\begin{table}[ht]
\centering
\caption{Experimental suite: each experiment tests a different aspect of the oscillator-spectral inductive bias.}
\label{tab:experiment-suite}
\begin{tabularx}{\linewidth}{@{}lXX@{}}
\toprule
Experiment & Problems & Purpose \\
\midrule
Exp.~1 & Convection, reaction, wave, beam, Taylor--Green, Schr\"odinger
& Forward accuracy across PDE families \\
Exp.~2 & noEnc, MLPbasis, linDec, noLinOSS, FD2, FP32
& Contribution of each architectural component \\
Exp.~3 & gPINN, RWF, Post-L-BFGS, TimeFD6
& Compatibility with PINN enhancements \\
Exp.~4 & KdV, SST advection--diffusion
& Inverse coefficient recovery \\
Exp.~5 & Hermite and P\"oschl--Teller bases
& Benefit of problem-adapted spectral bases \\
Exp.~6 & Triangle heat equation
& Non-rectangular geometry \\
Exp.~7 & Large-domain Euler--Bernoulli
& Spatial-domain scaling \\
\bottomrule
\end{tabularx}
\end{table}

\IfFileExists{T1_method_attributes.tex}{\input{T1_method_attributes.tex}}{}

\section{Implementation Details}
\label{app:implementation}

\textbf{Hardware and software:}
All experiments were performed on a single NVIDIA RTX~5090 Laptop GPU with
24~GiB VRAM, driver version 590.48.01, and CUDA~12.8. The software environment
used Ubuntu 24.04 LTS with Linux kernel 6.17, PyTorch~2.10.0+cu128, cuDNN
9.10.02, and Python~3.11.6. All runs were executed in a single-process,
single-GPU setting without concurrency.

\textbf{Numerical precision:}
Unless otherwise stated, all experiments are performed in double precision
(\texttt{float64}). A single-precision ablation (\texttt{float32}) is included
in \S\ref{app:ablations} and reported in the $-$FP64 row of
Table~\ref{tab:ablations-full}.

\textbf{Error metrics:}
We use relative mean absolute error (rMAE), relative root mean squared error
(rRMSE), and maximum absolute error (MaxErr). Given a test set $\mathcal{X}$, ground-truth solution
$u_{\mathrm{true}}$, and model prediction $u_{\mathrm{pred}}$, we define
\begin{align}
\mathrm{rMAE}
&=
\frac{
\sum_{x \in \mathcal{X}}
\left|u_{\mathrm{true}}(x)-u_{\mathrm{pred}}(x)\right|
}{
\sum_{x \in \mathcal{X}}
\left|u_{\mathrm{true}}(x)\right|
}, \\
\mathrm{rRMSE}
&=
\frac{
\sqrt{
\sum_{x \in \mathcal{X}}
\left(u_{\mathrm{true}}(x)-u_{\mathrm{pred}}(x)\right)^2
}
}{
\sqrt{
\sum_{x \in \mathcal{X}}
u_{\mathrm{true}}(x)^2
}
}, \\
\mathrm{MaxErr}
&=
\max_{x \in \mathcal{X}}
\left|u_{\mathrm{true}}(x)-u_{\mathrm{pred}}(x)\right|.
\end{align}
For experiments with multiple seeds, we report the mean and standard deviation
over seeds $s\in\{0,1,2\}$; seed-to-seed variability is visualized in
Figure~\ref{fig:seed-stability}. Single-seed experiments are explicitly marked.

\textbf{Optimization:}
All experiments use Adam followed by L-BFGS unless stated otherwise. The Adam
learning rate is $10^{-3}$. L-BFGS uses strong-Wolfe line search, history size
$50$, maximum inner iterations $20$, learning rate $1.0$, gradient tolerance
$10^{-12}$, and change tolerance $10^{-14}$. The default loss weights for
forward problems are $\lambda_{\mathrm{IC}} = 100$, $\lambda_{\mathrm{PDE}} = 1$. For inverse problems, we additionally use
$\lambda_{\mathrm{data}}=100$. Locally adaptive activation functions are
enabled in every coefficient-head MLP. Shared training hyperparameters for OSSM-PINN
and all four baselines are listed in
Tables~\ref{tab:s4-ossm-recipe}--\ref{tab:s8-neusa-recipe}. Per-problem
training-loss histories for OSSM-PINN-IMEX and OSSM-PINN-IM are shown in
Figures~\ref{fig:sup-loss-curves} and~\ref{fig:sup-loss-curves-im}.

% ======================================================================
\section{Forward-Problem Definitions and Additional Results}
\label{app:forward_problems}
% ======================================================================

This section provides the PDE definitions for all forward benchmarks, the full
numerical results table, and additional per-problem field comparisons.

\subsection{PDE Definitions}

\subsubsection{Convection Equation}

We consider the linear advection equation $u_t + \beta u_x = 0$ on
$[0,2\pi]\times[0,1]$ with periodic boundary conditions and
$u(x,0)=\sin x$. The exact solution is $u(x,t)=\sin(x-\beta t)$.
We report results for $\beta=50$ and $\beta=100$.

\subsubsection{Reaction Equation}

We study the logistic reaction equation $u_t = \rho u(1-u)$ on
$[0,2\pi]\times[0,1]$ with periodic boundary conditions and Gaussian IC
$u(x,0) = \exp\!\left(-{(x-\pi)^2}/{(2(\pi/4)^2)}\right)$.
The exact solution is $u(x,t) = h e^{\rho t} / (h(e^{\rho t}-1)+1)$
where $h=u(x,0)$. We use $\rho=5$.

\subsubsection{Wave Equation}

We consider $u_{tt}=4u_{xx}$ on $[0,1]\times[0,1]$ with homogeneous Dirichlet
conditions and IC $u(x,0)=\sin(\pi x)+\frac{1}{2}\sin(\beta\pi x)$,
$u_t(x,0)=0$. The exact solution is
$\sin(\pi x)\cos(2\pi t)+\frac{1}{2}\sin(\beta\pi x)\cos(2\beta\pi t)$
with $\beta=3$.

\subsubsection{Euler--Bernoulli Beam Equation}

We solve the forced beam equation $u_{tt}+u_{xxxx}=(1-16\pi^2)\sin x\cos(4\pi t)$
on $[0,\pi]\times[0,1]$ with simply-supported conditions
$u=u_{xx}=0$ at $x=0,\pi$ and IC $u(x,0)=\sin x$, $u_t(x,0)=0$.
The exact solution is $u(x,t)=\sin x\cos(4\pi t)$.

\subsubsection{Taylor--Green Vortex}

We solve the two-dimensional incompressible Navier--Stokes equations on
$[0,2\pi]^2$ with periodic boundary conditions and $t\in[0,10]$ at
$\mathrm{Re}=100$ ($\nu=10^{-2}$). The IC is $u(x,y,0)=\sin x\cos y$,
$v(x,y,0)=-\cos x\sin y$, $p(x,y,0)=\frac{1}{4}(\cos 2x+\cos 2y)$.

\subsubsection{High-Dimensional Schr\"odinger}

% We solve $i\psi_t = -\frac{1}{2}\Delta\psi + V(x)\psi$ on $[-1,1]^d$ with
% homogeneous Dirichlet boundary conditions for $d=5$ and $d=100$. This
% experiment evaluates scalability and tests the random-sample IC encoder.

We solve the linear Schr\"odinger equation
$i\psi_t = -\tfrac{1}{2}\Delta\psi$ on $[0,\pi]^d$ with
homogeneous Dirichlet boundary conditions $\psi|_{\partial\Omega}=0$
(infinite-well potential, $V\equiv 0$ inside the box) and
time horizon $t\in[0,\pi]$. The eigenstates of $-\tfrac12\Delta$
under these BCs are tensor products of $\sin(k_i x_i)$ with energies
$E_{\mathbf k}=\tfrac12\sum_i k_i^2$, so an IC formed from a few
eigenstates evolves as a closed-form sum of phase-rotated modes.

\noindent\textbf{$d=5$:}
\[
\psi(\mathbf x,0)=\Big(\tfrac{2}{\pi}\Big)^{\!5/2}\!\frac{1}{\sqrt 2}\!
\left[\prod_{i=1}^{5}\sin(x_i)+\prod_{i=1}^{5}\sin(2 x_i)\right],
\]
with exact solution $\psi(\mathbf x,t)=(2/\pi)^{5/2}(1/\sqrt2)
\big[e^{-i\,5t/2}\prod_i\sin x_i + e^{-i\,10\,t}\prod_i\sin 2x_i\big]$
(energies $E=5/2$ and $E=10$).

\noindent\textbf{$d=100$:}
\[
\psi(\mathbf x,0)=\sqrt{\tfrac{2}{\pi}}\,\frac{1}{d}\sum_{i=1}^{d}
\!\big[\sin(x_i)+0.3\,\sin(2 x_i)\big],\qquad d=100,
\]
with exact solution $\psi(\mathbf x,t)=\sqrt{2/\pi}(1/d)\sum_i
\big[e^{-i t/2}\sin x_i + 0.3\,e^{-i 2 t}\sin 2x_i\big]$
(per-axis energies $E_1=1/2$, $E_2=2$).

\subsection{Full Numerical Results}

Table~\ref{tab:s1-full-results} provides the complete per-problem breakdown
with rMAE, rRMSE, and maximum absolute error for all methods.

\begin{table}[ht]
  \caption{Full per-problem numerical results (rMAE, rRMSE, max error). Best in \textcolor{bestC}{\textbf{blue}}, second in \textcolor{secondC}{green}, third in \textcolor{thirdC}{red}. Italic factors show improvement over the best baseline.}
  \label{tab:s1-full-results}
  \centering
  \footnotesize
  \setlength{\tabcolsep}{2pt}
  \renewcommand{\arraystretch}{1.0}
  \resizebox{\textwidth}{!}{%
\begin{tabular}{c l l c c c c c c}
\toprule
& PDE & Metric & PINNsFormer & PINNMamba & ML-PINN & NeuSA & \makecell{OSSM-PINN-IM\\\scriptsize\emph{(ours)}} & \makecell{OSSM-PINN-IMEX\\\scriptsize\emph{(ours)}} \\
\midrule
\multirow{18}{*}{\rotatebox[origin=c]{90}{Forward}} & \multirow{3}{*}{Convection ($\beta{=}50$)} & rMAE & 6.68e-1 & 1.00e+0 & 1.00e+0 & \textcolor{thirdC}{3.24e-3} & \textcolor{secondC}{3.28e-5}\,{\scriptsize\itshape ($\times$99)} & \textbf{\textcolor{bestC}{3.04e-5}}\,{\scriptsize\itshape ($\times$107)} \\
 &  & rRMSE & 7.36e-1 & 1.00e+0 & 1.00e+0 & \textcolor{thirdC}{1.38e-2} & \textcolor{secondC}{3.83e-5}\,{\scriptsize\itshape ($\times$360)} & \textbf{\textcolor{bestC}{3.50e-5}}\,{\scriptsize\itshape ($\times$393)} \\
 &  & max error & 1.03e+0 & 1.05e+0 & 1.02e+0 & \textcolor{thirdC}{1.23e-1} & \textcolor{secondC}{1.05e-4}\,{\scriptsize\itshape ($\times$1.2k)} & \textbf{\textcolor{bestC}{9.64e-5}}\,{\scriptsize\itshape ($\times$1.3k)} \\
\cmidrule(lr){2-9}
 & \multirow{3}{*}{Convection ($\beta{=}100$)} & rMAE & 1.00e+0 & 1.01e+0 & 1.00e+0 & \textcolor{thirdC}{3.24e-3} & \textcolor{secondC}{5.55e-4}\,{\scriptsize\itshape ($\times$6)} & \textbf{\textcolor{bestC}{1.48e-5}}\,{\scriptsize\itshape ($\times$219)} \\
 &  & rRMSE & 1.01e+0 & 1.02e+0 & 1.00e+0 & \textcolor{thirdC}{1.38e-2} & \textcolor{secondC}{6.31e-4}\,{\scriptsize\itshape ($\times$22)} & \textbf{\textcolor{bestC}{1.76e-5}}\,{\scriptsize\itshape ($\times$784)} \\
 &  & max error & 1.14e+0 & 1.20e+0 & 1.04e+0 & \textcolor{thirdC}{1.23e-1} & \textcolor{secondC}{1.56e-3}\,{\scriptsize\itshape ($\times$79)} & \textbf{\textcolor{bestC}{5.26e-5}}\,{\scriptsize\itshape ($\times$2.3k)} \\
\cmidrule(lr){2-9}
 & \multirow{3}{*}{Reaction} & rMAE & \textcolor{thirdC}{1.12e-2} & 7.01e-1 & 9.81e-1 & 2.87e-1 & \textbf{\textcolor{bestC}{2.92e-3}}\,{\scriptsize\itshape ($\times$4)} & \textcolor{secondC}{3.07e-3}\,{\scriptsize\itshape ($\times$4)} \\
 &  & rRMSE & \textcolor{thirdC}{2.90e-2} & 7.56e-1 & 9.84e-1 & 4.49e-1 & \textbf{\textcolor{bestC}{8.75e-3}}\,{\scriptsize\itshape ($\times$3)} & \textcolor{secondC}{9.24e-3}\,{\scriptsize\itshape ($\times$3)} \\
 &  & max error & \textcolor{thirdC}{1.79e-1} & 1.39e+0 & 9.92e-1 & 9.48e-1 & \textbf{\textcolor{bestC}{7.17e-2}}\,{\scriptsize\itshape ($\times$3)} & \textcolor{secondC}{7.99e-2}\,{\scriptsize\itshape ($\times$2)} \\
\cmidrule(lr){2-9}
 & \multirow{3}{*}{Wave} & rMAE & 7.60e-2 & 7.32e-1 & 4.94e-1 & \textcolor{thirdC}{5.48e-3} & \textcolor{secondC}{4.29e-5}\,{\scriptsize\itshape ($\times$128)} & \textbf{\textcolor{bestC}{4.15e-5}}\,{\scriptsize\itshape ($\times$132)} \\
 &  & rRMSE & 7.67e-2 & 7.19e-1 & 4.83e-1 & \textcolor{thirdC}{2.18e-2} & \textcolor{secondC}{4.57e-5}\,{\scriptsize\itshape ($\times$476)} & \textbf{\textcolor{bestC}{4.27e-5}}\,{\scriptsize\itshape ($\times$510)} \\
 &  & max error & \textcolor{thirdC}{1.36e-1} & 1.08e+0 & 6.84e-1 & 1.51e-1 & \textcolor{secondC}{9.71e-5}\,{\scriptsize\itshape ($\times$1.4k)} & \textbf{\textcolor{bestC}{7.49e-5}}\,{\scriptsize\itshape ($\times$1.8k)} \\
\cmidrule(lr){2-9}
 & \multirow{3}{*}{Euler-Bernoulli (classical)} & rMAE & \multirow{3}{*}{\textcolor{stubGray}{\scriptsize\textsc{OOM}}} & \multirow{3}{*}{\textcolor{stubGray}{\scriptsize\textsc{OOM}}} & \textcolor{thirdC}{3.28e-1} & 3.35e+0 & \textbf{\textcolor{bestC}{6.58e-5}}\,{\scriptsize\itshape ($\times$5.0k)} & \textcolor{secondC}{6.72e-5}\,{\scriptsize\itshape ($\times$4.9k)} \\
 &  & rRMSE &  &  & \textcolor{thirdC}{3.32e-1} & 3.41e+0 & \textbf{\textcolor{bestC}{6.73e-5}}\,{\scriptsize\itshape ($\times$4.9k)} & \textcolor{secondC}{6.77e-5}\,{\scriptsize\itshape ($\times$4.9k)} \\
 &  & max error &  &  & \textcolor{thirdC}{4.31e-1} & 4.98e+0 & \textcolor{secondC}{1.14e-4}\,{\scriptsize\itshape ($\times$3.8k)} & \textbf{\textcolor{bestC}{9.79e-5}}\,{\scriptsize\itshape ($\times$4.4k)} \\
\cmidrule(lr){2-9}
 & \multirow{3}{*}{Euler-Bernoulli (extended)} & rMAE & \multirow{3}{*}{\textcolor{stubGray}{\scriptsize\textsc{OOM}}} & \multirow{3}{*}{\textcolor{stubGray}{\scriptsize\textsc{OOM}}} & \textcolor{thirdC}{1.06e+0} & 5.84e+0 & \textbf{\textcolor{bestC}{1.08e-5}}\,{\scriptsize\itshape ($\times$98.4k)} & \textcolor{secondC}{1.49e-5}\,{\scriptsize\itshape ($\times$71.1k)} \\
 &  & rRMSE &  &  & \textcolor{thirdC}{1.07e+0} & 6.25e+0 & \textbf{\textcolor{bestC}{1.39e-5}}\,{\scriptsize\itshape ($\times$76.6k)} & \textcolor{secondC}{1.82e-5}\,{\scriptsize\itshape ($\times$58.6k)} \\
 &  & max error &  &  & \textcolor{thirdC}{1.41e+0} & 1.08e+1 & \textbf{\textcolor{bestC}{3.29e-5}}\,{\scriptsize\itshape ($\times$42.7k)} & \textcolor{secondC}{4.04e-5}\,{\scriptsize\itshape ($\times$34.8k)} \\
\midrule
\midrule
\multirow{12}{*}{\rotatebox[origin=c]{90}{Forward high-dim}} & \multirow{3}{*}{Taylor-Green 2D} & rMAE & \multirow{3}{*}{\textcolor{stubGray}{\scriptsize\textsc{OOM}}} & \multirow{3}{*}{\textcolor{stubGray}{\scriptsize\textsc{OOM}}} & \multirow{3}{*}{\textcolor{stubGray}{\scriptsize\textsc{OOM}}} & \multirow{3}{*}{\textcolor{stubGray}{\scriptsize\textsc{OOM}}} & \textbf{\textcolor{bestC}{2.64e-3}} & \textcolor{secondC}{3.85e-3} \\
 &  & rRMSE &  &  &  &  & \textbf{\textcolor{bestC}{3.14e-3}} & \textcolor{secondC}{5.12e-3} \\
 &  & max error &  &  &  &  & \textbf{\textcolor{bestC}{3.65e-3}} & \textcolor{secondC}{8.46e-3} \\
\cmidrule(lr){2-9}
 & \multirow{3}{*}{Heat (triangular)} & rMAE & \multirow{3}{*}{\textcolor{stubGray}{\scriptsize\textsc{OOM}}} & \multirow{3}{*}{\textcolor{stubGray}{\scriptsize\textsc{OOM}}} & \multirow{3}{*}{\textcolor{stubGray}{\scriptsize\textsc{OOM}}} & \multirow{3}{*}{\textcolor{stubGray}{\scriptsize\textsc{IG}}} & \textbf{\textcolor{bestC}{7.66e-3}} & \textcolor{secondC}{7.67e-3} \\
 &  & rRMSE &  &  &  &  & \textbf{\textcolor{bestC}{7.50e-3}} & \textcolor{secondC}{7.52e-3} \\
 &  & max error &  &  &  &  & \textbf{\textcolor{bestC}{1.13e-2}} & \textcolor{secondC}{1.16e-2} \\
\cmidrule(lr){2-9}
 & \multirow{3}{*}{Schr{\"o}dinger (5D)} & rMAE & \multirow{3}{*}{\textcolor{stubGray}{\scriptsize\textsc{OOM}}} & \multirow{3}{*}{\textcolor{stubGray}{\scriptsize\textsc{OOM}}} & \multirow{3}{*}{\textcolor{stubGray}{\scriptsize\textsc{OOM}}} & \multirow{3}{*}{\textcolor{stubGray}{\scriptsize\textsc{OOM}}} & \textbf{\textcolor{bestC}{1.11e-3}} & \textcolor{secondC}{1.13e-3} \\
 &  & rRMSE &  &  &  &  & \textbf{\textcolor{bestC}{1.24e-3}} & \textcolor{secondC}{1.29e-3} \\
 &  & max error &  &  &  &  & \textbf{\textcolor{bestC}{2.43e-4}} & \textcolor{secondC}{3.75e-4} \\
\cmidrule(lr){2-9}
 & \multirow{3}{*}{Schr{\"o}dinger (100D)} & rMAE & \multirow{3}{*}{\textcolor{stubGray}{\scriptsize\textsc{OOM}}} & \multirow{3}{*}{\textcolor{stubGray}{\scriptsize\textsc{OOM}}} & \multirow{3}{*}{\textcolor{stubGray}{\scriptsize\textsc{OOM}}} & \multirow{3}{*}{\textcolor{stubGray}{\scriptsize\textsc{OOM}}} & \textcolor{secondC}{9.30e-4} & \textbf{\textcolor{bestC}{8.74e-4}} \\
 &  & rRMSE &  &  &  &  & \textcolor{secondC}{1.18e-3} & \textbf{\textcolor{bestC}{1.10e-3}} \\
 &  & max error &  &  &  &  & \textcolor{secondC}{2.48e-3} & \textbf{\textcolor{bestC}{2.21e-3}} \\
\midrule
\midrule
\multirow{6}{*}{\rotatebox[origin=c]{90}{Inverse}} & \multirow{3}{*}{KdV} & rMAE & \multirow{3}{*}{\textcolor{stubGray}{\scriptsize\textsc{OOM}}} & \multirow{3}{*}{\textcolor{stubGray}{\scriptsize\textsc{OOM}}} & \multirow{3}{*}{\textcolor{stubGray}{\scriptsize\textsc{OOM}}} & \multirow{3}{*}{\textcolor{stubGray}{\scriptsize\textsc{ST}}} & \textcolor{secondC}{6.41e-3} & \textbf{\textcolor{bestC}{6.06e-3}} \\
 &  & rRMSE &  &  &  &  & \textcolor{secondC}{6.56e-3} & \textbf{\textcolor{bestC}{6.22e-3}} \\
 &  & max error &  &  &  &  & \textbf{\textcolor{bestC}{1.61e-2}} & \textcolor{secondC}{1.65e-2} \\
\cmidrule(lr){2-9}
 & \multirow{3}{*}{SST 2D adv-diff} & rMAE & \multirow{3}{*}{\textcolor{stubGray}{\scriptsize\textsc{OOM}}} & \multirow{3}{*}{\textcolor{stubGray}{\scriptsize\textsc{OOM}}} & \multirow{3}{*}{\textcolor{stubGray}{\scriptsize\textsc{OOM}}} & \multirow{3}{*}{\textcolor{stubGray}{\scriptsize\textsc{OOM}}} & \textcolor{secondC}{8.06e-2} & \textbf{\textcolor{bestC}{8.05e-2}} \\
 &  & rRMSE &  &  &  &  & \textcolor{secondC}{1.81e-1} & \textbf{\textcolor{bestC}{1.81e-1}} \\
 &  & max error &  &  &  &  & \textcolor{secondC}{2.37e+0} & \textbf{\textcolor{bestC}{2.37e+0}} \\
\bottomrule
\end{tabular}}
  \\[3pt]
  {\scriptsize\textcolor{stubGray}{OOM: out-of-memory; IG: infeasible geometry; ST: stiff PDE.}}
\end{table}

\subsection{Additional Forward-Problem Figures}

\begin{figure}[ht]
  \centering
  \includegraphics[width=\textwidth]{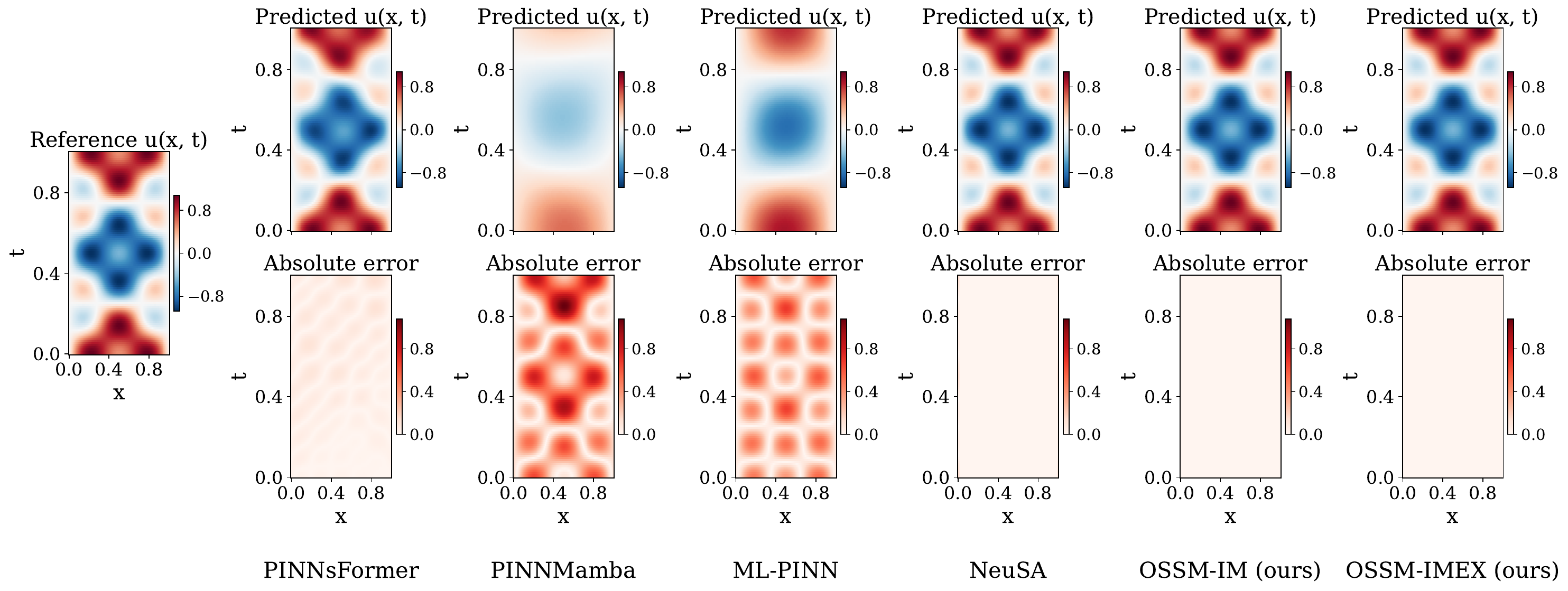}
  \caption{Wave equation: predicted $u(x,t)$ fields and absolute errors for each method.}
  \label{fig:fwd-wave}
\end{figure}

\begin{figure}[ht]
  \centering
  \includegraphics[width=0.8\textwidth]{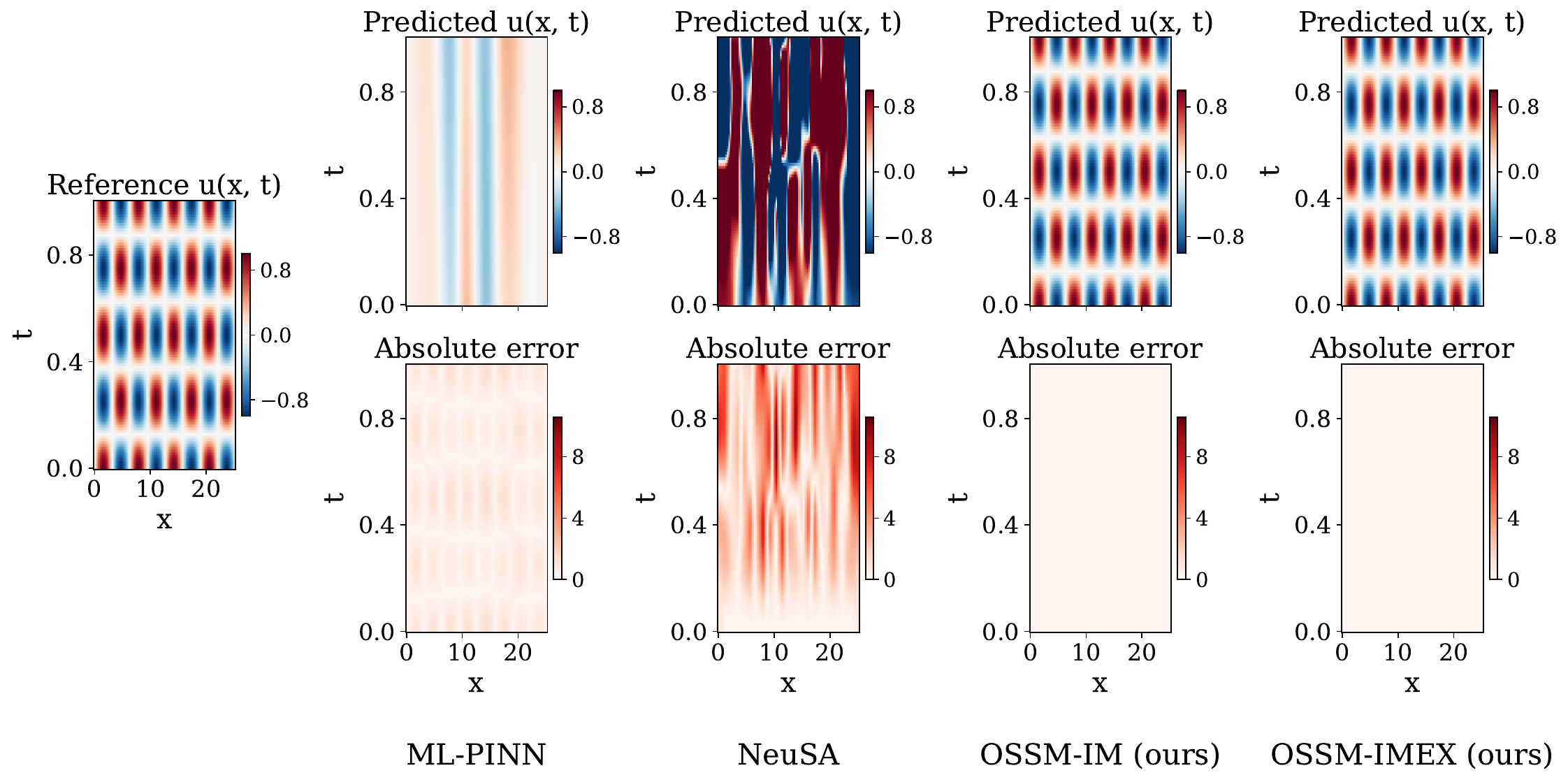}
  \caption{Euler--Bernoulli beam (extended domain $[0,8\pi]$): predicted $u(x,t)$ fields and absolute errors for each method.}
  \label{fig:fwd-eb-extended}
\end{figure}

\begin{figure}[ht]
  \centering
  \includegraphics[width=\textwidth]{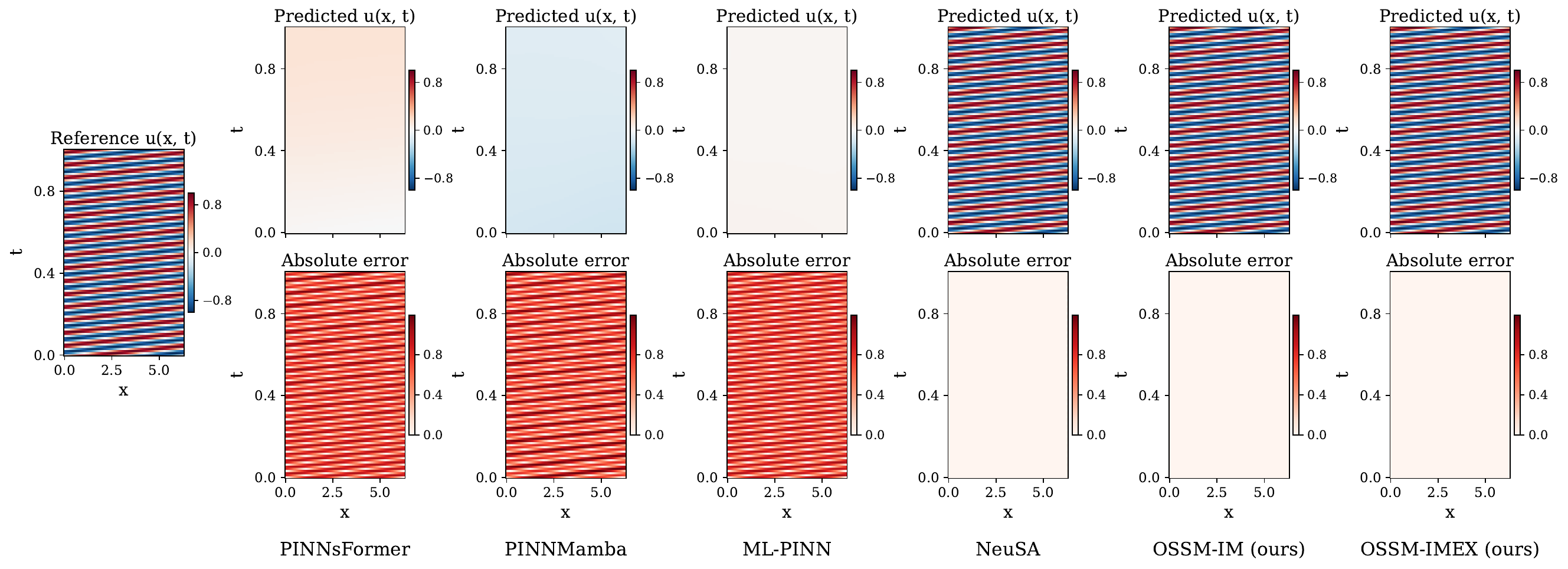}
  \caption{Convection ($\beta = 100$): predicted $u(x,t)$ fields and absolute errors for each method.}
  \label{fig:sup-convection-beta100}
\end{figure}

\begin{figure}[ht]
  \centering
  \includegraphics[width=0.714\textwidth]{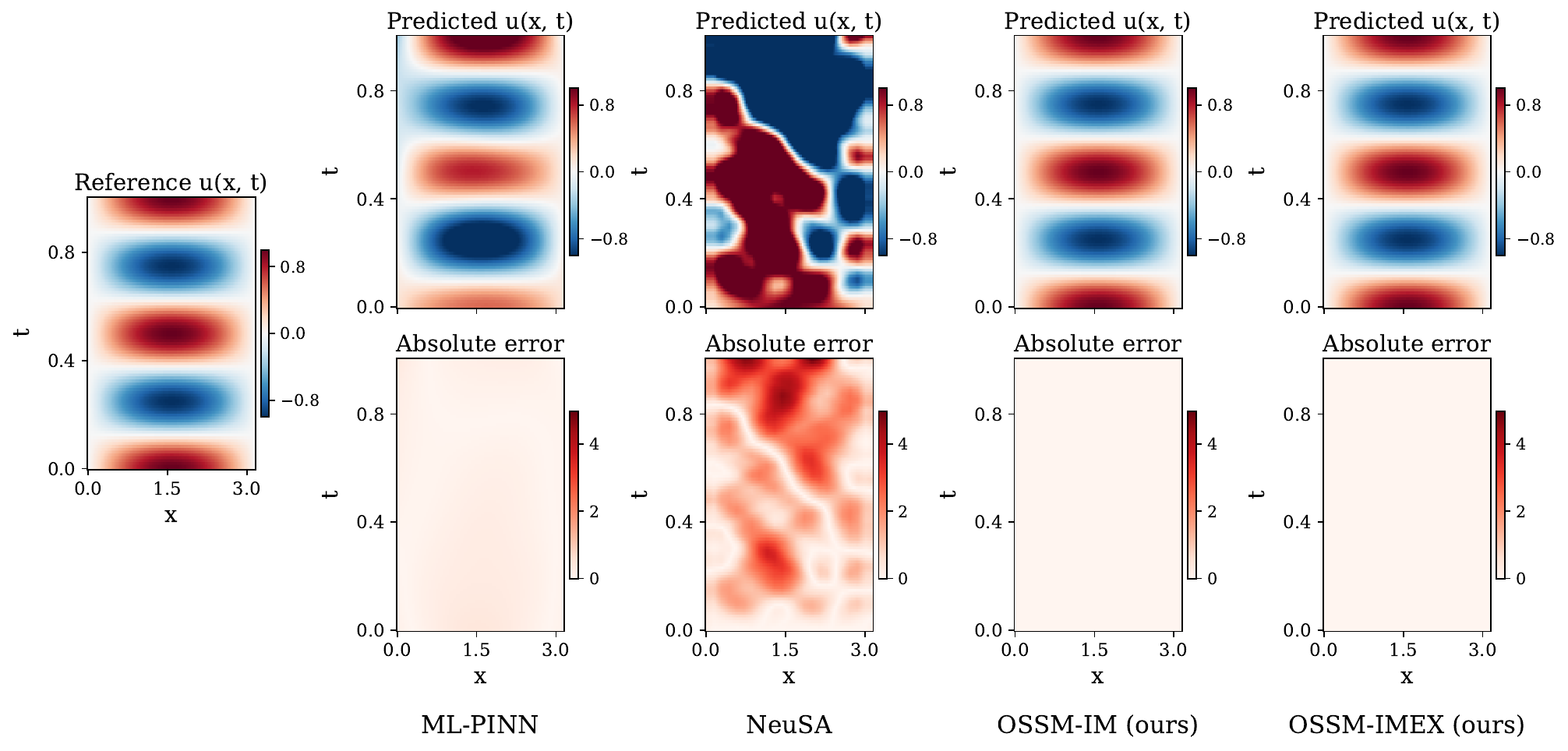}
  \caption{Euler--Bernoulli beam (classical, $[0,2\pi]$): predicted $u(x,t)$ fields and absolute errors for each method.}
  \label{fig:sup-eb}
\end{figure}

Figures~\ref{fig:fwd-wave} and~\ref{fig:fwd-eb-extended} show per-method
predicted fields and absolute errors for the wave and extended Euler--Bernoulli
benchmarks. On the wave equation, NeuSA's spectral rollout reproduces the
dominant mode but accumulates phase error; OSSM-PINN-IMEX is two orders of
magnitude more accurate. On the extended beam, PINNsFormer and PINNMamba
exceed the compute envelope; ML-PINN and NeuSA produce qualitatively
wrong fields, while OSSM-PINN resolves the fourth-order dynamics.
Figure~\ref{fig:sup-convection-beta100} shows the harder convection case
($\beta=100$), and Figure~\ref{fig:sup-eb} shows the classical
Euler--Bernoulli comparison.

% ======================================================================
\section{Spectral and Latent Dynamics}
\label{app:spectral_latent}
% ======================================================================

This section presents frequency-domain analyses, learned dispersion dynamics, and extended latent-state visualizations that probe the mechanism
behind OSSM-PINN's accuracy gains.

\textbf{Frequency-domain analysis:}
In each frequency-domain figure, Figure~\ref{fig:sup-freq-conv50}, ~\ref{fig:sup-freq-conv100}, ~\ref{fig:sup-freq-reaction}, ~\ref{fig:sup-freq-eb}, ~\ref{fig:sup-freq-eb-ext} the left panel shows predictions in physical
space at the final time, and the right panel shows the spatial Fourier spectrum
$|\hat{u}(k, t_{\max})|$ on a log scale. Across all benchmarks, OSSM-PINN
tracks the reference spectrum from the dominant mode down to the FFT noise
floor, while baselines either collapse to near-zero predictions (PINNsFormer,
PINNMamba, ML-PINN on convection) or inject spurious high-frequency content
(NeuSA on Euler--Bernoulli). The key observation is that NeuSA's explicit RK4
rollout introduces aliasing signatures visible as a flat spectral plateau at
high $k$, whereas the LinOSS parallel scan preserves spectral fidelity.

\begin{figure}[ht]
  \centering
  \includegraphics[width=\textwidth]{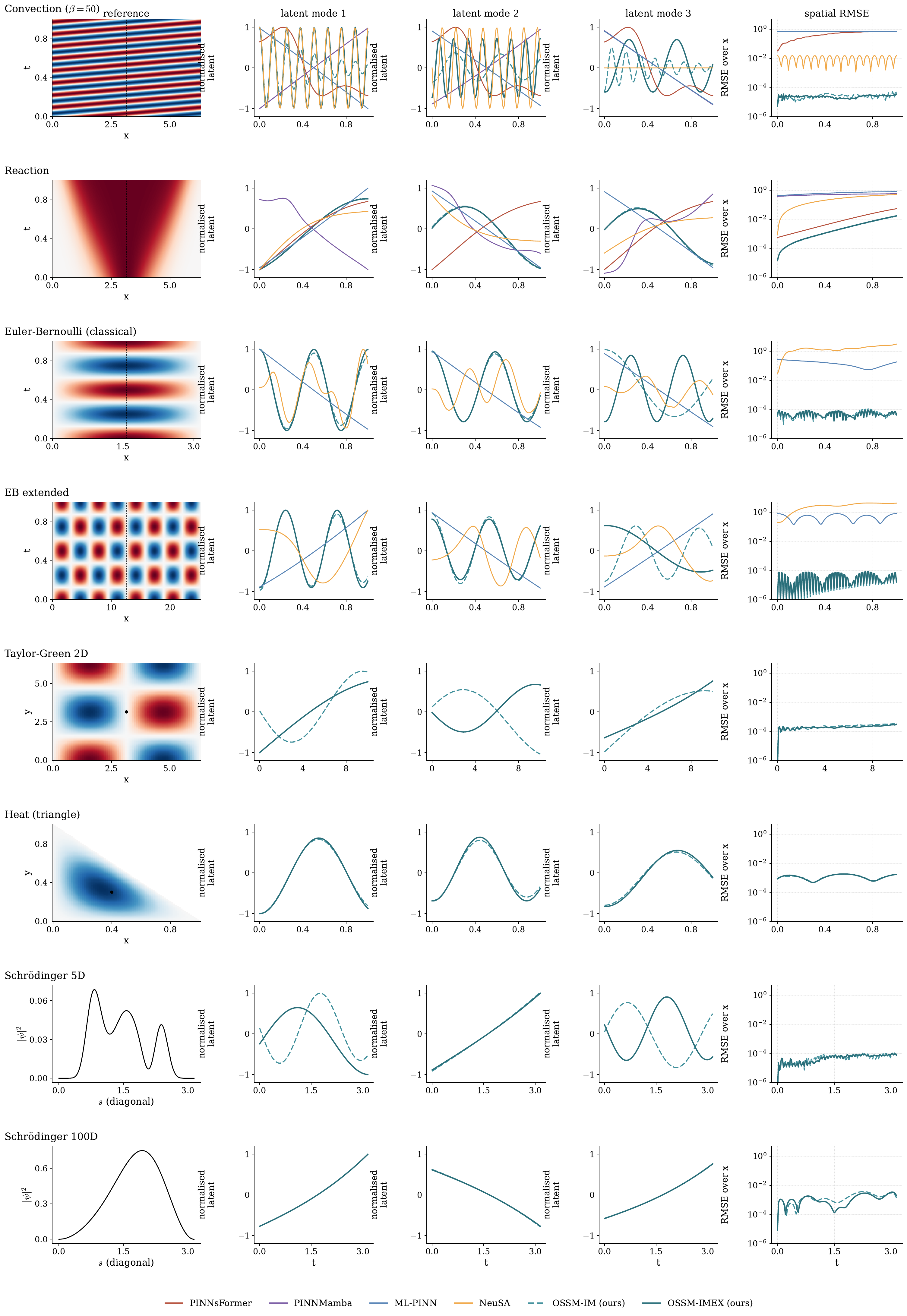}
  \caption{Latent dynamics on all benchmarks in addition to problems shown in Figure~\ref{fig:latent}. Columns: reference field, top-3 latent dimensions per method, spatial RMSE.}
  \label{fig:sup-latents-all}
\end{figure}

\textbf{Latent-state dynamics:}
The latent-dynamics analysis of Figure~\ref{fig:latent} (main text) is extended to all remaining benchmarks in Figure~\ref{fig:sup-latents-all}. The pattern
is consistent: whenever a baseline method runs, its latent decays into a
low-rank drift, while OSSM-PINN's LinOSS populates multiple oscillating
modes, with the spatial-RMSE column confirming the corresponding accuracy gap.

\begin{figure}[ht]
  \centering
  \includegraphics[width=\textwidth]{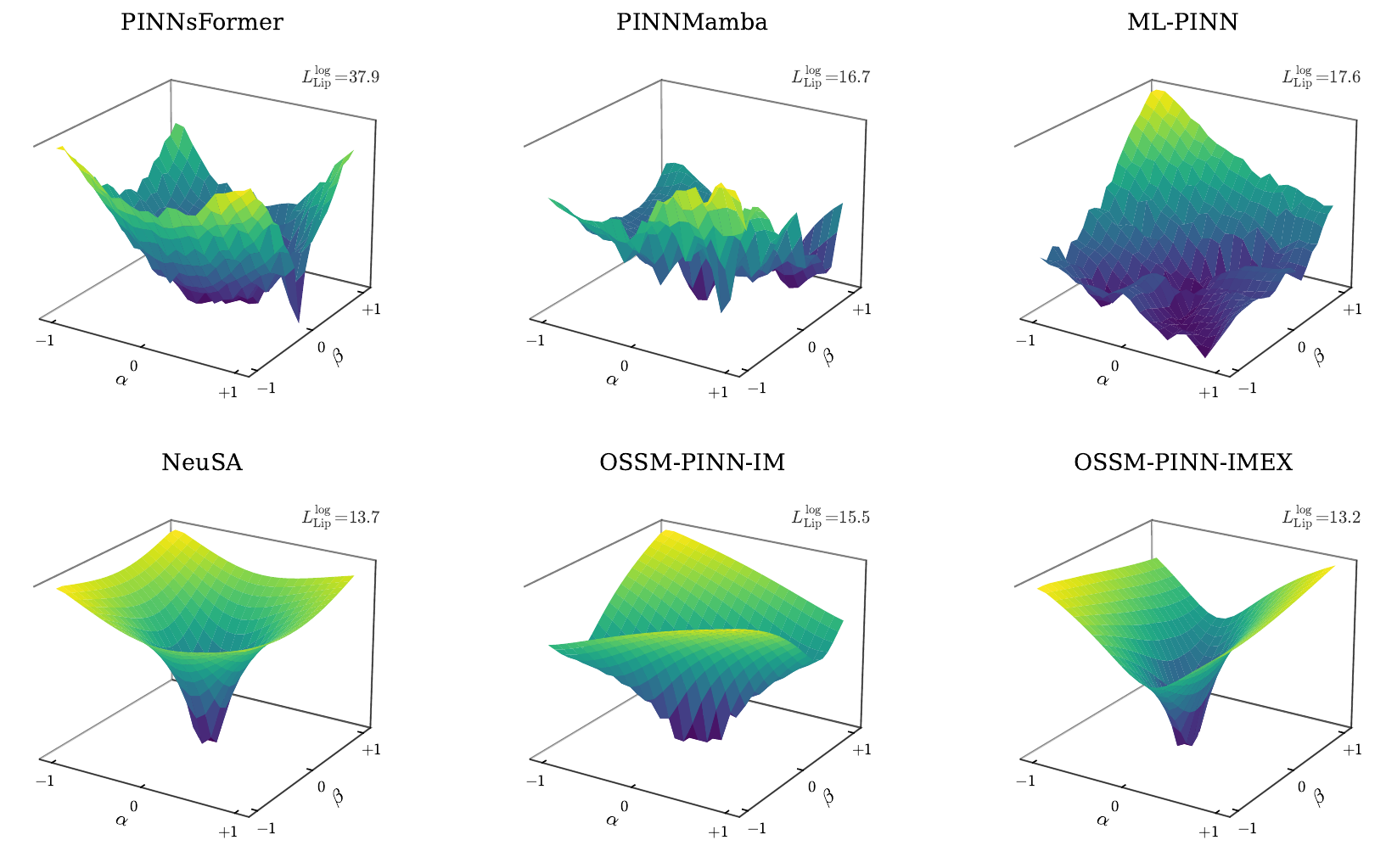}
  \caption{Loss-landscape on convection ($\beta=100$) along top-2 Hessian eigenvectors. Smaller $L_{\mathrm{Lip}}^{\log}$ indicates a smoother basin.}
  \label{fig:sup-landscape-conv100}
\end{figure}

\textbf{Loss landscape:}
The loss-landscape geometry on convection $\beta=100$ is visualized in Figure~\ref{fig:sup-landscape-conv100} using the top-2 Hessian eigenvectors at the trained
solution. OSSM-PINN-IMEX produces the smoothest basin
($L_{\mathrm{Lip}}^{\log}=13.2$), while PINNsFormer's landscape is dramatically
more rugged ($L_{\mathrm{Lip}}^{\log}=37.9$) with multiple visible local minima.
The Lipschitz constant is computed on the filter-normalized, log-scale surface
following the methodology of PINNsFormer \cite{zhao2024pinnsformer}.

\IfFileExists{figs/sup_fig_freq_convection_beta50.pdf}{%
\begin{figure}[ht]
  \centering
  \includegraphics[width=\textwidth]{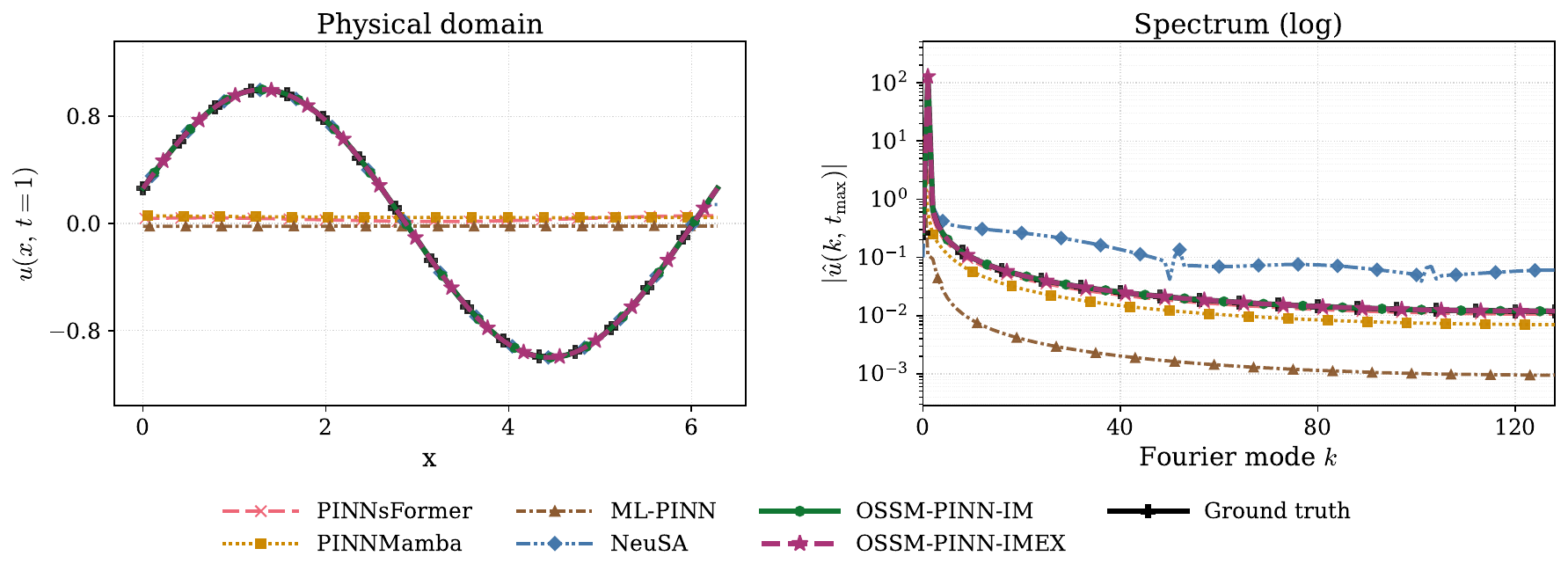}
  \caption{Frequency-domain comparison on convection ($\beta=50$) at $t = 1$: spatial Fourier magnitude spectrum $|\hat{u}(k)|$ of predicted vs.\ reference fields.}
  \label{fig:sup-freq-conv50}
\end{figure}}{}

\IfFileExists{figs/sup_fig_freq_convection_beta100.pdf}{%
\begin{figure}[ht]
  \centering
  \includegraphics[width=\textwidth]{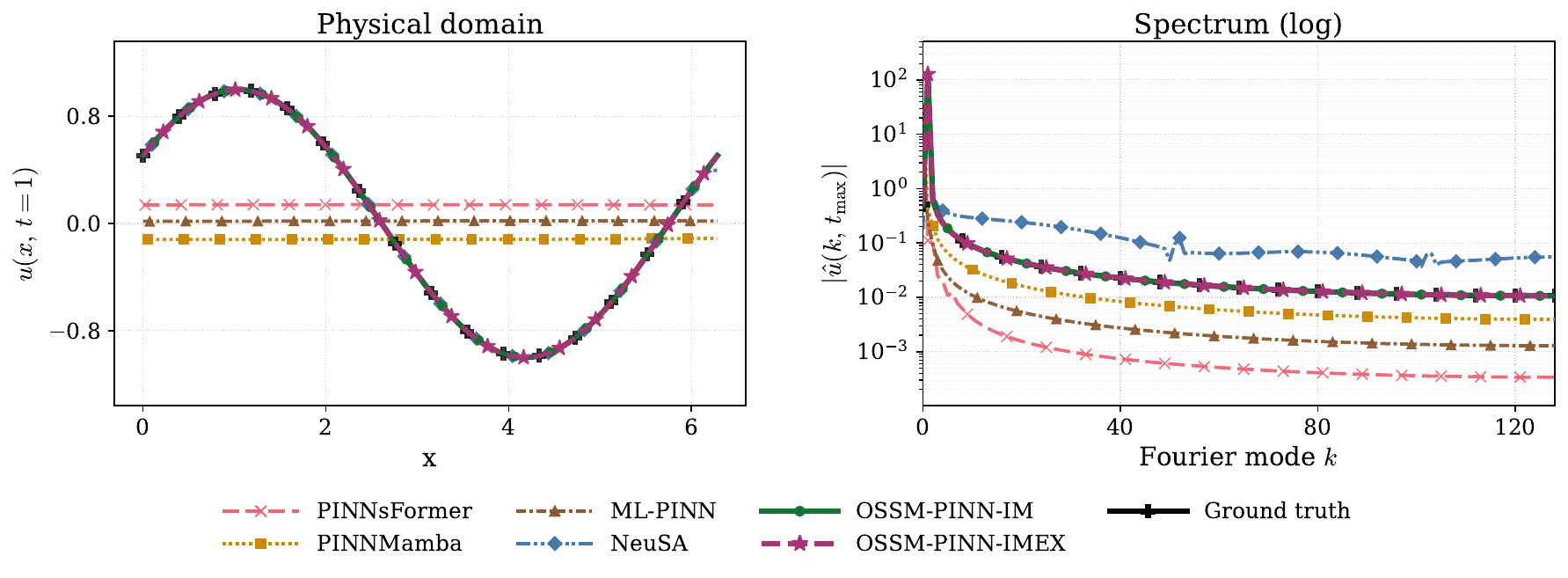}
  \caption{Frequency-domain comparison on convection ($\beta=100$) at $t = 1$: spatial Fourier magnitude spectrum $|\hat{u}(k)|$ of predicted vs.\ reference fields.}
  \label{fig:sup-freq-conv100}
\end{figure}}{}

\IfFileExists{figs/sup_fig_freq_reaction_rho5.pdf}{%
\begin{figure}[ht]
  \centering
  \includegraphics[width=0.75\textwidth]{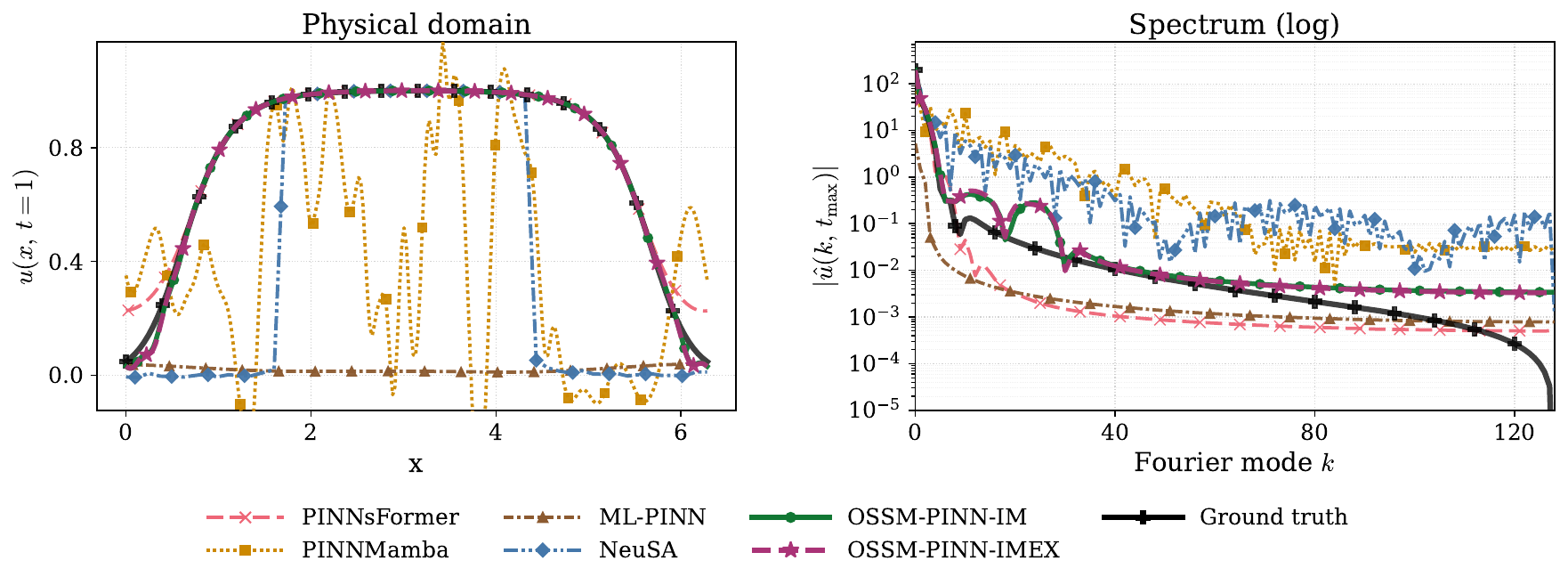}
  \caption{Frequency-domain comparison on reaction at the final time: spatial Fourier magnitude spectrum of predicted vs.\ reference fields.}
  \label{fig:sup-freq-reaction}
\end{figure}}{}

\IfFileExists{figs/sup_fig_freq_euler_bernoulli.pdf}{%
\begin{figure}[ht]
  \centering
  \includegraphics[width=\textwidth]{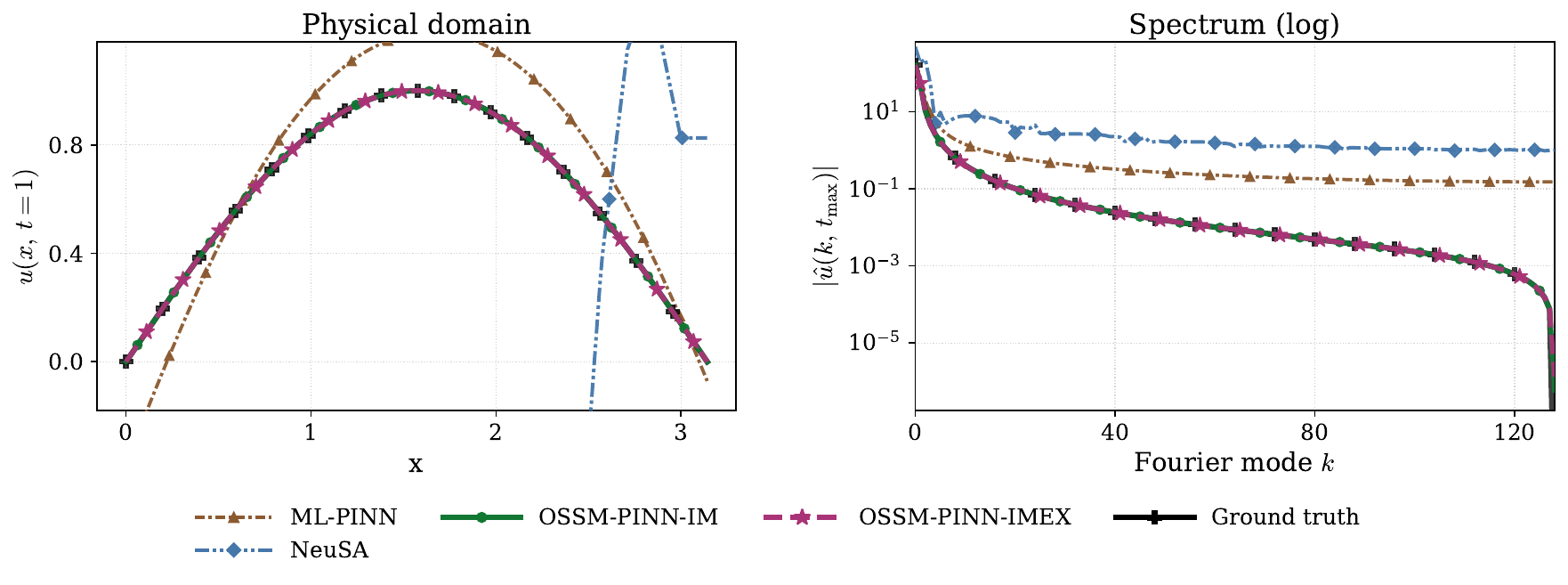}
  \caption{Frequency-domain comparison on Euler--Bernoulli beam (classical, $[0,2\pi]$) at the final time: spatial Fourier magnitude spectrum of predicted vs.\ reference fields.}
  \label{fig:sup-freq-eb}
\end{figure}}{}

\IfFileExists{figs/sup_fig_freq_euler_bernoulli_extended.pdf}{%
\begin{figure}[ht]
  \centering
  \includegraphics[width=\textwidth]{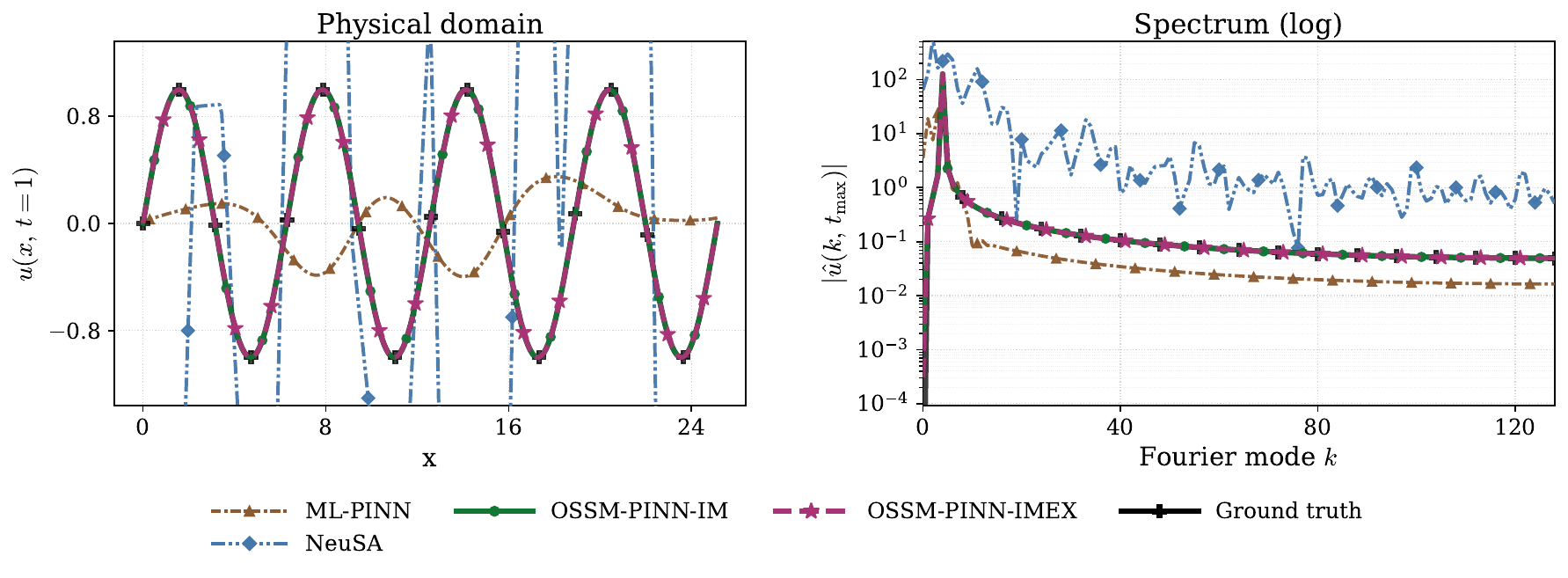}
  \caption{Frequency-domain comparison on extended Euler--Bernoulli at the final time: spatial Fourier magnitude spectrum of predicted vs.\ reference fields.}
  \label{fig:sup-freq-eb-ext}
\end{figure}}{}

\section{Ablation Studies}
\label{app:ablations}

\subsection{Architecture Ablations}
\label{app:ablations:arch}

We perform ablation studies to isolate the contribution of each component
(Exp.~2 in Table~\ref{tab:experiment-suite}). Each ablation removes or alters
one component at a time:
\begin{itemize}
    \item \textbf{$-$ IC encoder}: removes the initial-condition encoder; the initial
    oscillator state is fixed at zero.
    \item \textbf{$-$ Fourier basis}: replaces the Fourier basis with a learned MLP
    basis and adds a soft boundary-condition penalty.
    \item \textbf{$-$ MLP decoder}: replaces the spectral decoder with a single linear
    projection.
    \item \textbf{$-$ LinOSS}: removes the LinOSS rollout and uses a learnable
    constant latent state.
    \item \textbf{$-$ 6th-order FD}: replaces the sixth-order finite-difference stencil with
    a second-order stencil.
    \item \textbf{$-$ FP64}: switches precision from \texttt{float64} to
    \texttt{float32}.
\end{itemize}

Table~\ref{tab:ablations-full} reports results for both the IM and IMEX
discretizations on three benchmarks.

\begin{table}[ht]
  \caption{Architecture ablations (rMAE) on three benchmarks for both IM and IMEX discretizations.}
  \label{tab:ablations-full}
  \centering
  \footnotesize
  \setlength{\tabcolsep}{4pt}
  \begin{tabular}{l c c c c c c}
  \toprule
  Variant & \multicolumn{2}{c}{Conv.\ $\beta=50$} & \multicolumn{2}{c}{Euler-Bernoulli} & \multicolumn{2}{c}{Wave} \\
  \cmidrule(lr){2-3} \cmidrule(lr){4-5} \cmidrule(lr){6-7}
   & IM & IMEX & IM & IMEX & IM & IMEX \\
  \midrule
  Default & 3.28e-5 & 3.04e-5 & 6.58e-5 & 6.72e-5 & 4.29e-5 & 4.15e-5 \\
  $-$ IC encoder & 9.70e-5 & 5.37e-5 & 3.50e-2 & 3.84e-2 & 7.38e-2 & 7.30e-2 \\
  $-$ MLP decoder & 1.81e-4 & 1.27e-4 & 1.03e-4 & 4.34e-5 & 4.35e-5 & 6.79e-5 \\
  $-$ Fourier basis & 7.75e-1 & 7.10e-1 & 1.34e-2 & 9.95e-3 & 1.00e+0 & 1.00e+0 \\
  $-$ 6th-order FD & 2.65e-2 & 2.75e-2 & 1.24e-4 & 1.24e-4 & 3.92e-4 & 3.99e-4 \\
  $-$ FP64 & 2.35e-3 & 7.97e-4 & 3.03e-3 & 2.38e-3 & 1.15e-2 & 8.30e-3 \\
  $-$ LinOSS & 1.00e+0 & 1.00e+0 & 1.00e+0 & 1.00e+0 & 1.00e+0 & 1.00e+0 \\
  \bottomrule
  \end{tabular}
\end{table}

\subsection{State-Space-Cell Swap Ablation}
\label{app:ablations:cell}

To isolate the effect of the oscillatory eigenstructure, we hold the OSSM-PINN
backbone fixed (encoder, modewise decoder, Fourier basis, Adam $\to$ L-BFGS
recipe, same hidden size and parameter budget) and swap only the temporal cell.
Tables~\ref{tab:ssm-ablation-conv} and~\ref{tab:ssm-ablation-reac} report
results on convection and reaction respectively.

On convection ($\beta=50$), LinOSS is the only cell that learns the
high-frequency traveling wave; every non-oscillatory cell plateaus at rMAE
near unity (LRU \cite{orvieto2023resurrecting}/Mamba \cite{gu2024mamba}/NCDE \cite{kidger2020neural}/NRDE \cite{morrill2021neural}/Log-NCDE \cite{walker2024log}) or two orders of magnitude above
LinOSS (S5). On reaction, where the dynamics are monotonic and non-oscillatory,
LinOSS still leads the closest non-oscillatory cell (Mamba) by an order of
magnitude.

\begin{table}[ht]
\centering
\small
\caption{State-space-cell ablation on convection ($\beta{=}50$). Same backbone; only the temporal cell changes.}
\label{tab:ssm-ablation-conv}
\begin{tabular}{l r r r}
\toprule
Cell & rMAE $\downarrow$ & rRMSE $\downarrow$ & \#params \\
\midrule
\textbf{LinOSS-IMEX (ours)} & 3.039e-5 & 3.505e-5 & 93,459 \\
LRU      & 0.887 & 0.920 & 93,971 \\
S5       & 0.008 & 0.009 & 93,971 \\
Mamba    & 0.891 & 0.924 & 94,227 \\
NCDE     & 0.992 & 0.998 & 130,451 \\
NRDE     & 0.993 & 0.997 & 114,195 \\
Log-NCDE & 0.859 & 0.900 & 114,195 \\
\bottomrule
\end{tabular}
\end{table}

\begin{table}[ht]
\centering
\small
\caption{State-space-cell ablation on reaction. Same backbone as Table~\ref{tab:ssm-ablation-conv}.}
\label{tab:ssm-ablation-reac}
\begin{tabular}{l r r r}
\toprule
Cell & rMAE $\downarrow$ & rRMSE $\downarrow$ & \#params \\
\midrule
\textbf{LinOSS-IMEX (ours)} & 0.003 & 0.009 & 116,164 \\
LRU      & 0.136 & 0.262 & 116,676 \\
S5       & 0.095 & 0.200 & 116,676 \\
Mamba    & 0.072 & 0.161 & 116,932 \\
NCDE     & 0.109 & 0.222 & 153,156 \\
NRDE     & 0.356 & 0.495 & 136,900 \\
Log-NCDE & 0.340 & 0.488 & 136,900 \\
\bottomrule
\end{tabular}
\end{table}

\subsection{Plug-in Compatibility}
\label{app:plugin_compatibility}

OSSM-PINN is an architectural change compatible with existing PINN training
techniques (Exp.~3 in Table~\ref{tab:experiment-suite}). We test four
enhancements independently:
\begin{itemize}
    \item \textbf{gPINN}: adds a penalty on additional spatial gradient of the PDE
    residual, $\lambda_g\|\nabla_x\mathcal{F}(u)\|^2$.
    \item \textbf{Random weight factorization (RWF)}: factorizes each linear layer as
    $W=\operatorname{diag}(\exp s)V$.
    \item \textbf{Post-L-BFGS}: applies an additional L-BFGS refinement from a
    trained checkpoint.
    \item \textbf{TimeFD6}: enforces sixth-order finite differences for temporal
    derivatives throughout training.
\end{itemize}

Table~\ref{tab:additions-full} reports results for both IM and IMEX on
convection, wave, and Euler--Bernoulli.

\begin{table}[ht]
  \caption{Plug-in extensions (rMAE, IM and IMEX). Each row adds one modification on top of the default backbone.}
  \label{tab:additions-full}
  \centering
  \footnotesize
  \setlength{\tabcolsep}{4pt}
  \begin{tabular}{l c c c c c c}
  \toprule
  Variant & \multicolumn{2}{c}{Conv.\ $\beta=50$} & \multicolumn{2}{c}{Euler-Bernoulli} & \multicolumn{2}{c}{Wave} \\
  \cmidrule(lr){2-3} \cmidrule(lr){4-5} \cmidrule(lr){6-7}
   & IM & IMEX & IM & IMEX & IM & IMEX \\
  \midrule
  Default & 3.28e-5 & 3.04e-5 & 6.58e-5 & 6.72e-5 & 4.29e-5 & 4.15e-5 \\
  $+$ gPINN & 3.33e-5 & 2.74e-5 & 7.79e-5 & 5.28e-5 & 4.96e-5 & 6.75e-5 \\
  $+$ RWF & 3.29e-5 & 3.00e-5 & 4.41e-5 & 2.73e-5 & 4.51e-5 & 1.60e-5 \\
  $+$ post-L-BFGS & 2.78e-5 & 3.04e-5 & 5.19e-5 & 4.74e-5 & 2.74e-5 & 3.75e-5 \\
  $+$ time FD-6 & 3.28e-5 & 3.04e-5 & 6.46e-5 & 7.83e-5 & 5.06e-5 & 2.91e-5 \\
  \bottomrule
  \end{tabular}
\end{table}

\subsection{Hyperparameter Sensitivity}
\label{app:sensitivity}

Table~\ref{tab:sensitivity} sweeps $K\in\{8,16,32,64\}$ and
$N\in\{100,200,400,800\}$ on the 1D reaction problem. All 32 runs converge
without failure; sensitivity to the temporal rollout depth $N$ is essentially
flat (rMAE varies less than $10\%$ across the $8\times$ range in $N$), while
mild dependence on $K$ reflects the limited spatial frequency content of the
reaction equation.

\begin{table}[ht]
  \centering
  \footnotesize
  \setlength{\tabcolsep}{5pt}
  \caption{Hyperparameter sensitivity on 1D reaction.}
  \label{tab:sensitivity}
  \begin{tabular}{l c c c c c}
  \toprule
  Method & Metric & $K=8$ & $K=16$ & $K=32$ & $K=64$ \\
  \midrule
  \multirow{2}{*}{OSSM-PINN-IM}
    & rMAE  & 3.22--3.35e-3 & 2.68--4.69e-3 & 1.55--1.94e-2 & 2.04--2.36e-2 \\
    & rRMSE & 6.69--7.03e-3 & 9.22e-3--1.35e-2 & 4.48--5.52e-2 & 5.57--6.44e-2 \\
  \midrule
  \multirow{2}{*}{OSSM-PINN-IMEX}
    & rMAE  & 3.22--3.42e-3 & 2.54--2.82e-3 & 1.49--1.66e-2 & 1.90--2.03e-2 \\
    & rRMSE & 6.72--7.28e-3 & 8.80--9.21e-3 & 4.29--4.75e-2 & 5.48--5.63e-2 \\
  \bottomrule
  \multicolumn{6}{l}{\scriptsize Ranges are over $N\in\{100,200,400,800\}$; variation across rollout depth is $<10\%$ in all cases.}
  \end{tabular}
\end{table}

% ======================================================================
\section{Inverse Problems}
\label{app:inverse_problems}
% ======================================================================

\subsection{KdV Inverse Problem}

\begin{figure}[ht]
  \centering
  \includegraphics[width=0.55\textwidth]{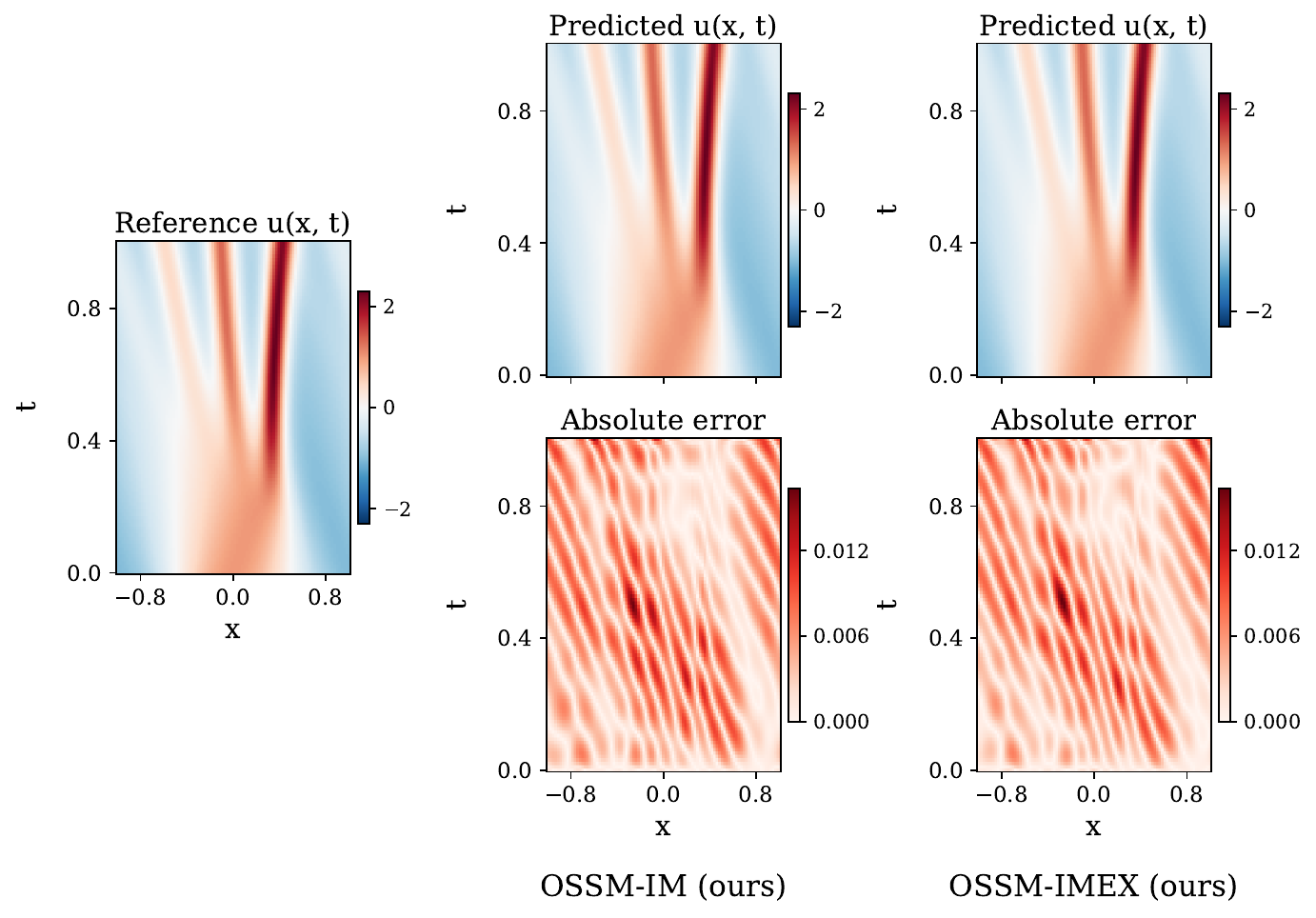}
  \caption{KdV inverse problem: predicted $u(x,t)$ fields and absolute errors. OSSM-PINN recovers both coefficients ($\lambda_1$, $\lambda_2$) to $<0.1\%$ relative error.}
  \label{fig:sup-inv-kdv}
\end{figure}

\begin{figure}[ht]
  \centering
  \includegraphics[width=0.78\textwidth]{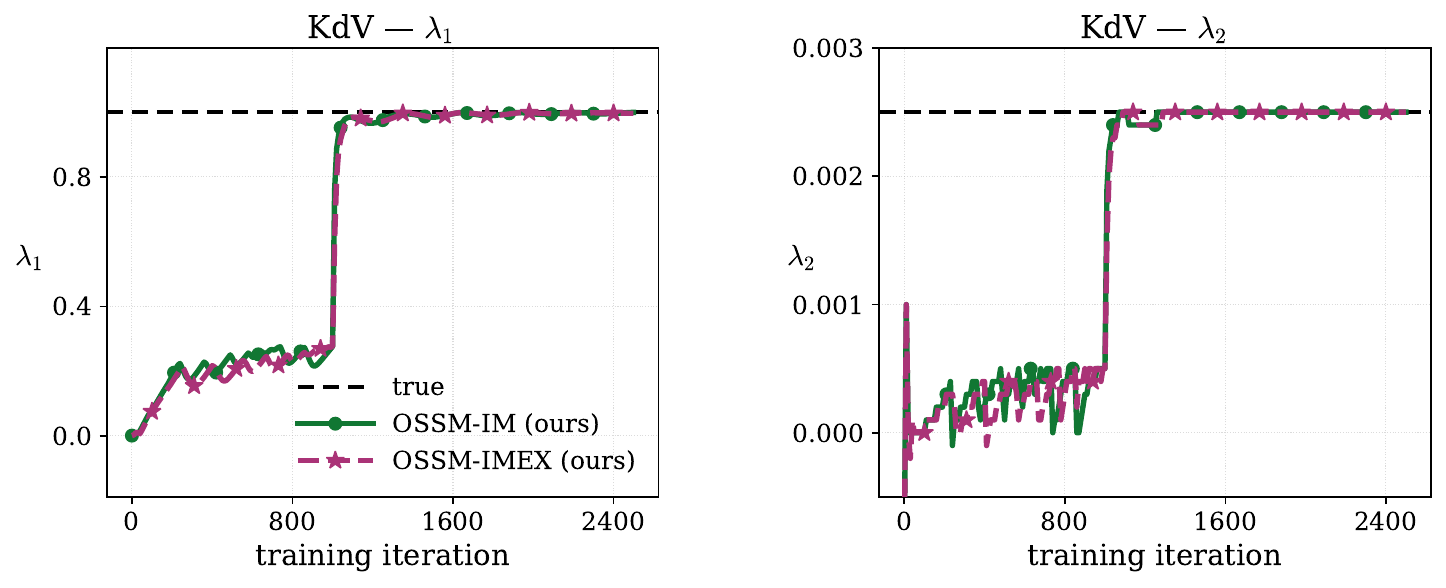}
  \caption{KdV coefficient convergence during training. Dashed line: true parameter value.}
  \label{fig:sup-inv-kdv-evolution}
\end{figure}

We consider $u_t+\lambda_1 u u_x+\lambda_2 u_{xxx}=0$ on $[-1,1]\times[0,1]$
with periodic boundary conditions and $u(x,0)=\cos(\pi x)$. True parameters:
$\lambda_1=1$, $\lambda_2=0.0025$. We use $1000$ randomly sampled space-time
observations. Recovered values and relative errors are listed in
Table~\ref{tab:s3-inverse-recovery}. Figure~\ref{fig:sup-inv-kdv} shows the
predicted fields and Figure~\ref{fig:sup-inv-kdv-evolution} shows the
coefficient convergence trajectories. Figure~\ref{fig:sup-freq-kdv} confirms
that the inferred forward solution matches the reference spectrum.

\subsection{Sea-Surface-Temperature Inverse Problem}

\begin{figure}[ht]
  \centering
  \includegraphics[width=\textwidth]{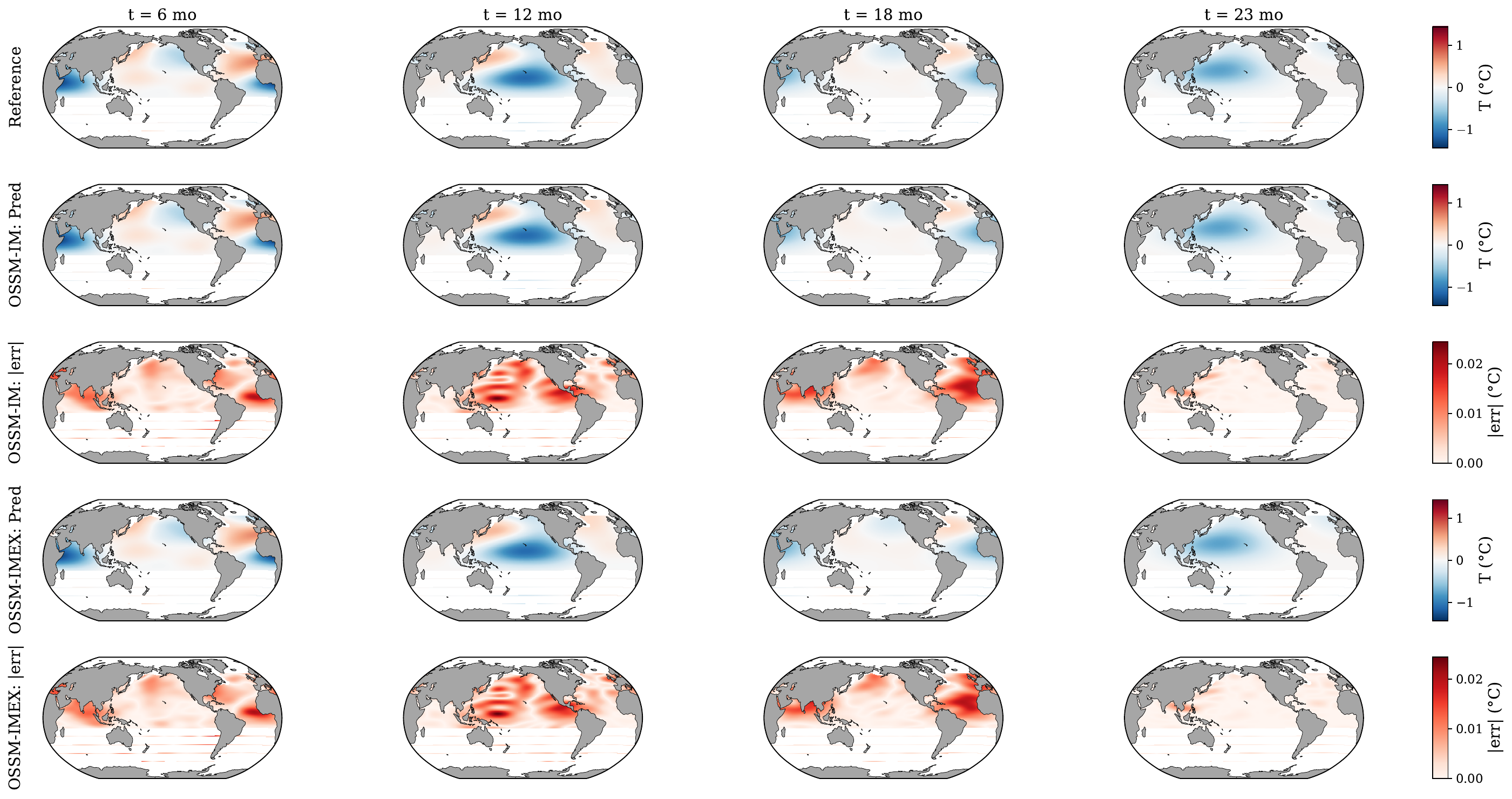}
  \caption{SST 2D advection--diffusion inverse problem: predicted temperature $T(x,y,t)$ at four monthly snapshots ($t\in\{6,12,18,23\}$\,mo). Rows: reference, OSSM-IM prediction, OSSM-IMEX prediction, and absolute errors.}
  \label{fig:sup-inv-sst}
\end{figure}

\begin{figure}[ht]
  \centering
  \includegraphics[width=\textwidth]{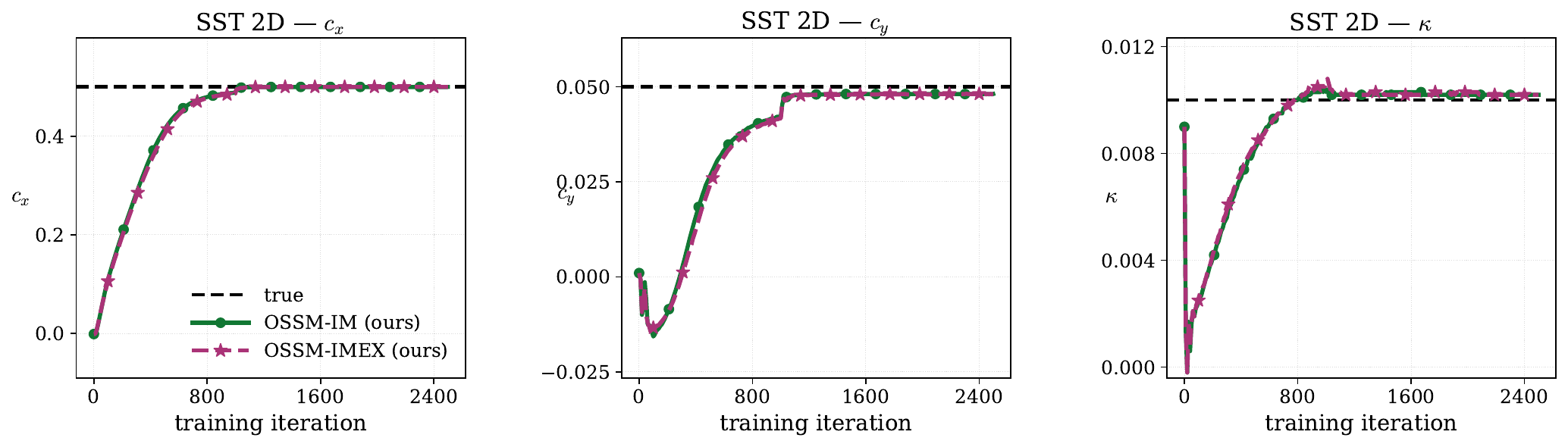}
  \caption{SST coefficient convergence during training ($c_x$, $c_y$, $\kappa$).}
  \label{fig:sup-inv-sst-evolution}
\end{figure}

We consider $\mathrm{u}_\mathrm{t}+c_\mathrm{x}\mathrm{u}_\mathrm{x}+c_\mathrm{y}\mathrm{u}_\mathrm{y}=\kappa(\mathrm{u}_{\mathrm{xx}}+\mathrm{u}_{\mathrm{yy}})$ on $[0,2\pi]^2$
with periodic boundary conditions and $t\in[0,23]$ months. The IC is a NOAA
OISST~\citep{huang2021oisst} anomaly field from January 2008. As the data does not specify the exact PDE that governs it. Advection-diffusion equation is considered with true parameters: $c_x=0.5$,
$c_y=0.05$, $\kappa=0.01$. Using true values the solution is found by a traditional method. We use $5000$ observations as the data for inverse problem.
Figure~\ref{fig:sup-inv-sst} shows predicted fields and
Figure~\ref{fig:sup-inv-sst-evolution} shows coefficient convergence.

\begin{table}[ht]
  \caption{Inverse parameter recovery: Ground-truth vs.\ OSSM-PINN-recovered coefficients.}
  \label{tab:s3-inverse-recovery}
  \centering
  \footnotesize
  \begin{tabular}{l l c c c c}
  \toprule
  Problem & Method & Parameter & True & Recovered & Rel.\ err.\ \\
  \midrule
  KdV (Raissi) & OSSM-IM (ours) & $\lambda_1$ & 1 & 0.99932 & 0.07\% \\
   &  & $\lambda_2$ & 0.0025 & 0.0024986 & 0.06\% \\
  \cmidrule(lr){3-6}
  KdV (Raissi) & OSSM-IMEX (ours) & $\lambda_1$ & 1 & 0.99656 & 0.34\% \\
   &  & $\lambda_2$ & 0.0025 & 0.0024935 & 0.26\% \\
  \cmidrule(lr){3-6}
  SST 2D & OSSM-IM (ours) & $c_x$ & 0.5 & 0.49963 & 0.07\% \\
   &  & $c_y$ & 0.05 & 0.048074 & 3.85\% \\
   &  & $\kappa$ & 0.01 & 0.010235 & 2.35\% \\
  \cmidrule(lr){3-6}
  SST 2D & OSSM-IMEX (ours) & $c_x$ & 0.5 & 0.49971 & 0.06\% \\
   &  & $c_y$ & 0.05 & 0.048107 & 3.79\% \\
   &  & $\kappa$ & 0.01 & 0.010215 & 2.15\% \\
  \bottomrule
  \end{tabular}
\end{table}

\IfFileExists{figs/sup_fig_freq_kdv_inverse.pdf}{%
\begin{figure}[ht]
  \centering
  \includegraphics[width=\textwidth]{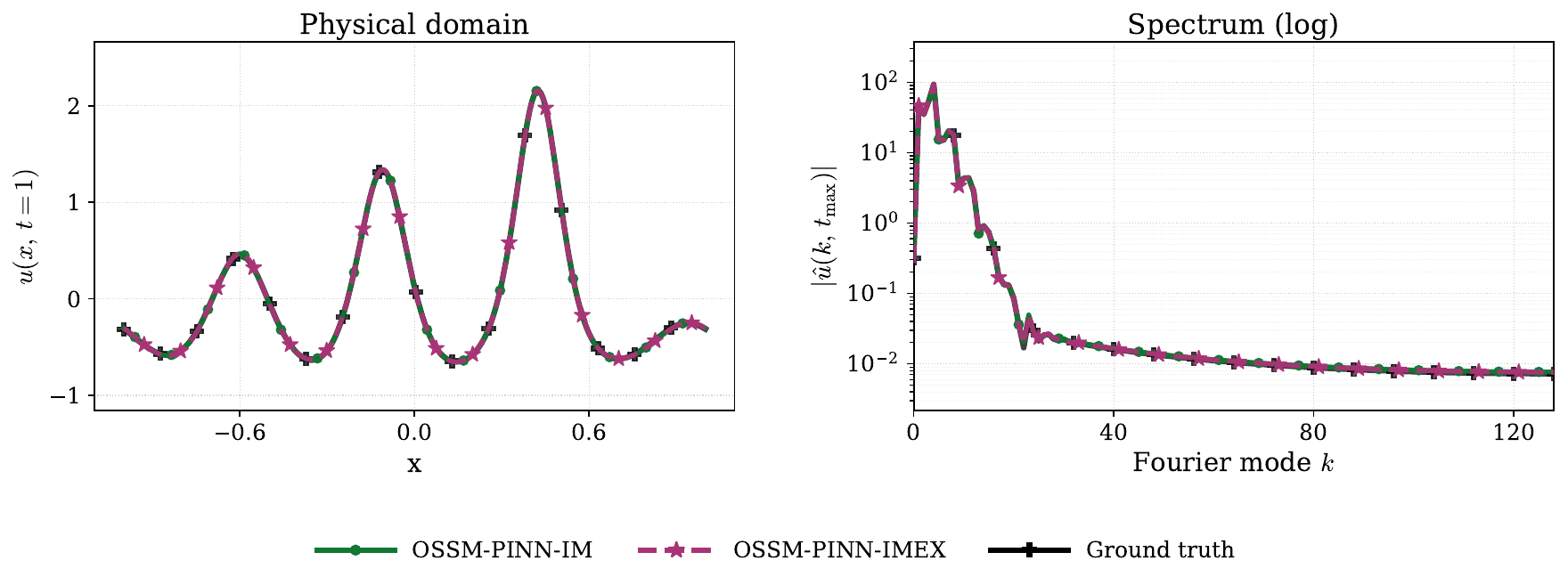}
  \caption{Frequency-domain validation of KdV inverse recovery at final time: spatial Fourier spectrum of the forward solution predicted with recovered coefficients matches the reference down to FFT noise floor.}
  \label{fig:sup-freq-kdv}
\end{figure}}{}

\section{Problem-Adapted Spectral Bases}
\label{app:adapted_basis}

We evaluate whether matching the spectral basis to the PDE's eigenstructure
improves accuracy (Exp.~5 in Table~\ref{tab:experiment-suite}). The analytical
basis functions are illustrated in Figure~\ref{fig:F7-bases}; quantitative
results are in Table~\ref{tab:basis-full}.

\subsubsection{Quantum Harmonic Oscillator}
\begin{figure}[ht]
  \centering
  \includegraphics[width=\textwidth]{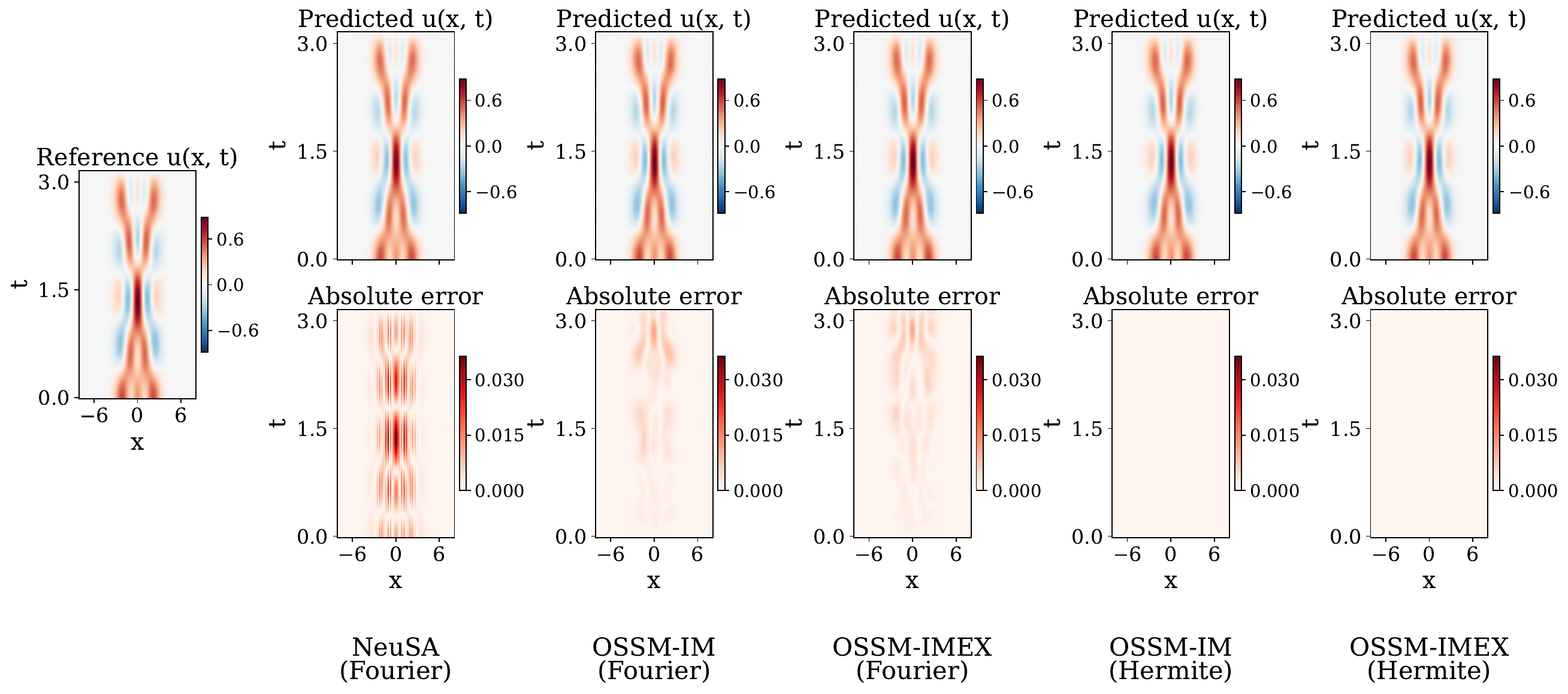}
  \caption{QHO 1D: predicted fields with Fourier vs.\ Hermite basis.}
  \label{fig:basis-qho}
\end{figure}

We solve $i\psi_t = -\frac{1}{2}\psi_{xx} + \frac{1}{2}\omega^2x^2\psi$ on
$[-10,10]\times[0,4\pi]$ with IC
$\psi(x,0)=(\phi_0(x)+\phi_1(x))/\sqrt{2}$ and exact solution
$\psi(x,t)=(\phi_0(x)e^{-iE_0t}+\phi_1(x)e^{-iE_1t})/\sqrt{2}$.
Using the Hermite basis reduces QHO rMAE from $9.5\times10^{-3}$ to
$1.9\times10^{-4}$, a $50\times$ improvement at no architectural cost
(Figure~\ref{fig:basis-qho}).

\subsubsection{Nonlinear Schr\"odinger Equation}
We consider the cubic NLS
$i\psi_t+\tfrac{1}{2}\psi_{xx}+|\psi|^2\psi=0$ on
$(x,t)\in[-5,5]\times[0,\pi/2]$ with periodic boundary conditions
($\psi$ and $\psi_x$ matched at $x=\pm 5$) and the soliton-like
initial condition
\[
\psi(x,0)=2\,\mathrm{sech}(x),
\]
The reference solution is the
analytical Satsuma--Yajima $N{=}2$ breather
\[
\psi(x,t)=\frac{4\,e^{i t/2}\!\left[\cosh(3x)+3\,e^{4it}\cosh(x)\right]}
              {\cosh(4x)+4\cosh(2x)+3\cos(4t)},
\]
which has period $\pi/2$ in $|\psi|$, so the chosen $t$-window covers
exactly one breather cycle. 

\subsubsection{P\"oschl--Teller Well}
\begin{figure}[ht]
  \centering
  \includegraphics[width=\textwidth]{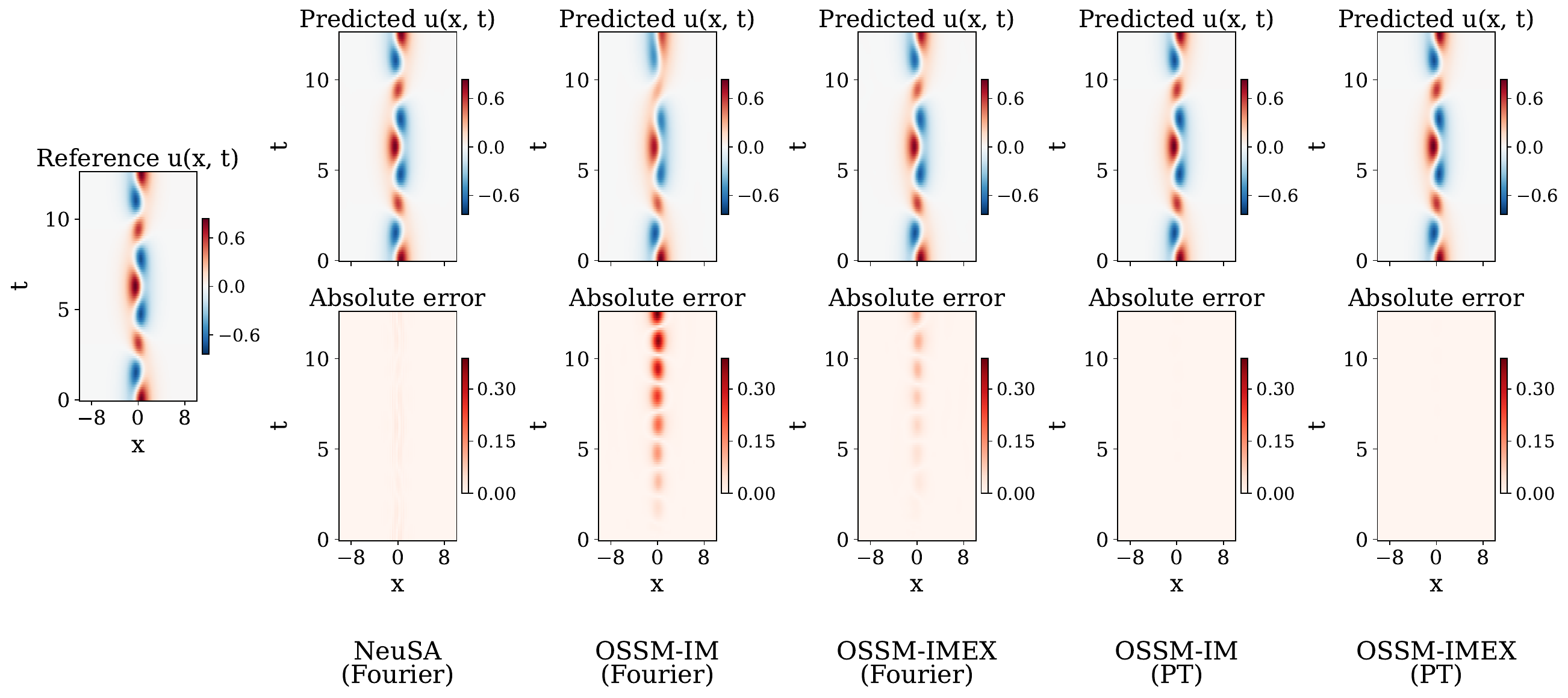}
  \caption{P\"oschl--Teller well: predicted fields with Fourier vs.\ PT basis.}
  \label{fig:basis-pt}
\end{figure}

We solve $i\psi_t+\frac{1}{2}\psi_{xx}+3\operatorname{sech}^2(x)\psi=0$ on
$[-10,10]\times[0,4\pi]$ with IC a superposition of the two exact
bound states $\phi_0(x)\propto \operatorname{sech}^2(x)$ and
$\phi_1(x)\propto \operatorname{sech}(x)\tanh(x)$. Using the PT eigenbasis
reduces rMAE from $6.6\times10^{-2}$ to $1.7\times10^{-3}$, a $40\times$
improvement (Figure~\ref{fig:basis-pt}).

\IfFileExists{figs/F7_bases.pdf}{%
\begin{figure}[ht]
  \centering
  \includegraphics[width=\textwidth]{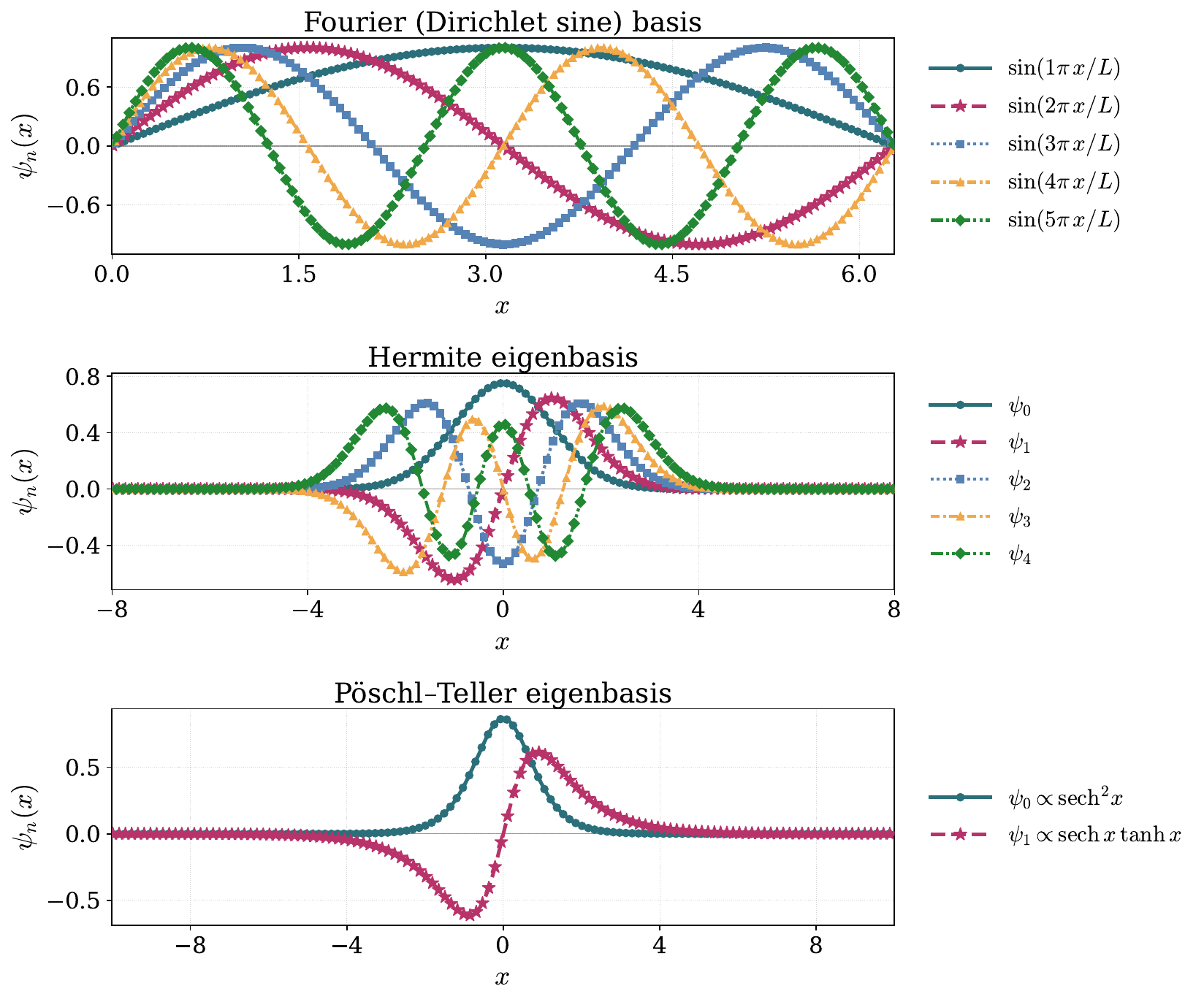}
  \caption{Problem-adapted basis functions.}
  \label{fig:F7-bases}
\end{figure}}{}

\begin{table}[ht]
  \caption{Spatial-basis ablation on quantum benchmarks. Best in \textcolor{bestC}{\textbf{blue}}, second in \textcolor{secondC}{green}, third in \textcolor{thirdC}{red}.}
  \label{tab:basis-full}
  \centering
  \footnotesize
  \setlength{\tabcolsep}{4pt}
  \begin{tabular}{l l l l c c}
  \toprule
  Problem & Method & Basis & Metric & IM & IMEX \\
  \midrule
  \multirow{6}{*}{QHO}
   & \multirow{2}{*}{NeuSA} & \multirow{2}{*}{Fourier}
        & rMAE  & \multicolumn{2}{c}{2.67e-2} \\
   & & & rRMSE & \multicolumn{2}{c}{3.14e-2} \\
   \cmidrule(l){2-6}
   & \multirow{4}{*}{OSSM-PINN}
       & \multirow{2}{*}{Fourier}
        & rMAE  & \textcolor{thirdC}{9.18e-3}\,{\scriptsize\itshape ($\times$3)} & 9.51e-3\,{\scriptsize\itshape ($\times$3)} \\
   & & & rRMSE & \textcolor{thirdC}{9.87e-3}\,{\scriptsize\itshape ($\times$3)} & 9.94e-3\,{\scriptsize\itshape ($\times$3)} \\
   \cmidrule(l){3-6}
   & & \multirow{2}{*}{Hermite}
        & rMAE  & \textbf{\textcolor{bestC}{1.47e-4}}\,{\scriptsize\itshape ($\times$181)} & \textcolor{secondC}{1.87e-4}\,{\scriptsize\itshape ($\times$143)} \\
   & & & rRMSE & \textbf{\textcolor{bestC}{1.46e-4}}\,{\scriptsize\itshape ($\times$215)} & \textcolor{secondC}{1.81e-4}\,{\scriptsize\itshape ($\times$173)} \\
  \midrule
  \multirow{6}{*}{P\"{o}schl--Teller}
   & \multirow{2}{*}{NeuSA} & \multirow{2}{*}{Fourier}
        & rMAE  & \multicolumn{2}{c}{\textcolor{thirdC}{1.77e-2}} \\
   & & & rRMSE & \multicolumn{2}{c}{\textcolor{thirdC}{2.39e-2}} \\
   \cmidrule(l){2-6}
   & \multirow{4}{*}{OSSM-PINN}
       & \multirow{2}{*}{Fourier}
        & rMAE  & 1.92e-1 & 6.63e-2 \\
   & & & rRMSE & 2.74e-1 & 8.65e-2 \\
   \cmidrule(l){3-6}
   & & \multirow{2}{*}{P\"{o}schl--Teller}
        & rMAE  & \textcolor{secondC}{2.19e-3}\,{\scriptsize\itshape ($\times$8)} & \textbf{\textcolor{bestC}{1.68e-3}}\,{\scriptsize\itshape ($\times$11)} \\
   & & & rRMSE & \textcolor{secondC}{2.96e-3}\,{\scriptsize\itshape ($\times$8)} & \textbf{\textcolor{bestC}{2.40e-3}}\,{\scriptsize\itshape ($\times$10)} \\
  \bottomrule
  \end{tabular}
\end{table}

\section{High-Dimensional, Geometry, and Large-Domain Experiments}
\label{app:highdim_geometry}

\subsection{Taylor--Green Vortex (2D)}

\begin{figure}[ht]
  \centering
  \includegraphics[width=\textwidth]{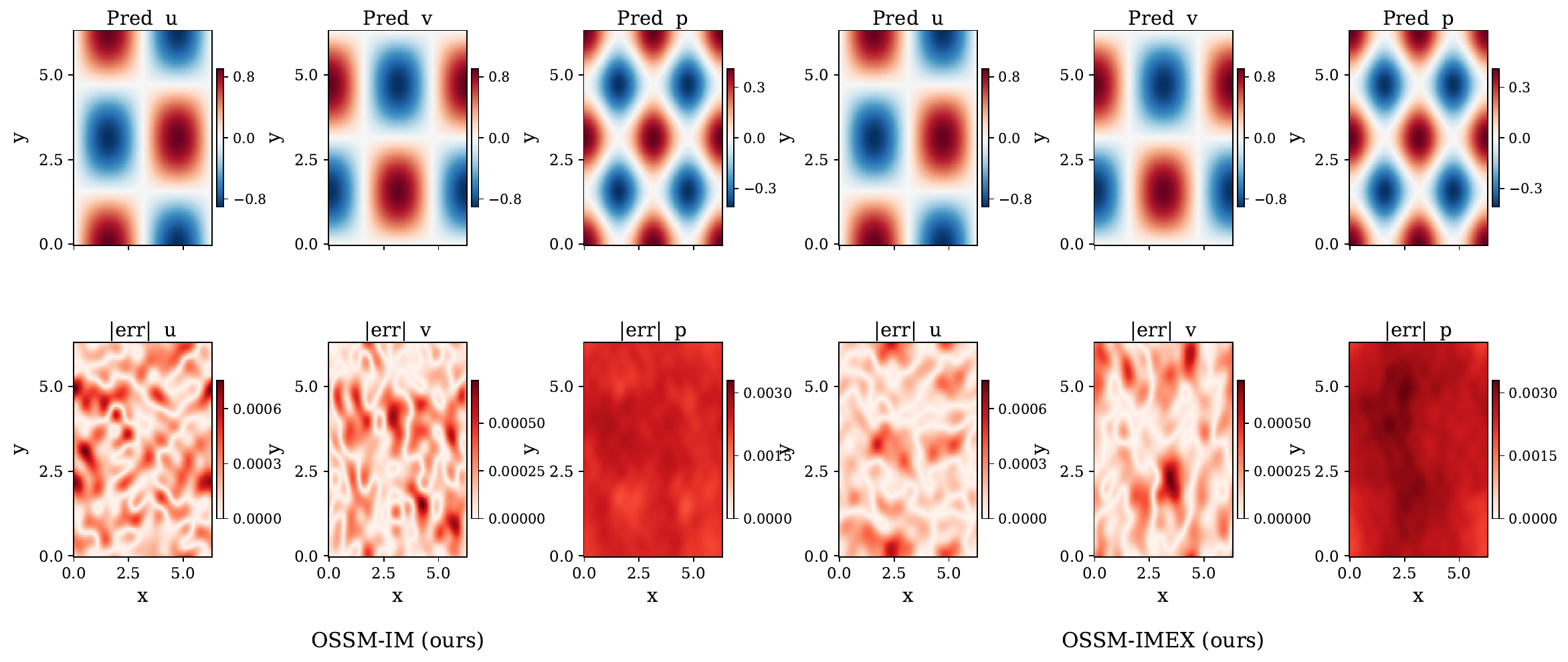}
  \caption{Taylor--Green vortex 2D ($\mathrm{Re} = 100$): predicted velocity components $u(x,y)$, $v(x,y)$ and pressure $p(x,y)$, with absolute errors.}
  \label{fig:sup-tgv}
\end{figure}

Figure~\ref{fig:sup-tgv} shows the predicted velocity and pressure fields at
$t = t_{\max}/2$. The temporal evolution under $\mathrm{Re} = 100$ is dominated
by a slow $e^{-2 \nu t}$ decay ($\nu = 0.01$, $t \le 10$). All four baselines
exceeded the single-GPU memory budget.

\subsection{Heat Equation on Triangular Domain}
\label{app:nonrectangular_geometry}

We solve $u_t = 0.1\Delta u + f(x,y,t)$ on the right triangle
$\mathcal{T}=\{(x,y):x\ge 0,\;y\ge 0,\;x+y\le 1\}$ with $t\in[0,1]$ and
homogeneous Dirichlet boundary conditions. The manufactured solution is
$u(x,y,t)=\sin(\pi x)\sin(\pi y)\sin(\pi(1-x-y))\cos(2\pi t)$, which
vanishes on all three edges including the hypotenuse $x+y=1$. The forcing is
$f(x,y,t)=u_t-0.1\Delta u$. All baselines are either OOM or geometry-infeasible.

\begin{figure}[ht]
  \centering
  \includegraphics[width=\textwidth]{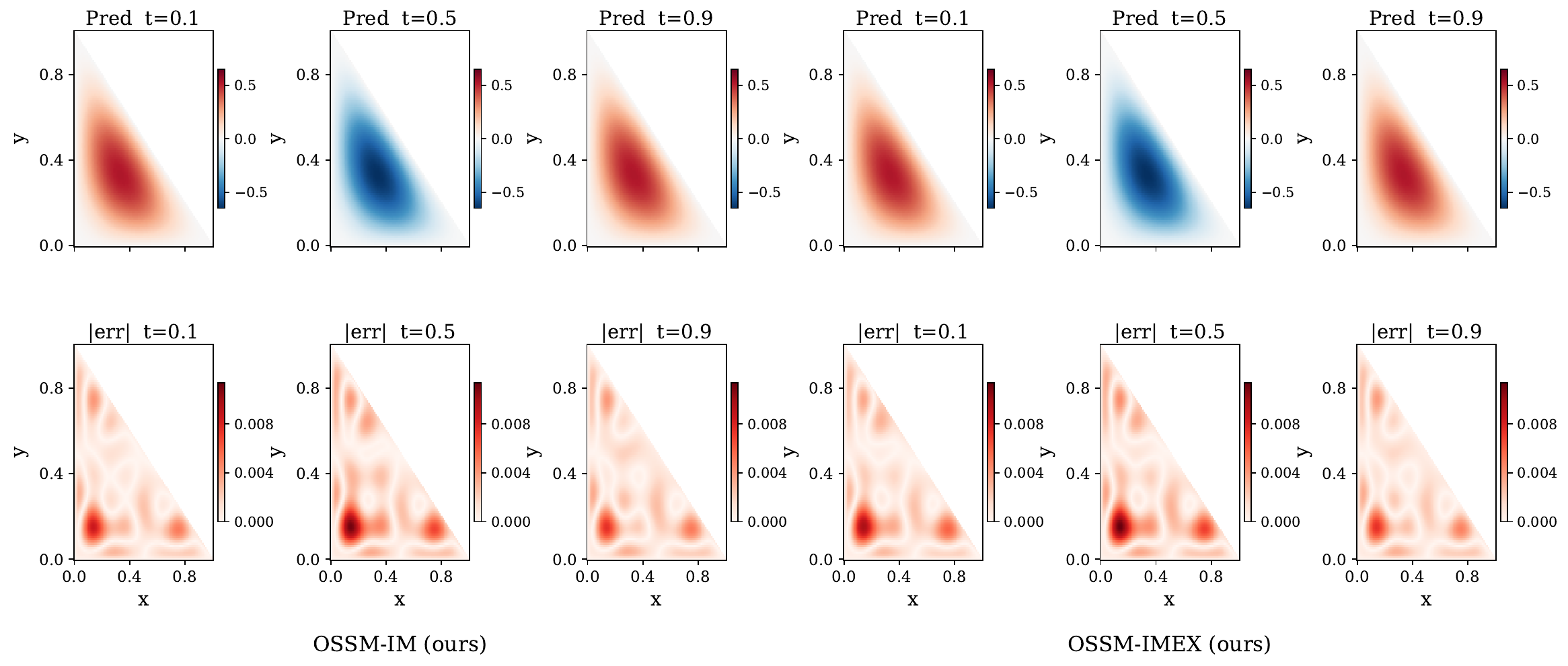}
  \caption{Heat equation on a triangular domain: predicted $u(x,y,t)$ at $t \in \{0.1, 0.5, 0.9\}$, showing reference, OSSM-IM prediction, and absolute error.}
  \label{fig:sup-triangle}
\end{figure}

\subsection{High-Dimensional Schr\"odinger (5D and 100D)}

\begin{figure}[ht]
  \centering
  \includegraphics[width=\textwidth]{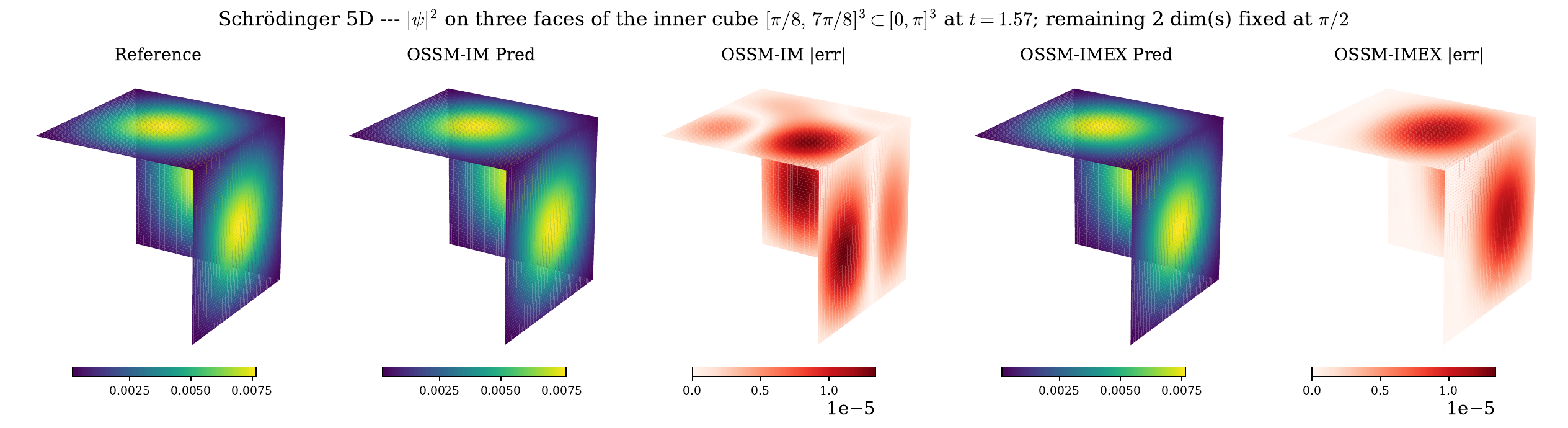}
  \caption{Schr\"odinger 5D: $|\psi(x_1,\ldots,x_5,t)|^2$ at $t = \pi/2$, showing a slice in $(x_1,x_2,x_3)$. Reference, OSSM-IM prediction, and absolute error.}
  \label{fig:sup-schrodinger-5d}
\end{figure}

\begin{figure}[ht]
  \centering
  \includegraphics[width=\textwidth]{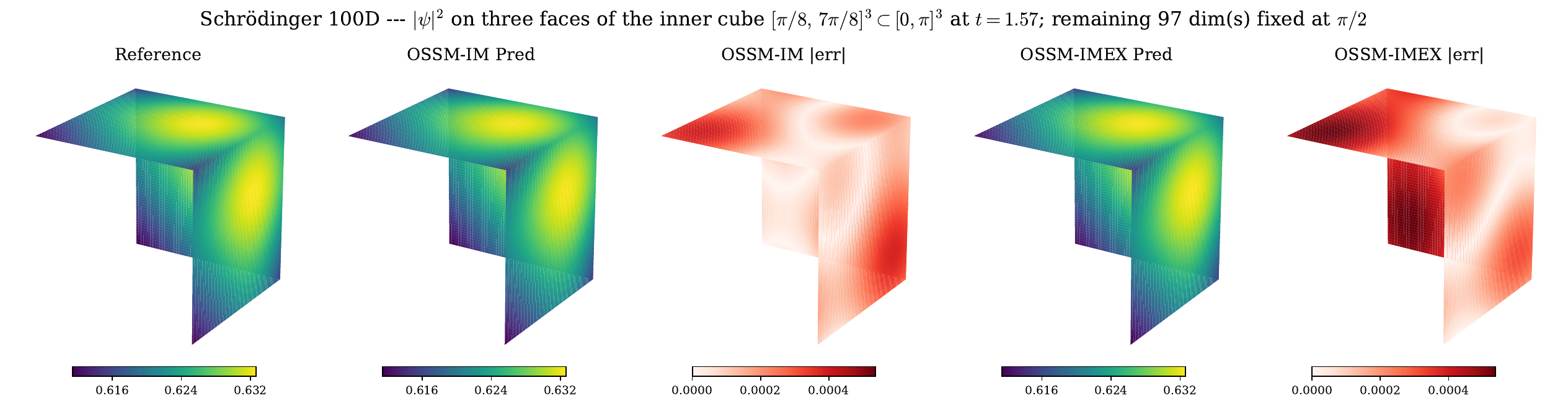}
  \caption{Schr\"odinger 100D: $|\psi(x_1,\ldots,x_{100},t)|^2$ at $t = \pi/2$, showing a slice in $(x_1,x_2, x_3)$. Reference, OSSM-IM prediction, and absolute error.}
  \label{fig:sup-schrodinger-100d}
\end{figure}

Figures~\ref{fig:sup-schrodinger-5d} and~\ref{fig:sup-schrodinger-100d} show
$|\psi|^2$ onto the three visible faces of an inner cube at $t =
\pi/2$. The three varying coordinates are $x_1, x_2, x_3$; remaining
dimensions are fixed at $\pi/2$. All four baselines are infeasible at these
dimensionalities.

\subsection{Large-Domain Euler--Bernoulli Beam}
\label{app:large_domain}

We extend the Euler--Bernoulli beam from $x\in[0,\pi]$ to $x\in[0,8\pi]$
(Exp.~7 in Table~\ref{tab:experiment-suite}), keeping the same PDE, forcing,
IC, and exact solution $u(x,t)=\sin x\cos(4\pi t)$. The field comparison is shown in
Figure~\ref{fig:fwd-eb-extended} and the spectral comparison in
Figure~\ref{fig:sup-freq-eb-ext}.

% ======================================================================
\section{Hyperparameters and Training Recipes}
\label{app:hyperparameters}

Shared hyperparameters for OSSM-PINN are listed in Table~\ref{tab:s4-ossm-recipe}; baseline hyperparameters are in Tables~\ref{tab:s5-psf-recipe}--\ref{tab:s8-neusa-recipe}.

\begin{table}[ht]
  \caption{OSSM-PINN hyperparameters.}
  \label{tab:s4-ossm-recipe}
  \centering
  \footnotesize
  \setlength{\tabcolsep}{6pt}
  \renewcommand{\arraystretch}{1.05}
  \begin{tabular}{l l}
  \toprule
  Hyperparameter & Value \\
  \midrule
  LinOSS hidden width                              & $128$ \\
  Coefficient MLP width                             & $128$ \\
  Locally-adaptive activation                       & enabled \\
  Discretization                                    & IM / IMEX \\
  Adam learning rate                                & $10^{-3}$ \\
  L-BFGS history                                    & $50$ \\
  L-BFGS line search                                & strong Wolfe \\
  L-BFGS learning rate                              & $1.0$ \\
  PDE residual $\lambda_{\mathrm{pde}}$            & $1$ \\
  Initial condition $\lambda_{\mathrm{ic}}$         & $10^{2}$ \\
  Data fit (inverse) $\lambda_{\mathrm{data}}$      & $10^{2}$ \\
  GPU                                               & $1\times$ NVIDIA RTX 5090, 24\,GB \\
  Framework                                         & PyTorch 2.x \\
  \bottomrule
  \end{tabular}
\end{table}

\begin{table}[ht]
  \caption{PINNsFormer hyperparameters.}
  \label{tab:s5-psf-recipe}
  \centering
  \footnotesize
  \setlength{\tabcolsep}{6pt}
  \renewcommand{\arraystretch}{1.05}
  \begin{tabular}{l l}
  \toprule
  Hyperparameter & Value \\
  \midrule
  Embedding width                                  & $64$ \\
  FFN hidden width                                  & $288$ \\
  Pseudo-sequence length                            & $3$ \\
  Attention heads                                   & $2$ \\
  Encoder / decoder layers                          & $1 / 1$ \\
  Activation                                        & wavelet $\omega_1 \sin(x) + \omega_2 \cos(x)$ \\
  Pseudo-sequence step                  & $10^{-3}$ \\
  Adam learning rate                                & $10^{-3}$ \\
  L-BFGS history                                    & $50$ \\
  L-BFGS line search                                & strong Wolfe \\
  L-BFGS learning rate                              & $1.0$ \\
  PDE residual $\lambda_{\mathrm{pde}}$            & $1$ \\
  Initial condition $\lambda_{\mathrm{ic}}$         & $10^{2}$ \\
  Boundary $\lambda_{\mathrm{bc}}$                  & $1$ \\
  \bottomrule
  \end{tabular}
\end{table}

\begin{table}[ht]
  \caption{PINNMamba hyperparameters.}
  \label{tab:s6-pmb-recipe}
  \centering
  \footnotesize
  \setlength{\tabcolsep}{6pt}
  \renewcommand{\arraystretch}{1.05}
  \begin{tabular}{l l}
  \toprule
  Hyperparameter & Value \\
  \midrule
  Embedding width                                  & $80$ \\
  SSM state size                                    & $16$ \\
  Convolution width                                 & $4$ \\
  Expansion factor                                  & $2$ \\
  Pseudo-sequence length                            & $2$ \\
  Number of Mamba blocks                            & $2$ \\
  Activation                                        & wavelet $\omega_1 \sin(x) + \omega_2 \cos(x)$ \\
  Pseudo-sequence step $\Delta t$                  & $10^{-3}$ \\
  Adam learning rate                                & $10^{-3}$ \\
  L-BFGS history                                    & $50$ \\
  L-BFGS line search                                & strong Wolfe \\
  L-BFGS learning rate                              & $1.0$ \\
  PDE residual $\lambda_{\mathrm{pde}}$            & $1$ \\
  Initial condition $\lambda_{\mathrm{ic}}$         & $10^{2}$ \\
  Boundary $\lambda_{\mathrm{bc}}$                  & $1$ \\
  \bottomrule
  \end{tabular}
\end{table}

\begin{table}[ht]
  \caption{ML-PINN hyperparameters.}
  \label{tab:s7-mlpinn-recipe}
  \centering
  \footnotesize
  \setlength{\tabcolsep}{6pt}
  \renewcommand{\arraystretch}{1.05}
  \begin{tabular}{l l}
  \toprule
  Hyperparameter & Value \\
  \midrule
  Mamba embedding width                             & $80$ \\
  Mamba SSM state size                              & $8$ \\
  Mamba convolution width                           & $4$ \\
  Mamba expansion factor                            & $2$ \\
  Mamba blocks                                      & $1$ \\
  LSTM hidden width                                 & $16$ \\
  LSTM layers                                       & $1$, bidirectional \\
  Pseudo-sequence length                            & $4$ \\
  Pseudo-sequence step                  & $10^{-2}$ \\
  Activation                                        & wavelet $\omega_1 \sin(x) + \omega_2 \cos(x)$ \\
  Adam learning rate                                & $10^{-3}$ \\
  L-BFGS history                                    & $50$ \\
  L-BFGS line search                                & strong Wolfe \\
  L-BFGS learning rate                              & $1.0$ \\
  PDE residual $\lambda_{\mathrm{pde}}$            & $1$ \\
  Initial condition $\lambda_{\mathrm{ic}}$         & $10^{2}$ \\
  Boundary $\lambda_{\mathrm{bc}}$                  & $1$ \\
  \bottomrule
  \end{tabular}
\end{table}

\begin{table}[ht]
  \caption{NeuSA hyperparameters.}
  \label{tab:s8-neusa-recipe}
  \centering
  \footnotesize
  \setlength{\tabcolsep}{6pt}
  \renewcommand{\arraystretch}{1.05}
  \begin{tabular}{l l}
  \toprule
  Hyperparameter & Value \\
  \midrule
  Spectral basis                                    & Fourier (periodic) / sine (Dirichlet) \\
  Spectral state size                           & $51$ \\
  Neural correction MLP                             & hidden $272$, depth $2$ \\
  Correction weight $\varepsilon$                  & $0.1$ \\
  Time integrator                                   & explicit RK4 \\
  Adam learning rate                                & $10^{-3}$ \\
  L-BFGS history                                    & $50$ \\
  L-BFGS line search                                & strong Wolfe \\
  L-BFGS learning rate                              & $1.0$ \\
  PDE residual $\lambda_{\mathrm{pde}}$            & $1$ \\
  Initial condition $\lambda_{\mathrm{ic}}$         & $10^{2}$ \\
  Boundary $\lambda_{\mathrm{bc}}$                  & $1$ \\
  \bottomrule
  \end{tabular}
\end{table}

\section{Computational Cost}
\label{app:cost}

Table~\ref{tab:s2-cost} reports trainable parameter count, peak GPU memory
(MiB), and single-batch inference latency (ms) for every (problem, method)
pair. OSSM-PINN consistently uses less peak GPU memory than all
sequence-model baselines that can run (as low as 144~MiB on wave versus
1,454~MiB for PINNsFormer and 6,301~MiB for PINNMamba), and inference
latency is $8$--$23\times$ lower.

\begin{table}[ht]
  \caption{Computational cost: parameters (k), peak GPU memory (MiB), and inference latency (ms).}
  \label{tab:s2-cost}
  \centering
  \footnotesize
  \setlength{\tabcolsep}{2pt}
  \renewcommand{\arraystretch}{1.0}
  \resizebox{\textwidth}{!}{%
\begin{tabular}{c l l c c c c c c}
\toprule
& PDE & Quantity & PINNsFormer & PINNMamba & ML-PINN & NeuSA & \makecell{OSSM-PINN-IM\\\scriptsize\emph{(ours)}} & \makecell{OSSM-PINN-IMEX\\\scriptsize\emph{(ours)}} \\
\midrule
\multirow{18}{*}{\rotatebox[origin=c]{90}{Forward}} & \multirow{3}{*}{Convection ($\beta{=}50$)} & \#params (k) & 108.0 & 107.6 & 94.9 & 102.3 & 93.5 & 93.5 \\
 &  & peak GPU (MiB) & 547 & 1734 & 1269 & 98 & 153 & 154 \\
 &  & latency (ms) & 88.65 & 15.66 & 9.52 & 31.95 & 3.84 & 3.79 \\
\cmidrule(lr){2-9}
 & \multirow{3}{*}{Convection ($\beta{=}100$)} & \#params (k) & 108.0 & 107.6 & 94.9 & 102.3 & 93.5 & 93.5 \\
 &  & peak GPU (MiB) & 547 & 1734 & 1270 & 134 & 215 & 215 \\
 &  & latency (ms) & 51.58 & 16.09 & 11.01 & 31.18 & 7.85 & 7.82 \\
\cmidrule(lr){2-9}
 & \multirow{3}{*}{Reaction} & \#params (k) & 132.7 & 139.1 & 128.8 & 129.8 & 116.2 & 116.2 \\
 &  & peak GPU (MiB) & 454 & 1763 & 1320 & 87 & 154 & 154 \\
 &  & latency (ms) & 73.61 & 25.85 & 16.87 & 28.93 & 2.58 & 2.61 \\
\cmidrule(lr){2-9}
 & \multirow{3}{*}{Wave} & \#params (k) & 124.5 & 139.1 & 128.0 & 124.2 & 109.8 & 109.8 \\
 &  & peak GPU (MiB) & 1454 & 6301 & 3661 & 85 & 144 & 144 \\
 &  & latency (ms) & 65.28 & 25.77 & 14.04 & 32.47 & 2.59 & 2.59 \\
\cmidrule(lr){2-9}
 & \multirow{3}{*}{Euler-Bernoulli (classical)} & \#params (k) & \multirow{3}{*}{\textcolor{stubGray}{\scriptsize\textsc{OOM}}} & \multirow{3}{*}{\textcolor{stubGray}{\scriptsize\textsc{OOM}}} & 94.9 & 102.5 & 92.3 & 92.3 \\
 &  & peak GPU (MiB) &  &  & 9409 & 76 & 131 & 130 \\
 &  & latency (ms) &  &  & 10.34 & 32.50 & 2.46 & 2.51 \\
\cmidrule(lr){2-9}
 & \multirow{3}{*}{Euler-Bernoulli (extended)} & \#params (k) & \multirow{3}{*}{\textcolor{stubGray}{\scriptsize\textsc{OOM}}} & \multirow{3}{*}{\textcolor{stubGray}{\scriptsize\textsc{OOM}}} & 94.9 & 102.5 & 93.3 & 93.3 \\
 &  & peak GPU (MiB) &  &  & 9410 & 76 & 187 & 187 \\
 &  & latency (ms) &  &  & 10.38 & 32.52 & 2.59 & 2.56 \\
\midrule
\midrule
\multirow{12}{*}{\rotatebox[origin=c]{90}{Forward high-dim}} & \multirow{3}{*}{Taylor-Green 2D} & \#params (k) & \multirow{3}{*}{\textcolor{stubGray}{\scriptsize\textsc{OOM}}} & \multirow{3}{*}{\textcolor{stubGray}{\scriptsize\textsc{OOM}}} & \multirow{3}{*}{\textcolor{stubGray}{\scriptsize\textsc{OOM}}} & \multirow{3}{*}{\textcolor{stubGray}{\scriptsize\textsc{OOM}}} & 604.6 & 604.6 \\
 &  & peak GPU (MiB) &  &  &  &  & 809 & 809 \\
 &  & latency (ms) &  &  &  &  & 4.05 & 3.95 \\
\cmidrule(lr){2-9}
 & \multirow{3}{*}{Heat (triangular)} & \#params (k) & \multirow{3}{*}{\textcolor{stubGray}{\scriptsize\textsc{OOM}}} & \multirow{3}{*}{\textcolor{stubGray}{\scriptsize\textsc{OOM}}} & \multirow{3}{*}{\textcolor{stubGray}{\scriptsize\textsc{OOM}}} & \multirow{3}{*}{\textcolor{stubGray}{\scriptsize\textsc{IG}}} & 615.6 & 615.6 \\
 &  & peak GPU (MiB) &  &  &  &  & 603 & 602 \\
 &  & latency (ms) &  &  &  &  & 1.95 & 1.89 \\
\cmidrule(lr){2-9}
 & \multirow{3}{*}{Schr{\"o}dinger (5D)} & \#params (k) & \multirow{3}{*}{\textcolor{stubGray}{\scriptsize\textsc{OOM}}} & \multirow{3}{*}{\textcolor{stubGray}{\scriptsize\textsc{OOM}}} & \multirow{3}{*}{\textcolor{stubGray}{\scriptsize\textsc{OOM}}} & \multirow{3}{*}{\textcolor{stubGray}{\scriptsize\textsc{OOM}}} & 173.4 & 173.4 \\
 &  & peak GPU (MiB) &  &  &  &  & 405 & 405 \\
 &  & latency (ms) &  &  &  &  & 2.75 & 2.62 \\
\cmidrule(lr){2-9}
 & \multirow{3}{*}{Schr{\"o}dinger (100D)} & \#params (k) & \multirow{3}{*}{\textcolor{stubGray}{\scriptsize\textsc{OOM}}} & \multirow{3}{*}{\textcolor{stubGray}{\scriptsize\textsc{OOM}}} & \multirow{3}{*}{\textcolor{stubGray}{\scriptsize\textsc{OOM}}} & \multirow{3}{*}{\textcolor{stubGray}{\scriptsize\textsc{OOM}}} & 216.7 & 216.7 \\
 &  & peak GPU (MiB) &  &  &  &  & 4328 & 4328 \\
 &  & latency (ms) &  &  &  &  & 5.22 & 6.40 \\
\midrule
\midrule
\multirow{6}{*}{\rotatebox[origin=c]{90}{Inverse}} & \multirow{3}{*}{KdV} & \#params (k) & \multirow{3}{*}{\textcolor{stubGray}{\scriptsize\textsc{OOM}}} & \multirow{3}{*}{\textcolor{stubGray}{\scriptsize\textsc{OOM}}} & \multirow{3}{*}{\textcolor{stubGray}{\scriptsize\textsc{OOM}}} & \multirow{3}{*}{\textcolor{stubGray}{\scriptsize\textsc{ST}}} & 116.2 & 116.2 \\
 &  & peak GPU (MiB) &  &  &  &  & 151 & 151 \\
 &  & latency (ms) &  &  &  &  & 1.97 & 1.92 \\
\cmidrule(lr){2-9}
 & \multirow{3}{*}{SST 2D adv-diff} & \#params (k) & \multirow{3}{*}{\textcolor{stubGray}{\scriptsize\textsc{OOM}}} & \multirow{3}{*}{\textcolor{stubGray}{\scriptsize\textsc{OOM}}} & \multirow{3}{*}{\textcolor{stubGray}{\scriptsize\textsc{OOM}}} & \multirow{3}{*}{\textcolor{stubGray}{\scriptsize\textsc{OOM}}} & 202.4 & 202.4 \\
 &  & peak GPU (MiB) &  &  &  &  & 312 & 313 \\
 &  & latency (ms) &  &  &  &  & 1.45 & 1.37 \\
\bottomrule
\end{tabular}}
\end{table}

\section{Training Dynamics and Seed Stability}
\label{app:training_dynamics}

\begin{figure}[ht]
  \centering
  \includegraphics[width=\textwidth]{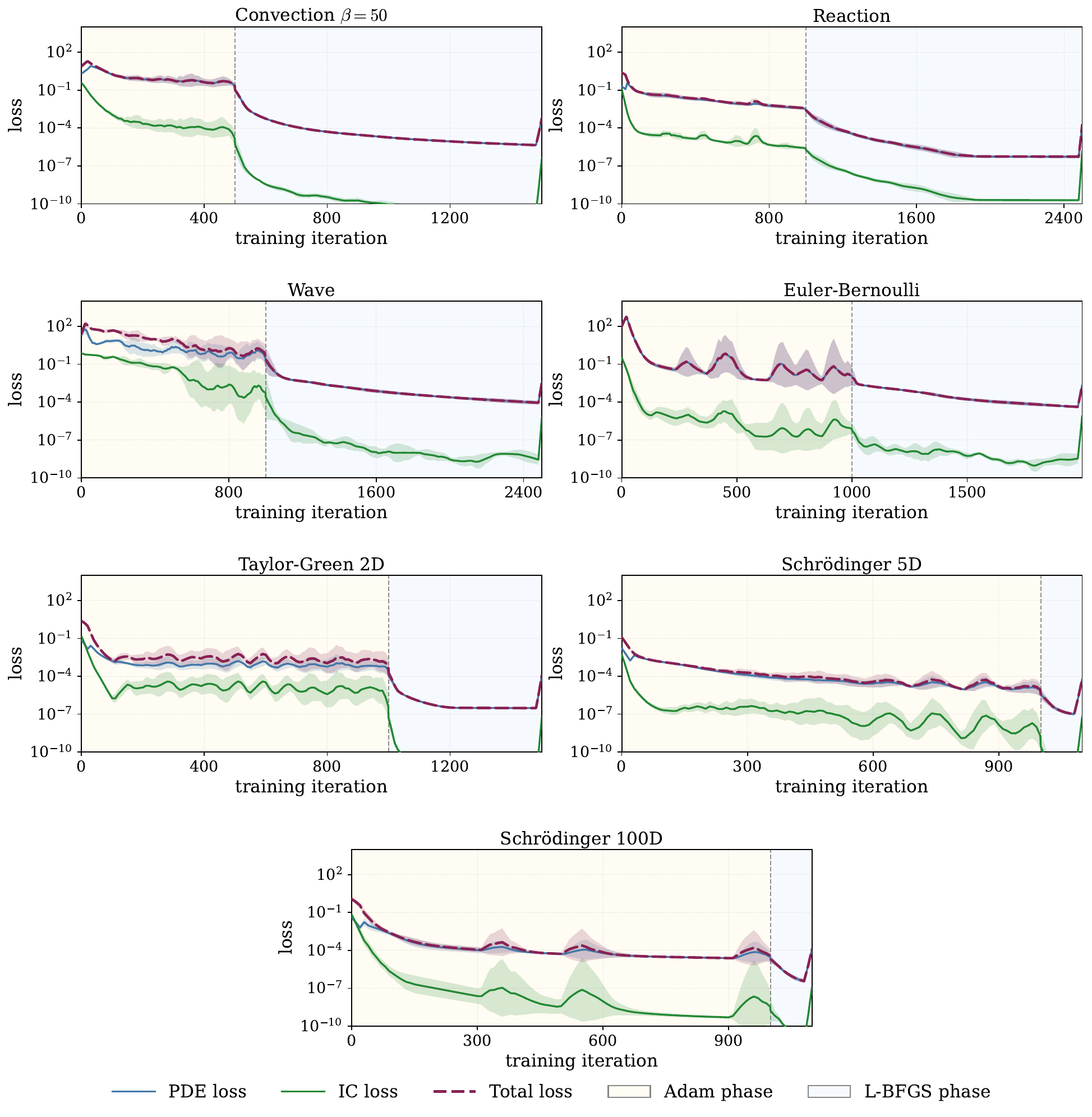}
  \caption{Training-loss histories for OSSM-PINN-IMEX (three seeds).}
  \label{fig:sup-loss-curves}
\end{figure}

\begin{figure}[ht]
  \centering
  \includegraphics[width=\textwidth]{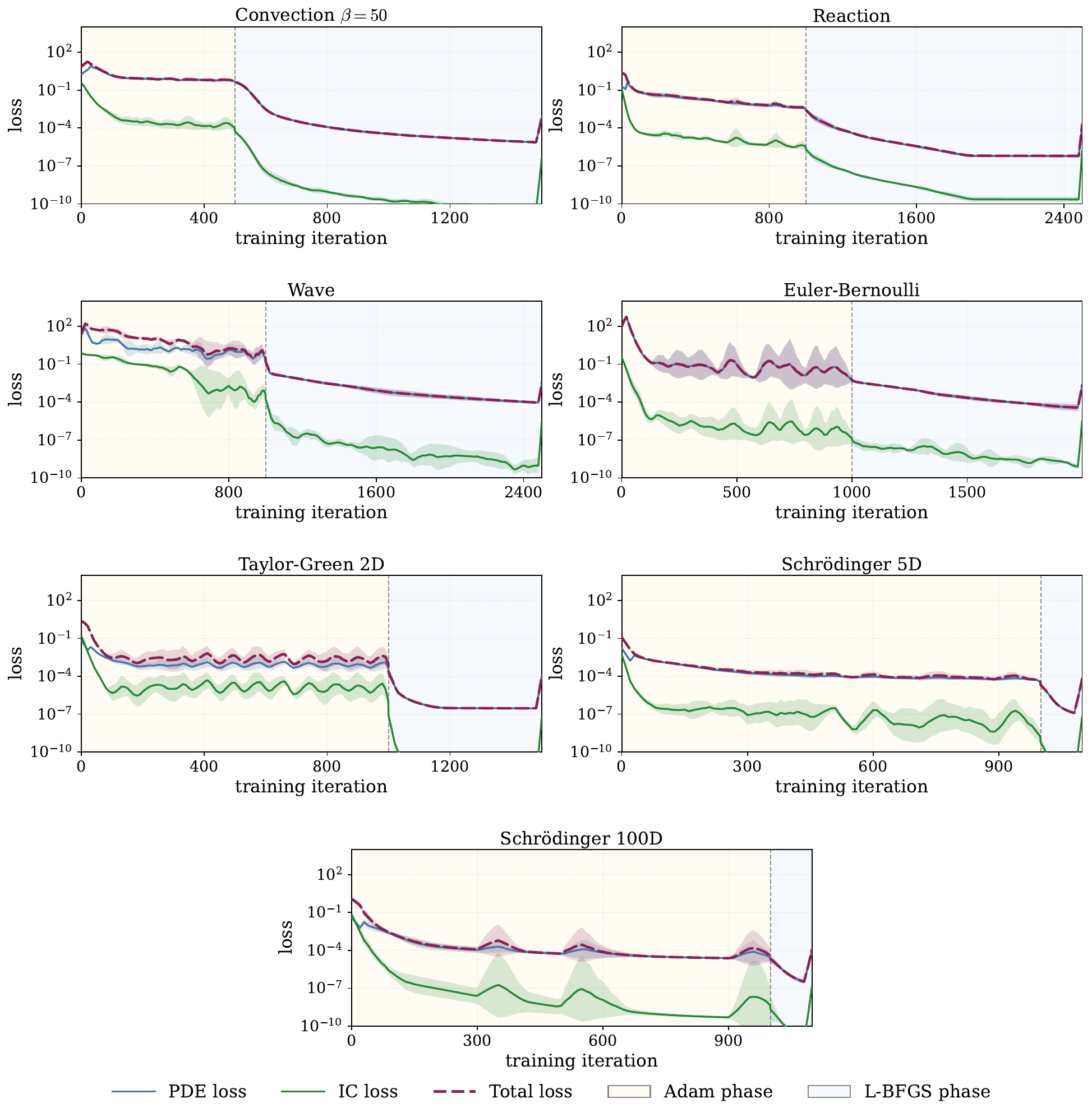}
  \caption{Training-loss histories for OSSM-PINN-IM (three seeds).}
  \label{fig:sup-loss-curves-im}
\end{figure}

\textbf{Loss curves:}
Figures~\ref{fig:sup-loss-curves} and~\ref{fig:sup-loss-curves-im} show
per-problem training-loss histories for OSSM-PINN-IMEX and OSSM-PINN-IM,
aggregated over three seeds. Each panel shows the PDE residual loss, the IC
loss, and the total weighted loss; background tints separate the Adam warm-up
(cream) from the L-BFGS phase (light blue).

\begin{figure}[ht]
  \centering
  \includegraphics[width=0.70\textwidth]{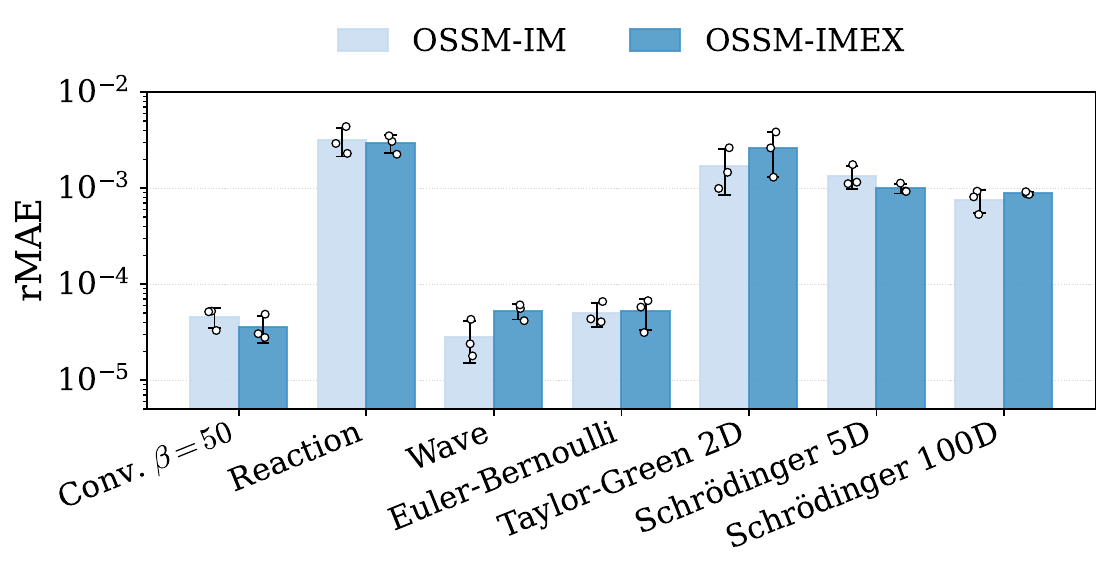}
  \caption{Seed stability: mean rMAE $\pm 1$ std over three seeds for seven benchmarks.}
  \label{fig:seed-stability}
\end{figure}

\textbf{Seed stability:}
Figure~\ref{fig:seed-stability} shows mean rMAE over three seeds for seven
benchmarks; error bars show $\pm 1$ std and open circles mark individual seed
values. Variability between seeds stays well within the same order of magnitude
as the mean for both discretizations.

\end{document}

%% file: F3_algorithm.tex
\begin{wrapfigure}{r}{0.50\textwidth}
\vspace{-20pt}
\begin{minipage}{0.50\textwidth}
\begin{algorithm}[H]
\caption{OSSM-PINN training.}
\label{alg:ossm-pinn}
\small
\begin{algorithmic}[1]
\Require Operator $\mathcal{F}$, IC $\mathrm{u}_0$, basis $\{\phi_k\}$;
\textcolor{ourTeal}{\textit{(inv)}} data
$\{(\mathrm{x}_i,\mathrm{t}_i,\mathrm{u}_i^{\text{obs}})\}$,
loss weights $\boldsymbol\lambda$
\For{$\text{iter} = 1,\dots,N$}
  \State $(y_0, z_0) \gets g_e(\mathrm{u}_0)$;\quad
         $\{h_n\}_{n=0}^{N_t} \gets \mathrm{LinOSS}(y_0, z_0)$
  \State $\widehat{\mathrm{u}}(\mathrm{x}_i, \mathrm{t}_n)
         \gets \sum_k c_k(h_n)\,\phi_k(\mathrm{x}_i)$
  \State $\mathcal{L} \gets
         \lambda_{\text{PDE}}\|\mathcal{R}(\widehat{\mathrm{u}})\|^2
         + \lambda_{\text{IC}}\|\widehat{\mathrm{u}}(\cdot,0)
         {-}\mathrm{u}_0\|^2$
  \State \textcolor{ourTeal}{\textit{(inv)}}
         $\;\mathcal{L} \mathrel{+}=
         \lambda_{\text{D}}\sum_i
         (\widehat{\mathrm{u}}(\mathrm{x}_i,\mathrm{t}_i){-}
         \mathrm{u}_i^{\text{obs}})^2$
  \State Update $\theta$
         \textcolor{ourTeal}{(\,and unknown PDE parameters
         if inverse\,)}
\EndFor
\end{algorithmic}
\end{algorithm}
\end{minipage}
\vspace{-10pt}
\end{wrapfigure}

%% file: T1_method_attributes.tex
\newcommand{\Tcmark}{\textcolor{ourTeal}{$\checkmark$}}      % teal check
\newcommand{\Txmark}{\textcolor{psfRed}{$\times$}}            % red cross
\newcommand{\Tpmark}{\textcolor{neusaOrange}{$\sim$}}        % amber tilde

\begin{table}[t]
  \centering
  \footnotesize
  \setlength{\tabcolsep}{6pt}
  \renewcommand{\arraystretch}{1.18}
  \caption{Method-attribute comparison across PINN-family
  architectures.  \Tcmark{} = supported, \Txmark{} = not supported,
  \Tpmark{} = partial / problem-dependent.}
  \label{tab:T1-attributes}
  \begin{tabular}{l c c c}
  \toprule
   & PINNsFormer / Mamba & NeuSA & \makecell{\textbf{OSSM-PINN}\\\textbf{(ours)}} \\
  \midrule
  No PDE-class hand-coded operator                & \Tcmark & \Txmark & \Tcmark \\
  Causal time treatment                           & \Tpmark & \Tcmark & \Tcmark \\
  Inductive bias for oscillatory dynamics         & \Txmark & \Tcmark & \Tcmark \\
  Learnable time-evolution operator               & \Txmark & \Txmark & \Tcmark \\
  Symplectic time-stepping option                 & \Txmark & \Txmark & \Tcmark \\
  Modal / interpretable latent state              & \Txmark & \Tcmark & \Tcmark \\
  Hard BC enforcement via spatial basis           & \Txmark & \Tcmark & \Tcmark \\
  High-frequency dynamics without failure modes   & \Txmark & \Tpmark & \Tcmark \\
  Higher-order spatial operators                  & \Txmark & \Tpmark & \Tcmark \\
  Inverse-problem capability                      & \Tcmark & \Txmark & \Tcmark \\
  Curse-of-dim friendly ($d \ge 5$)               & \Txmark & \Txmark & \Tcmark \\
  \bottomrule
  \end{tabular}
\end{table}